\documentclass{article} 

\usepackage[accepted]{icml2026}

\usepackage{amsmath,amsfonts,bm}

\def\eqref#1{equation~\ref{#1}}

\def\1{\bm{1}}

\def\vh{{\bm{h}}}

\def\vx{{\bm{x}}}
\def\vy{{\bm{y}}}

\def\mH{{\bm{H}}}

\DeclareMathAlphabet{\mathsfit}{\encodingdefault}{\sfdefault}{m}{sl}
\SetMathAlphabet{\mathsfit}{bold}{\encodingdefault}{\sfdefault}{bx}{n}

\def\gA{{\mathcal{A}}}

\def\gC{{\mathcal{C}}}
\def\gD{{\mathcal{D}}}
\def\gE{{\mathcal{E}}}

\def\gM{{\mathcal{M}}}

\def\gO{{\mathcal{O}}}
\def\gP{{\mathcal{P}}}

\def\gR{{\mathcal{R}}}
\def\gS{{\mathcal{S}}}
\def\gT{{\mathcal{T}}}

\usepackage[T1]{fontenc}
\usepackage{hyperref}
\usepackage{url}
\usepackage{xspace}
\usepackage{booktabs}
\usepackage{multirow}
\usepackage{amsmath}
\usepackage{array}
\usepackage{caption}
\usepackage{graphicx}
\usepackage{subcaption}
\usepackage{algorithm}
\usepackage{algorithmic}
\usepackage{enumitem}
\usepackage{colortbl}
\usepackage{mathtools}
\usepackage{amssymb}
\usepackage[table]{xcolor}
\usepackage{amssymb} 
\usepackage{pifont}  
\usepackage[most]{tcolorbox}
\usepackage{listings}
\usepackage{minted}
\usepackage{xspace}
\usepackage{boxedminipage}
\usepackage{adjustbox}
\usepackage{tabularx}
\usepackage{wrapfig}
\tcbuselibrary{listings} 

\usepackage[capitalize,noabbrev]{cleveref}

\newcommand{\cmark}{\checkmark} 
\newcommand{\xmark}{\ding{55}}  

\definecolor{Blue}{rgb}{0.152, 0.294, 0.925}
\definecolor{darkgray}{gray}{0.3}
\definecolor{chocolate}{HTML}{D2691E}
\definecolor{maroon}{HTML}{A00000}
\definecolor{indigo}{HTML}{4B0082}
\definecolor{green}{HTML}{008000}
\definecolor{newred}{HTML}{a91e1e}
\definecolor{cadmiumgreen}{rgb}{0.0, 0.42, 0.24}
\definecolor{forestgreen}{rgb}{0.13, 0.55, 0.13}

\newcommand{\ours}{\textsc{Acon}\xspace}

\newcommand{\eg}{\textit{e.g.,}\xspace}
\definecolor{Red}{rgb}{0.768, 0.054, 0.054}
\definecolor{Green}{rgb}{0,0.4,0.7}
\hypersetup{
    colorlinks=true,
    citecolor=teal,
    linkcolor=Red,
    urlcolor=Green,
}
\definecolor{lightgray}{gray}{0.9}
\definecolor{lightblue}{RGB}{230,245,255}

\newcommand{\UTIL}{{\protect\color{chocolate} $\overline{\underline{\textsc{ut}}}$}\xspace}
\newcommand{\COMP}{{\protect\color{indigo} $\overline{\underline{\textsc{co}}}$}\xspace}

\newtcolorbox[auto counter, number within=section]{examplebox}[2][]{
  colback=white, colframe=black, boxrule=0.5pt,
  sharp corners, left=4pt, right=4pt, top=4pt, bottom=4pt,
  enhanced, breakable,
  title=Example~\thetcbcounter: #2, #1
}

\newtcolorbox{observationbox}{
  colback=gray!5, colframe=gray!40, boxrule=0.4pt,
  enhanced, breakable
}

\setminted{breaklines, breakanywhere, fontsize=\small}
\newtcblisting[auto counter, number within=section]{compressbox}[2][]{
  colback=white,
  colframe=green!60!black,  
  boxrule=0.5pt,
  sharp corners,
  left=4pt, right=4pt, top=4pt, bottom=4pt,
  enhanced, breakable,
  width=0.95\linewidth,
  title=Compression~\thetcbcounter: #2,
  listing only,
  listing engine=listings,
  listing options={
    basicstyle=\ttfamily\small,
    breaklines=true,
    columns=fullflexible,
    keepspaces=true
  },
  #1
}

\usepackage[textsize=tiny]{todonotes}

\icmltitlerunning{ACON: Optimizing Context Compression for Long-horizon LLM Agents}

\begin{document}
\twocolumn[
  \icmltitle{\ours: Optimizing Context Compression for Long-horizon LLM Agents}



  \icmlsetsymbol{msintern}{*}

  \begin{icmlauthorlist}
    \icmlauthor{Minki Kang}{kaist,msintern}
    \icmlauthor{Wei-Ning Chen}{ms}
    \icmlauthor{Dongge Han}{ms}
    \icmlauthor{Huseyin A. Inan}{ms}
    \icmlauthor{Lukas Wutschitz}{ms}
    \icmlauthor{Yanzhi Chen}{ms,cam}
    \icmlauthor{Robert Sim}{ms}
    \icmlauthor{Saravan Rajmohan}{ms}
  \end{icmlauthorlist}

  \icmlaffiliation{kaist}{KAIST}
  \icmlaffiliation{ms}{Microsoft}
  \icmlaffiliation{cam}{University of Cambridge}

  \icmlcorrespondingauthor{Minki Kang}{minkikang@kaist.ac.kr}

  \icmlkeywords{Machine Learning, ICML}

  \vskip 0.3in
]




\printAffiliationsAndNotice{\textsuperscript{*}Work done during the internship at Microsoft.}  

\newcommand{\fix}{\marginpar{FIX}}
\newcommand{\new}{\marginpar{NEW}}

\begin{abstract}
Large language models (LLMs) are increasingly deployed as agents in dynamic real-world environments, where success depends on maintaining precise records of actions and observations. 
However, the resulting unbounded context growth in long-horizon agentic tasks makes two critical bottlenecks: prohibitive inference memory costs and reasoning degradation due to irrelevant information.
Existing compression methods fail to fully address this, often relying on brittle heuristics or requiring parameter updates impractical for proprietary or large-scale LLMs.
We introduce \textbf{Agent Context Optimization} (\ours), a unified framework that optimally compresses both observations and history into concise, informative representations. Distinct from prior works, \ours employs an optimization in natural language space: it iteratively refines compression guidelines based on failure analysis of the agent, ensuring critical state information is preserved without model fine-tuning. To further minimize computational overhead, we distill the optimized compressor into smaller models. 
Experiments on AppWorld, OfficeBench, and Multi-objective QA demonstrate that \ours reduces peak token usage by 26–54\% while improving task success over existing compression baselines.
Notably, it enables smaller LMs to function effectively as long-horizon agents, achieving up to 46\% performance improvement by mitigating context distraction.
Our code is available at \href{https://github.com/microsoft/acon}{https://github.com/microsoft/acon}.

\end{abstract}

\section{Introduction}

\begin{figure*}[ht]
    \centering
    \includegraphics[width=1.00\linewidth]{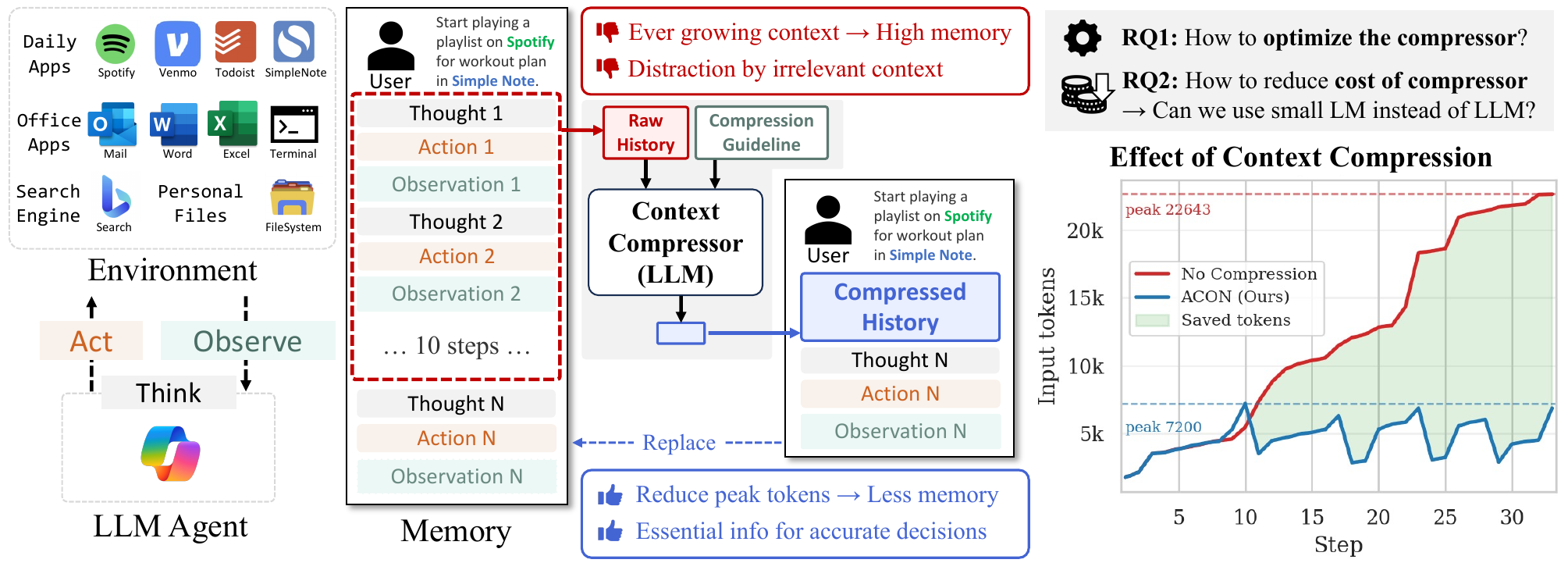}
    \vspace{-0.0in}
    \caption{\textbf{Motivation: Unbounded context in LLM agents.} Continuous interactions lead to ever-growing contexts that incur high memory usage (red line). This motivates the need for compression, yet raises key questions on \textit{how to optimize the compressor} and \textit{reduce its cost}. We address these with \ours through a \textbf{compression guideline optimization} and \textbf{compressor distillation}, effectively reducing peak tokens (blue line) while preserving essential information.}
    \label{fig:motivation}
    \vspace{-0.0in}
\end{figure*}

Large language models (LLMs) have become the backbone of AI agents, enabling them to plan and act in dynamic environments~\citep{ReAct}.
However, these tasks often unfold over extended horizons, requiring the agent to maintain a continuous record of observations, tool outputs, and evolving states.
In such settings, context is not auxiliary but foundational; losing a single detail such as a file path or an API parameter can derail the entire workflow.
As interactions accumulate, context grows unbounded as shown in~\autoref{fig:motivation}, making two major bottlenecks. First, the inference memory cost of transformers scales with context length, resulting long-horizon reasoning computationally prohibitive due to the massive KV cache requirements~\citep{transformer}.
Second, excessively long contexts dilute relevant information, distracting the model with outdated or extraneous details and degrading decision quality~\citep{irrelevant-context}.

\begin{figure}[t]
    \centering
    \includegraphics[width=1.00\linewidth]{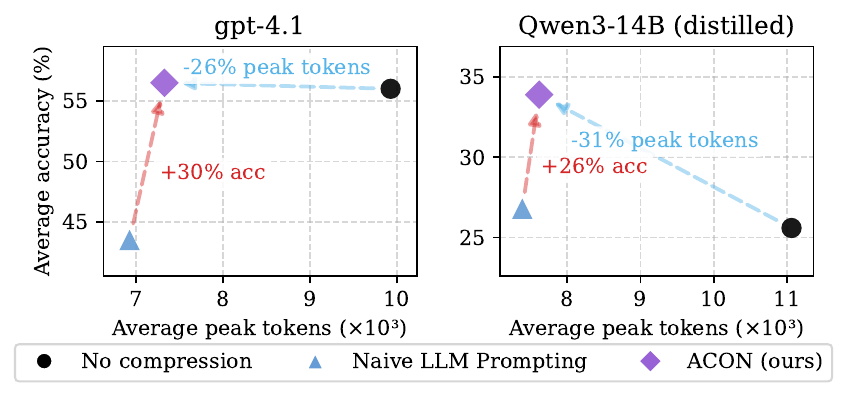}
    \vspace{-0.2in}
    \caption{\textbf{Accuracy-Peak tokens trade-off} on AppWorld benchmark~\citep{appworld}. We compare average accuracy versus peak tokens in history compression. \ours (ours) reduces token cost while preserving accuracy for the large model (gpt-4.1) relative to a naive prompting baseline, and even improves accuracy on smaller model (Qwen-14B). More results are in~\autoref{sec:exp}.}
    \vspace{-0.25in}
    \label{fig:first}
\end{figure}

The challenge of managing this context is particularly critical in productivity scenarios, such as email management or workflow automation, where agents must coordinate across heterogeneous tools~\citep{appworld, OfficeBench}.
Unlike simpler conversational tasks, these environments demand the preservation of diverse signal types: factual history, action-outcome relationships, success preconditions, and future decision cues.
Naive strategies like token truncation or generic summarization are insufficient, as they easily discard these critical details essential for multi-step reasoning. Consequently, effective compression must balance aggressive reduction with the precise retention of task-relevant state information.

Existing compression approaches, however, fail to fully address these agent-specific needs. Dialogue-oriented systems~\citep{packer2023memgpt} focus on conversational coherence rather than state tracking, while document-centric methods~\citep{LongLLMLingua} assume single-step reasoning where context can be discarded after use. While recent agent-focused methods attempt to bridge this gap, they face significant limitations: heuristic-based approaches~\citep{Mind2Act, openhands-condenser} are often brittle and narrowly specialized, limiting their robustness. Meanwhile, model optimization-based approaches~\citep{MEM1, Contextfolding} typically entangle compression with the agent model, making them difficult to apply directly to proprietary, API-based LLMs, where gradient-based updates to the underlying model are infeasible.

To address these challenges, we introduce \textbf{Agent Context Optimization (\ours)}, a unified framework for optimizing the compression of both environment observations and interaction histories.
Distinct from heuristic approaches that rely on static rules and model optimization-based methods that require model parameter updates, \ours introduces a \textbf{compression guideline optimization} directly in natural language space.
This method refines compressor prompts through failure analysis ensuring that critical environment-specific signals are preserved after compression without altering the agent model weight~\citep{APO, OPRO, TextGrad, blueprint}.
It makes \ours purely model-agnostic and directly applicable to agents based on proprietary, API-based LLMs.

\ours yields three key advantages over previous works.
First, the guideline optimization enables environment-specific compression rules to be derived consistently across diverse agentic tasks, overcoming the brittleness of handcrafted heuristics.
Second, by retaining essential information, optimally compressed contexts not only reduce memory costs but also improve decision quality, allowing smaller models to act more effectively by mitigating distraction.
Third, we validate that these optimized compressors can be distilled into smaller models, demonstrating that the compression module itself can be deployed with minimal computational overhead.

We validate \ours on three multi-step agent benchmarks: AppWorld~\citep{appworld}, OfficeBench~\citep{OfficeBench}, and Multi-objective QA~\citep{NQ, MEM1}, each requiring 15+ interaction steps. Our empirical results demonstrate clear advantages of \ours:
(1) lowers peak token usage of agents by 26--54\% while improving task success compared to existing compression baselines
(2) enables effective distillation of the context compressor into smaller models, preserving 95\% of the teacher’s accuracy, thereby reducing the overhead of the compression 
(3) allows small LMs to function more effectively as agents, improving performance by 32\% on AppWorld, 20\% on OfficeBench, and 46\% on Multi-objective QA by mitigating the distraction of long contexts.
Our result highlights on AppWorld benchmark are in~\autoref{fig:first}.

In summary, our work makes the following contributions:
\begin{itemize}[itemsep=0.7mm, parsep=1pt, leftmargin=*]
\vspace{-0.1in}
\item We propose \textbf{Agent Context Optimization} (\ours), a framework for optimizing compression of both environment observations and interaction histories, tailored to multi-step, long-horizon agentic tasks.
\item We develop a failure-driven compression guideline optimization. This approach is model-agnostic, making it readily applicable to any LLM, including proprietary API-based models, without requiring weight updates.
\item We enable cost-efficient deployment of optimized compressors by distilling them into smaller models, preserving over \textbf{95\%} of the teacher's performance while reducing the overhead of compression.
\item We validate \ours on AppWorld, OfficeBench, and Multi-objective QA, showing that it reduces peak token usage by \textbf{26--54\%} while improving task success over existing compression baselines with LLMs, and enabling small LMs to achieve \textbf{20--46\%} performance improvements.
\end{itemize}
\section{Related Works}

\paragraph{Long-horizon LLM agents.}
Large language model (LLM) agents extend pretrained models beyond static single-step reasoning tasks (e.g., RAG-based QA, math problem solving, or code generation) to interactive decision-making in dynamic environments~\citep{ReAct, CodeAct, Kimi-K2, ChatGPTAgent}.
Unlike chatbots or solvers that return an answer in one pass, agents must iteratively observe their surroundings, select tools, and execute actions while revising their plans based on feedback~\citep{ALFWorld, SWE-bench, WebArena, BrowserComp, OSWorld, WindowsAgentArena}. 
Recent work highlights the importance of \emph{long-horizon LLM agents}, which tackle tasks that unfold over dozens to hundreds of steps and require coordination across multiple applications and tools~\citep{METR, appworld, OfficeBench, OdysseyBench}. 
A central challenge in these scenarios lies in managing the \emph{dynamic long context}, where the agent must retain multi-step interaction histories and handle diverse observations produced by heterogeneous environments.

\paragraph{Context compression for LLMs.}
Managing this ever-growing context has been a longstanding challenge, and a variety of approaches have been proposed to compress LLM inputs.
Prior works on context compression can be broadly grouped into three directions: document- or retrieval-based compression~\citep{efficient-long-context-lm, selective-context, recomp, CompAct, MEM1, LongLLMLingua, TACO-RL}, dialogue memory summarization~\citep{A-Mem, maharana2024evaluating, wang2025recursively}, and low-level KV cache compression~\citep{LightThinker}. While each line of research has demonstrated benefits in its respective setting, they remain insufficient for the dynamic and heterogeneous contexts required by long-horizon agents, where the relevance of information frequently shifts as the agent progresses.

Beyond general compression, several recent works have explored context compression specifically for LLM agents~\citep{Mind2Act, ContextualizeWeb, openhands-condenser, MemAgent}. However, these approaches either rely on naive prompting or target narrow domains, limiting their broader applicability.
Another related line of works treats context compression as an agent action~\citep{MEM1, Agentfold, Contextfolding}, employing reinforcement learning to optimize the model for both compression and agent action policy.
However, such methods inherently update the model to couple reasoning with compression, typically requiring access to internal weights.
In addition, ReSum~\citep{ReSum} shares the motivation of extending long-horizon agents through summarized contexts, but it optimizes the policy model to better utilize summaries.

\textbf{In contrast,} we introduce a universal optimization framework for agent context compression that is applicable to any arbitrary LLMs. Our framework distinguishes itself by supporting both history and observation compression and providing a generalizable optimization process for the compression. Since our approach is entirely \textbf{model-agnostic}, it remains equally effective for both open-source models and proprietary API-based LLMs.
A detailed analysis is provided in \autoref{appendix:sec:results}.
\section{Agent Context Optimization (\ours)}
We present Agent Context Optimization (\ours), a unified framework for optimized history and observation compression in long-horizon LLM agents.
We begin by formulating the agent task and defining context cost in~\autoref{sec:problem_formulation}.
Next, in~\autoref{sec:context_opt_prompting}, we introduce generative compression with LLMs for both history and observation, and formalize the associated optimization objective and its challenges.
We then propose our optimization method in~\autoref{sec:prompt_opt}, followed by a distillation that enables smaller models for compressions to reduce the compression cost (\autoref{sec:distillation}).

\subsection{Problem Formulation}
\label{sec:problem_formulation}
\paragraph{Task.}
An agentic task is formulated as a Partially Observable Markov Decision Process (POMDP) $\gE = \left\langle \gS, \gA, \gO, \gT, \gR\right\rangle$ with state space $\gS$, action space $\gA$, and observation space $\gO$.
The transition function $\gT(s, a) \rightarrow (s', o)$ is deterministic, and it is determined by the environment. Specifically, it executes an action $a \in \gA$ in the environment and returns the next state and observation.
The reward function $\gR$ returns the reward given the terminal state $s_T$.
The terminal state is arrived when the transition function receives the special action (e.g., $\texttt{finish\_task}$).

An LLM agent interacts with the environment to get information for making a decision to achieve a given task $o_0$ through multiple steps.
For each step $t$, the LLM $\gM$ generates the action $a_t$ followed by its reasoning at each step~\citep{ReAct, CodeAct} given the interaction history $\vh_{t-1} = (o_0, a_0, o_1, a_1, \cdots, o_{t-1}, a_{t-1})$ and the latest observation $o_t$: 
\begin{equation}
\label{eq:agent}
\vspace{-0.05in}
    \gM(o_t, \vh_{t-1}; \theta, \gP_{\sf agent}) \mapsto a_t,
\vspace{-0.01in}
\end{equation}
where $\theta$ refers to the pre-trained parameters of the LLM and $\gP_{\sf agent}$ is the prompt that consists of a general environment description, tools, output format, and few-shot examples in natural language.

\begin{figure*}
    \centering
    \includegraphics[width=1.0\linewidth]{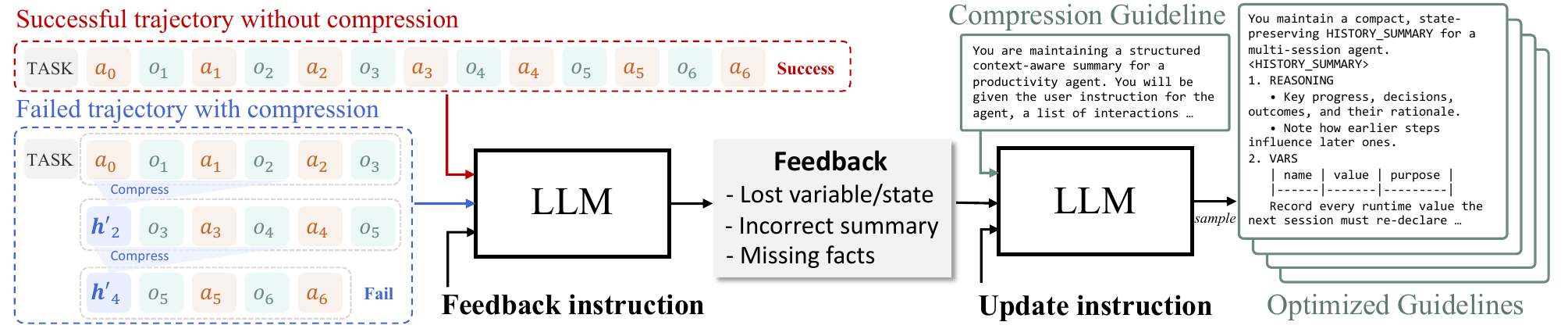}
    \vspace{-0.2in}
    \caption{\textbf{Compression Guideline Optimization.} Feedback is generated by contrasting successful trajectories (no compression) with failed ones (with compression). The collected feedback is then used by LLM to refine the compression guidelines.}
    \label{fig:prompt_opt}
    \vspace{-0.2in}
\end{figure*}

\paragraph{Cost function for context.}
We assume that the LLM agent's parameters $\theta$ and the task and system prompt $\gP_{\sf agent}$ are fixed.
We define a cost function $\gC$ that measures the cost of encoding the dynamic context during action generation at each step such as $\gO(n)$ computational cost of a transformer for decoding given $n$ input tokens.
The cost function takes the interaction history $\vh_{t-1}$, and the latest observation $o_t$ as input and returns the per-step cost:

\begin{equation}
\label{eqn:cost}
C(\mH) = \sum\nolimits_{t=1}^{T}\gC(\vh_{t-1}, o_t),
\end{equation}
where $C$ is the total cost of completing the task, $\mH = \{ \vh_{t-1}, o_t \}_{t=1}^T$ denotes the sequence of history and observation of each step.
Typically, $C$ is proportional to the summation of token lengths of action and observations in each step, $\vh_{t-1}$ and $o_t$.
While the cost of system prompt is static, the costs from interaction histories are \textbf{unbounded}, leaving the user with only two options: terminate the task early or truncate the context heuristically to a maximum length.
This raises the central question: \emph{how can we compress context more effectively than such heuristics?}

\subsection{History \& Observation Compression with LLMs}
\label{sec:context_opt_prompting}
To address this challenge, we use an LLM $f(\cdot; \phi, \gP)$, parameterized by pre-trained weights $\phi$ and a compression guideline $\gP$, to minimize context cost defined in~\autoref{eqn:cost} (e.g., \textit{summarize the given interaction history}). 
As in~\autoref{eq:agent}, the LLM receives two inputs at each step: the interaction history $\vh_{t-1}$ and the latest observation $o_t$.
This introduces two options for context compression:

\paragraph{History compression.}
The interaction history accumulates both environment observations and agent actions.
In long-horizon tasks, this history can grow excessively large. 
To manage its length, we apply history compression only when the length exceeds a predefined threshold $T_{\sf hist}$:
\begin{equation}
\label{eqn:history_comp}
\vh_t' =
f(\vh_t;\phi,\gP_{\sf hist}) \;\;\text{if } |\vh_t| > T_{\sf hist}, \quad
\vh_t \;\;\text{otherwise}.
\end{equation}
The compressed history $\vh_t'$ replaces the raw history in~\autoref{eq:agent}.
This selective compression ensures that the overhead of invoking the compressor is incurred only when necessary~\citep{openhands-condenser}.

\paragraph{Latest observation compression.} 
Given an action $a$, the environment returns an observation $o$ according to the transition function $\mathcal{T}(s, a) \rightarrow (s', o)$. 
We similarly apply observation compression only when the observation length exceeds a threshold $T_{\sf obs}$:
\begin{equation}
\label{eqn:obs_comp}
o_t' =
f(o_t, \vh_{t-1}; \phi, \gP_{\sf obs}) \;\;\text{if } |o_t| > T_{\sf obs}, \quad
o_t \;\;\text{otherwise}.
\end{equation}
This mechanism avoids unnecessary overhead when $o_t$ is already short, while still reducing redundant or distracting content in long observations~\citep{recomp, Mind2Act, ContextualizeWeb}. 
The compressed one $o_t'$ replaces the raw one in~\autoref{eq:agent} and is stored in the interaction history $\vh$.

In both cases, the compressor LLM selects information to preserve based on its learned prior knowledge of importance. 
However, there is \textbf{no guarantee} that the salient details required for successful task completion are retained. 
The agent context effectively serves as a \textbf{world model of the environment}, encompassing diverse forms of information such as  causal relations (\eg email leaves drafts), evolving states (\eg account balance), preconditions (\eg login required), and task-relevant decision cues (\eg due dates).
Effective context compression must therefore accommodate this heterogeneous and dynamic nature of agent context, ensuring that the most critical signals are preserved for long-horizon reasoning and task success.

\paragraph{Optimization objective.}
We optimize the compressor parameters \(\psi \triangleq (\phi,\gP)\) to maximize task reward while minimizing context cost.
At each step \(t\), the compressor produces either a compressed history $\vh_{t}' = f_{\sf hist}(\vh_t;\psi)$ or observation $o_t' = f_{\sf obs}(o_t,\vh_{t-1};\psi)$.
Let the compressed context be
\begin{gather}
\label{eq:compressed-context}
\mH'(\psi) \;=\; \{\, \vh_{t-1}',\, o_t'\,\}_{t=1}^{T},\\
C(\mH'(\psi)) \;=\; \sum\nolimits_{t=1}^{T} \gC(\vh_{t-1}', o_t').
\end{gather}
With the agent \(\gM(\cdot;\theta,\gP_{\sf agent})\) fixed, the environment induces a trajectory \(\tau(\psi)\) and terminal state \(s_T(\psi)\) when the agent conditions on \(\mH'(\psi)\). 
Our learning objective is
\begin{equation}
\label{eq:optim}
\max_{\psi}\; 
\underbrace{\mathbb{E}\!\left[\,\gR\!\big(s_T(\psi)\big)\,\right]}_{\text{maximize}}
\;-\;
\lambda\,
\underbrace{\mathbb{E}\!\left[\,C\!\big(\mH'(\psi)\big)\,\right]}_{\text{minimize}},
\quad \lambda \ge 0,
\end{equation}
where $\lambda$ is a multiplier and the expectations are over tasks. 

\paragraph{Challenges.}
The optimization objective in~\autoref{eq:optim} is difficult to optimize in practice because there is no gold supervision for compression, the reward is sparse and only revealed at the end of the trajectory, and the context cost is defined over discrete quantities, which precludes direct gradient computation. 
While these properties naturally motivate reinforcement learning (RL)~\citep{RL}, applying RL introduces additional obstacles: 
(1) updating the parameters $\phi$ of a LLM with RL can be computationally prohibitive, 
(2) environment rollouts are extremely expensive since each reward requires multi-step executions of both agent and compressor, and 
(3) policy gradient estimates suffer from high variance since compression quality is only indirectly evaluated through eventual task success.

\subsection{Optimizing Compression Guidelines}
\label{sec:prompt_opt}
To overcome these challenges, we propose to optimize \textbf{compression guidelines $\gP$} (natural language prompts) for context compression, rather than fine-tuning model parameters $\phi$.
Trajectories under compressed contexts provide \emph{dense signals} about the quality of compression. 
For example, if the agent fails with compressed context while succeeding without compression, this indicates that the compressed context may have lost crucial information. 
Such trajectory-level comparisons yield richer feedback than scalar rewards (e.g., binary task success).

We instantiate this idea as prompt optimization using an LLM as the optimizer, where the natural language prompt $\gP$ is refined via feedback expressed in natural language~\citep{OPRO, TextGrad, DSPy, MetaSPO}.
We introduce \textbf{compression guideline optimization} based on \emph{contrastive task feedback}.

On the training set $\gD_{\sf train}$, we run the LLM agent both without and with context compression to obtain baseline context $\mH$ and compressed context $\mH'$. 
We collect tasks where the agent succeeds with $\mH$ but fails with $\mH'$, forming a contrastive subset $\gD_{\sf cont}$. 
For each task in $\gD_{\sf cont}$, we query a optimizer LLM with the context before and after compression to obtain natural language feedback:
\begin{gather}
\text{Feedback} = \text{LLM}(\text{Feedback Instruction}, \mH, \mH').
\end{gather}
This feedback serves as a natural language gradient~\citep{TextGrad}, indicating how the compression guideline $\gP$ should be refined. 
We then aggregate feedback from multiple trajectories:
\begin{gather}
\label{eqn:prompt_opt}
\gP^{(1)} = \text{LLM}(\text{Update Inst.}, \gP^{(0)}, \textstyle \Vert_{i=1}^n \text{Feedback}_i),
\end{gather}
where $\Vert$ is concatenation of feedbacks from each task, which corresponds to a batch optimization step in textual gradient descent~\citep{TextGrad}.
We also generate multiple candidate prompts $\{ \gP_k^{(1)} \}$, evaluate them on $\gD_{\sf cont}$, and select the best-performing one.
We refer this process as \emph{utility maximization} step \UTIL as it primarily maximizes the first term (task reward) of~\autoref{eq:optim}.

However, optimizing only for reward may neglect the context cost (second term in~\autoref{eq:optim}).
To address this, motivated by alternating optimization, we perform a second iteration that conditions only on successful task with compressed context, asking the LLM to generate feedback about which information was actually used during execution. 
This refines $\gP^{(1)} \rightarrow \gP^{(2)}$, encouraging shorter yet sufficient contexts.
We refer this additional process as \emph{compression maximization} step \COMP as it minimizes the second term (context cost) of~\autoref{eq:optim}.

We illustrate overall process in~\autoref{fig:prompt_opt}. Algorithm~\ref{alg:prompt_opt} and prompts are in~\autoref{sec:appendix:exp_setup}.

\begin{table*}[t]
\centering
\caption{Results across different difficulty levels on \textbf{Appworld} benchmark (test-normal). Each block reports accuracy (task goal completion score), steps, peak input tokens ($10^3$), and dependency ($10^6$) for agents. Best results in each column are highlighted in bold. Rows in blue background indicate the results from \colorbox{lightblue}{\textbf{ours}}.
\ours consistently improves accuracy while reducing peak tokens and dependency, with \ours~\UTIL \COMP achieving the best overall performance.}
\vspace{-0.0in}
\label{tab:1_appworld_main}
\resizebox{\textwidth}{!}{
\begin{tabular}{lccccccccccccc}
\toprule
\multirow{2}{*}{Method} & 
\multicolumn{4}{c}{Average (168)} & 
\multicolumn{3}{c}{Easy (57)} & 
\multicolumn{3}{c}{Medium (48)} & 
\multicolumn{3}{c}{Hard (63)} \\
\cmidrule(lr){2-5} \cmidrule(lr){6-8} \cmidrule(lr){9-11} \cmidrule(lr){12-14}
& Acc. $\uparrow$ & Steps $\downarrow$ & Peak $\downarrow$ & Dep.$\downarrow$
& Acc. $\uparrow$ & Peak $\downarrow$ & Dep.$\downarrow$
& Acc. $\uparrow$ & Peak $\downarrow$ & Dep.$\downarrow$
& Acc. $\uparrow$ & Peak $\downarrow$ & Dep.$\downarrow$ \\
\midrule
\rowcolor{lightgray}\multicolumn{14}{c}{\textbf{Agent:} \texttt{gpt-4.1} / \textbf{Compressor:} \texttt{gpt-4.1}}\\
\midrule
No compression & 56.0 & \bf 16.14 & 9.93 & 5.96 & 80.7 & 7.57 & 2.98 & 47.9 & 10.10 & 5.36 & \bf 39.7 & 11.95 & 9.11 \\
\midrule[0.1pt]
\multicolumn{14}{l}{\textbf{History Compression}}\\
FIFO & 45.8 & 28.48 & 6.73 & 5.69 & 84.2 & 5.85 & 2.89 & 39.6 & 7.26 & 6.24 & 15.9 & \bf 7.14 & 7.80 \\
Retrieval & 27.4 & 33.17 & 8.39 & 6.68 & 61.4 & 7.40 & 3.97 & 12.5 & 8.74 & 7.72 & 7.9 & 9.02 & 8.33 \\
LLMLingua & 39.3 & 24.42 & 7.50 & 6.37 & 66.7 & 6.38 & 3.04 & 37.5 & 8.04 & 7.39 & 15.9 & 8.09 & 8.59 \\
Prompting & 43.5 & 24.01 & 6.93 & 5.29 & 66.7 & 6.36 & 2.84 & 41.7 & 7.10 & 5.36 & 23.8 & 7.31 & 7.48 \\
\rowcolor{lightblue} \ours ~\UTIL & 51.2 & 20.92 & 7.17 & 4.49 & 77.2 & 6.45 & 2.43 & 50.0 & 7.39 & 4.47 & 28.6 & 7.65 & \bf 6.37 \\
\rowcolor{lightblue} \ours ~\UTIL \COMP & \textbf{56.5} & 22.82 & 7.33 & 4.69 & \textbf{86.0} & 7.09 & 2.84 & \textbf{56.2} & 7.48 & 4.43 & 30.2 & 7.44 & 6.55 \\
\midrule[0.1pt]
\multicolumn{14}{l}{\textbf{Observation Compression}}\\
LLMLingua & 32.1 & 18.16 & 8.17 & 6.01 & 54.4 & 5.78 & 2.33 & 29.2 & 8.24 & 5.23 & 14.3 & 10.29 & 9.92 \\
Prompting & 42.3 & 17.38 & \bf 6.58 & \textbf{4.09} & 64.9 & \textbf{4.92} & \textbf{1.88} & 35.4 & \bf 6.96 & \bf 4.11 & 27.0 & 7.79 & 6.07 \\
\rowcolor{lightblue} \ours ~\UTIL & 47.0 & 16.67 & 7.62 & 5.08 & 70.2 & 5.87 & 2.21 & 45.8 & 7.79 & 5.00 & 27.0 & 9.07 & 7.73 \\
\rowcolor{lightblue} \ours ~\UTIL \COMP & 53.6 & 18.12 & 7.43 & 4.93 & 82.5 & 5.66 & 2.63 & 47.9 & 7.30 & 4.43 & 31.8 & 9.14 & 7.50 \\
\bottomrule
\end{tabular}
}
\vspace{-0.0in}
\end{table*}

\subsection{Distilling Context Compression into Small Models}
\label{sec:distillation}
While compression guideline optimization enables effective compression, repeatedly invoking the large LLM for compression adds substantial overhead.
To reduce this cost, we \textbf{distill the compressor into a smaller model}. 
The teacher with optimized guideline $\gP^*$ (parameters $\phi_{\sf T}$) generates compressed outputs $\vy$ from input $\vx$, which supervise the student (parameters $\phi_{\sf S}$). 
We train the student with a cross-entropy objective~\citep{SeqKD} with input-output pair $(\vx,\vy)$, 
where $(\vx,\vy) = (\vh_t, \vh_t')$ for~\autoref{eqn:history_comp} or $(\vx,\vy) = ((\vh_{t-1}, o_t), o_t')$ for~\autoref{eqn:obs_comp}:
\begin{equation}
\min_{\phi_{\sf S}}\;
\mathbb{E}_{(\vx,\vy) \sim \mathcal{D}^{+}_{\sf train}}
\Bigg[
-\sum_{n=1}^{L_\vy} \log f(\vy_n \mid \vx, \vy_{<n}; \phi_{\sf S}, \gP^*)
\Bigg],
\end{equation}
where $\mathcal{D}^{+}_{\sf train}$ denotes tasks where the teacher succeeds with compressed context. 

Once trained, the student replaces the teacher during inference, decoupling decision making from compression. 
This two-stage pipeline, guideline optimization then distillation, achieves effective compression with a much smaller model ($|\phi_{\sf T}| \gg |\phi_{\sf S}|$):
\begin{equation}
f(\cdot; \phi_{\sf T}, \gP)
\xrightarrow{\;\text{prompt optim.}\;}
f(\cdot; \phi_{\sf T}, \gP^*)
\xrightarrow{\;\text{distill}\;}
f(\cdot; \phi_{\sf S}, \gP^*).
\end{equation}

\section{Experiments}
\label{sec:exp}

We evaluate \ours on three challenging benchmarks that require multi-step interactions across diverse domains. 
Our experiments are designed to address the following key questions:
\begin{itemize}[itemsep=0.5mm, parsep=1pt, leftmargin=*]
\vspace{-0.1in}
    \item How well does \ours improve token efficiency while preserving performance? (\autoref{sec:exp:main})
    \item Does distilling the compressor reduce its size while maintaining agent performance? (\autoref{sec:exp:compressor_distillation})
    \item Can \ours help small, distilled LM agents perform better under long contexts? (\autoref{sec:exp:agent_distillation})
\vspace{-0.1in}
\end{itemize}

\begin{table*}[ht]
\centering
\caption{Results on \textbf{OfficeBench} and \textbf{8-objective QA} benchmarks. We report performance metrics (acc/EM/F1) along with steps, peak input tokens ($10^3$), and dependency ($10^6$). Best values are in \textbf{bold}. Rows in blue are \colorbox{lightblue}{\textbf{ours}}. \ours consistently improves accuracy/efficiency trade-offs.}
\vspace{-0.0in}
\label{tab:qa_officebench_main}
\begin{subtable}[t]{0.448\textwidth}
\centering
\caption{OfficeBench}
\vspace{-0.0in}
\label{tab:4_officebench_main}
\resizebox{1.0\textwidth}{!}{
\setlength{\tabcolsep}{6pt}
\begin{tabular}{lcccc}
\toprule
Method & Acc. $\uparrow$ & Steps $\downarrow$ & Peak $\downarrow$ & Dep. $\downarrow$ \\
\midrule
\rowcolor{lightgray} \multicolumn{5}{c}{\textbf{Agent:} \texttt{gpt-4.1} / \textbf{Compressor:} \texttt{gpt-4.1}} \\
\midrule
No Compression                     & \bf 76.84 & 11.52 & 7.27 & 4.43 \\
\midrule
\multicolumn{5}{l}{\textbf{History Compression}} \\
FIFO                  & 67.37 & 12.26 & {\bf 4.02} & 2.64 \\
Retrieval             & 65.26 & 16.20 & 4.33 & 2.06 \\
LLMLingua             & 70.53 & 10.89 & 4.65 & 1.85 \\
Prompting             & 71.58 & {\bf 10.13} & 4.40 & {\bf 1.10} \\
\rowcolor{lightblue} \ours~\UTIL& 74.74 & 13.13 & 4.93 & 3.85 \\
\rowcolor{lightblue} \ours~\UTIL\COMP & 72.63 & 11.54 & 4.54 & 1.91 \\
\midrule
\multicolumn{5}{l}{\textbf{Observation Compression}} \\
LLMLingua             & 71.58 & 11.89 & 7.38 & 6.14 \\
Prompting             & 55.79 & 12.24 & 6.44 & 2.68 \\
\rowcolor{lightblue} \ours~\UTIL & 73.68 & 10.83 & 6.55 & 3.85 \\
\rowcolor{lightblue} \ours~\UTIL\COMP & 72.63 & 10.28 & 6.17 & 2.88 \\
\bottomrule
\end{tabular}
}
\end{subtable}\hfill
\begin{subtable}[t]{0.522\textwidth}
\centering
\caption{8-objective QA}
\vspace{-0.0in}
\label{tab:3_qa_main}
\resizebox{\textwidth}{!}{
\setlength{\tabcolsep}{7pt}
\begin{tabular}{lccccc}
\toprule
Method & EM $\uparrow$ & F1 $\uparrow$ & Steps $\downarrow$ & Peak $\downarrow$ & Dep. $\downarrow$ \\
\midrule
\rowcolor{lightgray}  \multicolumn{6}{c}{\textbf{Agent:} \texttt{gpt-4.1} / \textbf{Compressor:} \texttt{gpt-4.1}} \\
\midrule
No compression                 & 0.366 & 0.488 & 15.78 & 10.35 & 3.32 \\
\midrule
\multicolumn{6}{l}{\bf History Compression} \\
FIFO                           & 0.293 & 0.388 & 19.26 & 5.09 & 2.51 \\
Retrieval                      & 0.331 & 0.438 & 20.06 & 5.11 & 2.62 \\
LLMLingua                      & 0.363 & 0.481 & 17.68 & 5.68 & 2.24 \\
Prompting                      & \bf 0.376 & 0.478 & 18.70 & 4.73 & 1.66 \\
\rowcolor{lightblue} \ours~\UTIL      & 0.373 & \bf 0.494 & 17.14 & 4.71 & 1.57 \\
\rowcolor{lightblue} \ours~\UTIL\COMP & 0.335 & 0.458 & 17.79 & \bf 4.65 & \bf 1.50 \\
\midrule
\multicolumn{6}{l}{\bf Observation Compression} \\
LLMLingua                      & 0.320 & 0.414 & 14.23 & 5.16 & 1.35 \\
Prompting                      & 0.288 & 0.397 & \bf 11.64 & \bf 3.41 & \bf 0.45 \\
\rowcolor{lightblue} \ours~\UTIL      & 0.364 & 0.475 & 16.33 & 4.97 & 1.28 \\
\rowcolor{lightblue} \ours~\UTIL\COMP & 0.336 & 0.461 & 14.00 & 4.22 & 0.81 \\
\bottomrule
\end{tabular}
}
\end{subtable}
\vspace{-0.0in}
\end{table*}

\begin{figure*}
    \centering
    \includegraphics[width=0.93\linewidth]{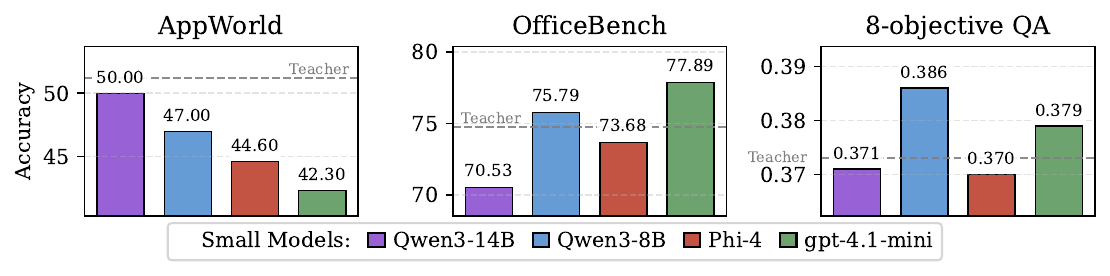}
    \vspace{-0.0in}
      \caption{\textbf{Results of distilled compressors on history compression} with gpt-4.1 as the agent. Grey dotted lines denote performance using the gpt-4.1 teacher compressor. Student models (Qwen3-14B, Qwen3-8B, Phi-4) are distilled from gpt-4.1 compressor using the optimized compression guideline after~\UTIL step, and evaluated across all benchmarks. We also include gpt-4.1-mini without distillation, showing that even a small model can serve as an effective compressor without additional training.}
    \label{fig:exp:history_distill}
\vspace{-0.1in}
\end{figure*}

\subsection{Experimental Setup}
\paragraph{Benchmarks \& Metrics.}
We focus on long-horizon agentic task benchmarks that require $10+$ interaction steps on average:
\textbf{(1) AppWorld}~\citep{appworld}: Main benchmark with 9 simulated apps (e.g., Venmo, Spotify, SimpleNote) and $\sim$100 simulated users. Performance is measured by task completion score. 
\textbf{(2) OfficeBench}~\citep{OfficeBench}: Productivity tasks across 6 apps (e.g., Word, Excel, Email), operating on simulated documents. Performance is measured by benchmark-defined accuracy functions.
\textbf{(3) 8-objective QA}~\citep{NQ, MEM1}: QA benchmark where agents interact with a search tool to answer 8 questions and output a consolidated answer set. Performance is the average of Exact Match (EM) and F1 scores across 8 questions.

In addition to task-specific performance metrics, we report three token efficiency metrics following prior work~\citep{LightThinker, MEM1}:
\textbf{(1) Steps:} The average number of interaction steps per task.
\textbf{(2) Peak Tokens:} The maximum context length encountered across all steps. 
\textbf{(3) Dependency:} The cumulative dependency of each generated action on prior tokens, measuring how much generation relies on the context history. 
Full details are provided in the~\autoref{sec:appendix:exp_setup}.

Throughout all experiments, we use ReAct agent~\citep{ReAct}. For the detailed tool format, we follow the convention of each benchmark.

\paragraph{Baselines.}
\textbf{(1) No Compression:} full uncompressed context.  
\textbf{(2) FIFO:} keep the most recent $k$ interactions, discarding earlier ones~\citep{SWE-agent}.  
\textbf{(3) Retrieval:} select $k$ past interactions most similar to the current query via embedding search~\citep{A-Mem}.  
\textbf{(4) LLMLingua:} extractive compression with an encoder-only LM~\citep{LLMLingua, LLMLingua2}.  
\textbf{(5) Prompting:} naive baseline using a general compression instruction~\citep{openhands-condenser, ContextualizeWeb}.
\paragraph{Our Methods.}
We evaluate two versions of \ours. 
\textbf{(1) \ours~\UTIL} utilizes an \emph{optimized guideline} for context compression after utility maximization step.
\textbf{(2) \ours~\UTIL\COMP} applies compression maximization~\COMP after utility maximization~\UTIL, aiming for shorter but informative compression. We apply only single step per each step for experiments.
Additional analysis on the number of steps is in~\autoref{appendix:sec:results}.

\begin{figure*}[ht]
    \centering
    \includegraphics[width=0.93\linewidth]{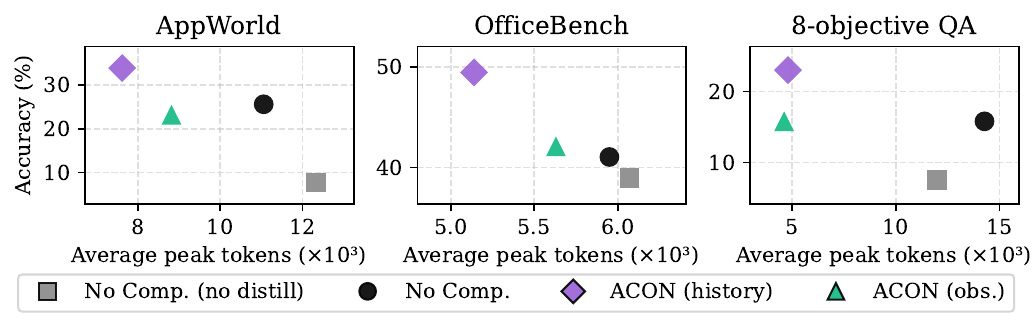}
    \vspace{-0.1in}
    \caption{\textbf{Performance-efficiency trade-off of the Qwen3-14B agent} distilled from gpt-4.1 trajectories. For distilled compressors, we use the same distillation setting as in~\autoref{fig:exp:history_distill}. Compared to the baseline without compression, our framework \ours provides compressed trajectories combined with a distilled compressor, enabling the distilled agent to achieve consistently higher accuracy while requiring substantially fewer peak input tokens across all benchmarks.}
    \label{fig:exp:distilled_agent}
\vspace{-0.1in}
\end{figure*}

\begin{figure*}[ht]
    \centering
    \begin{subfigure}[b]{0.32\textwidth} 
        \centering
        \includegraphics[width=\textwidth]{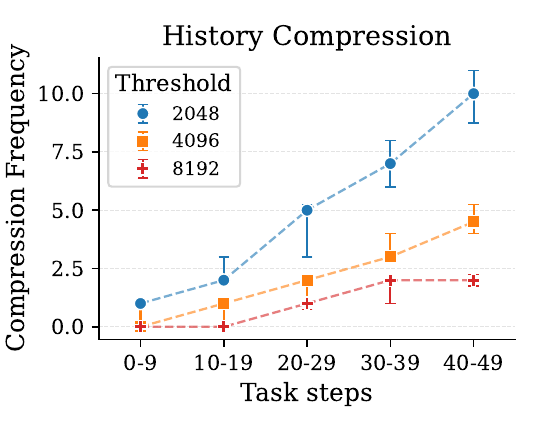}
        \label{fig:sub1}
    \end{subfigure}
    \hfill 
    \begin{subfigure}[b]{0.32\textwidth} 
        \centering
        \includegraphics[width=\textwidth]{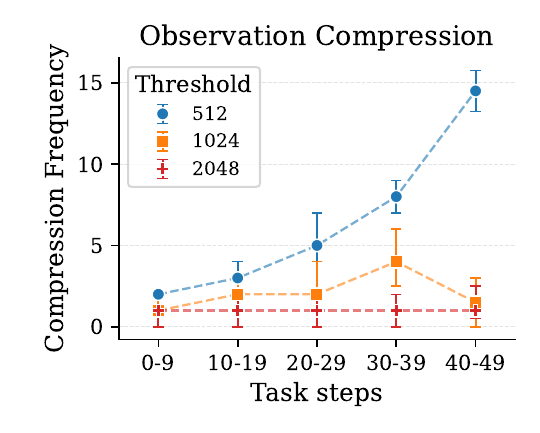}
        \label{fig:sub2}
    \end{subfigure}
    \hfill 
    \begin{subfigure}[b]{0.32\textwidth} 
        \centering
        \includegraphics[width=\textwidth]{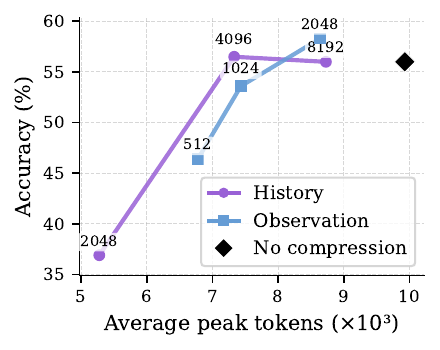}
        \label{fig:sub3}
    \end{subfigure}
    \vspace{-0.3in}
    \caption{\textbf{Ablation studies on thresholds for compression} on AppWorld with gpt-4.1. (1) the number of compressions (compression frequency) for each length of task trajectories (task steps). (2) the performance comparison for each threshold setting.}
    \label{fig:exp:threshold_ablation}
\vspace{-0.15in}
\end{figure*}

\subsection{Overall performance and token efficiency}
\label{sec:exp:main}
In~\autoref{tab:1_appworld_main} and~\autoref{tab:qa_officebench_main}, we first evaluate \ours using on gpt-4.1~\citep{GPT-4.1} for both agent and compressor, which already achieves strong results on benchmarks. 

For history compression, as shown in~\autoref{tab:1_appworld_main}, on AppWorld, \textbf{\ours reduces peak tokens by over 25\%} while preserving the accuracy of the no compression upper bound, outperforming all baselines that suffer severe degradation on medium and hard tasks spanning longer steps.  
On OfficeBench (\autoref{tab:4_officebench_main}), \ours lowers peak context size by nearly 30\% while maintaining accuracy above 74\%.  
On 8-objective QA (\autoref{tab:3_qa_main}), \ours even surpasses the no compression baseline in EM/F1 while reducing peak tokens and dependency by 54.5\% and 61.5\%, respectively.
For observation compression, \ours consistently outperforms all baselines confirming that compression guideline optimization is effective for compressing not only history but also raw observations.

Applying only the utility maximization step (\UTIL) improves performance while reducing token cost across all benchmarks, whereas the compression maximization step (\COMP) further lowers token cost but may slightly hurt accuracy depending on the environment.
This trade-off suggests a practical guideline for choosing between the two variants.
In verbose and noisy environments such as AppWorld, where observations contain redundant API outputs and distractors, \UTIL\COMP is often preferable because aggressive pruning can improve both efficiency and focus.
In contrast, in high-fidelity information-seeking tasks such as OfficeBench and 8-objective QA, \UTIL is generally safer because over-compression may remove subtle facts needed for final decisions.
Therefore, \UTIL\COMP is recommended when observations contain substantial redundancy, while \UTIL is preferable when precise fact retention is critical.

\subsection{Compressor distillation}
\label{sec:exp:compressor_distillation}
We distill the compressor with optimized guidelines after~\UTIL step into smaller models such as Qwen3-14B, Qwen3-8B~\citep{Qwen3}, and Phi-4~\citep{Phi4} using LoRA~\citep{lora}.
As shown in~\autoref{fig:exp:history_distill}, the \textbf{distilled compressors retain over 95\% of the performance of the gpt-4.1 teacher} (indicated by the grey dotted line) while reducing computational overhead.
We also observe that gpt-4.1-mini, even without any distillation, can serve as an effective lightweight compressor on OfficeBench and QA.
This indicates that small models can reliably replace large LLM-based compressors when equipped with optimized guidelines.
These results confirm that small models are sufficient for compression, enabling the expensive LLM to be only reserved for the agent.

\subsection{\ours for distilled small agents}
\label{sec:exp:agent_distillation}
We examine whether \ours also benefits smaller LLM agents, which are particularly vulnerable to long-horizon inefficiency. Without compression, models such as Qwen3-14B often fail on medium and hard tasks due to distracting context. 
As shown in~\autoref{fig:exp:distilled_agent}, \ours substantially improves their performance: on AppWorld, Qwen3-14B achieves a \textbf{32.4\% relative improvement} (from 25.6\% to 33.9\%), and on 8-objective QA, it shows a \textbf{45.6\% gain} (from 0.158 to 0.23 EM). 
These results demonstrate that \ours acts as an equalizer, enabling smaller agents with concise but informative contexts to approach the performance of  larger models.

\subsection{Analysis}
\label{sec:analysis}

\paragraph{Compression threshold: moderate value yields the best trade-off.}
In~\autoref{fig:exp:threshold_ablation}, we provide ablations on threshold for compression in~\autoref{eqn:history_comp} and~\autoref{eqn:obs_comp}.
Results show that smaller thresholds reduce tokens but incur more frequent compression calls and degrade accuracy, while larger thresholds preserve accuracy with higher cost. 
Moderate values (4096 for history, 1024 for observation) provide the best trade-off, maintaining accuracy close to no compression while still reducing peak tokens substantially.

\paragraph{Prompt optimizer: \texttt{o3} + contrastive feedback works best.}
We analyze how the choice of optimizer and the use of contrastive feedback affect compression guideline quality. 
As shown in \autoref{tab:exp:prompt_opt_ablation}, the default \texttt{o3} with contrastive feedback yields the best performance, while removing contrastive feedback (only using failed trajectories) or switching to other models results in lower accuracy.
Although \texttt{o3} shows the best performance, we also demonstrate that the optimizer model can be replaced to weaker models such as gpt-4.1, showing it still yields sufficiently fine guideline compared to the baseline guideline.

\begin{table}
\centering
\caption{\textbf{Ablation studies on the prompt optimizer} in AppWorld, gpt-4.1 agent and history compressor. Default is o3 optimizer with task contrastive feedback.}
\vspace{-0.0in}
\label{tab:exp:prompt_opt_ablation}
\setlength{\tabcolsep}{6pt}
\resizebox{0.85\linewidth}{!}{
\begin{tabular}{ccc}
\toprule
Optimizer model & Task contrastive & Average Acc. \\
\midrule
\rowcolor{lightblue}\textbf{\texttt{o3}} & \cmark & \textbf{51.2} \\
\texttt{o3} & \xmark & 50.6 (-0.6) \\
\texttt{gpt-4.1} & \cmark & 47.6 (-3.6) \\
\texttt{gpt-5} & \cmark & 50.6 (-0.6) \\
\bottomrule
\end{tabular}}
\vspace{-0.1in}
\end{table}

\paragraph{Optimization cost: practical and lightweight.}
One might assume that leveraging a reasoning model like \texttt{o3} would be cost-prohibitive.
However, our guideline optimization step is remarkably lightweight, demanding \emph{less than \$2 per benchmark}.
This can be negligible compared to the expense of trajectory rollouts on the training dataset.
For instance, utilizing gpt-4.1 on AppWorld train set requires a total rollout cost of approximately \$20.
Furthermore, even this data collection cost is substantially lower than that of RL-based methods such as GRPO~\citep{GRPO}, which requires extensive rollout for advantage estimation.
Detailed API cost computation and breakdowns are provided in~\autoref{subsec:eval_metrics} and~\ref{appendix:subsec:detailed_results}.

\paragraph{API cost analysis of compressors across different models.}
While using a compressor reduces context length, it incurs additional computational costs beyond the agent. In~\autoref{sec:distillation}, we proposed a distillation strategy to mitigate this overhead. To quantify the cost efficiency, we analyze the expenses using API pricing as a proxy.
With gpt-4.1 as the baseline compressor, the cost remains significant at \$0.045 per example. Switching to gpt-4.1-mini reduces the compression cost to \$0.014, achieving a 69.2\% reduction. 
However, the most substantial gain is observed with distilled Qwen3-14B, where the cost decreases to \$0.0004. This represents a 99.1\% reduction compared to the teacher model, effectively minimizing the cost burden of context compression.

\paragraph{Practical efficiency trade-off.}
We further quantify the practical overhead of generative compression on AppWorld.
For API cost, we estimate the end-to-end cost of both the agent and compressor, excluding input caching.
For latency, we run the agent and compressor locally on a single A100 GPU rather than using API endpoints, which avoids confounding factors such as network overhead and server-side queueing.
As shown in~\autoref{tab:practical_efficiency}, ACON reduces end-to-end API cost by compressing the context of agent, but introduces additional wall-clock latency due to the extra compressor call.

\begin{table}[t]
\centering
\small
\caption{
\textbf{Practical efficiency on AppWorld.} API cost is estimated per task with a gpt-4.1 agent and Qwen3-14B compressor. Latency is measured as median wall-clock time per task with a Qwen3-14B agent and Qwen3-14B compressor.
}
\label{tab:practical_efficiency}
\setlength{\tabcolsep}{6pt}
\resizebox{0.85\linewidth}{!}{
\begin{tabular}{lcc}
\toprule
Method & API Cost $\downarrow$ & Latency $\downarrow$ \\
\midrule
No Compression & \$0.331 & 73.24s \\
\ours (history) & \$0.285 & 87.68s \\
\ours (observation) & \$0.272 & 101.92s \\
\bottomrule
\vspace{-0.2in}
\end{tabular}}
\end{table}

We include more experimental results including experiments with different agent models, additional ablation studies, case study, and qualitative examples of context compression in~\autoref{appendix:sec:results} and~\ref{appendix:sec:examples}.

\section{Conclusion}
We presented \textbf{Agent Context Optimization (\ours)}, a unified framework that systematically compresses both interaction histories and environment observations for long-horizon LLM agents. Unlike prior work that relies on naive prompting or narrow domains, \ours introduces compression guideline optimization in natural language space, enabling adaptive and model-agnostic compression.
Experiments on AppWorld, OfficeBench, and Multi-objective QA show that ACON reduces peak tokens by 26--54\% while improving task success over existing compression baselines, with small degradation relative to full-context baseline.
Beyond memory efficiency, we demonstrate that optimized compressors can be distilled into smaller models, substantially lowering overhead without sacrificing performance. 
Moreover, by supplying concise yet informative contexts, \ours allows small agents such as Qwen3-14B to approach the performance of much larger models.
Overall, our findings highlight that \ours lays a foundation for more general, memory-efficient, and deployable long-horizon LLM agents.

\paragraph{Limitations and Future Work.}
While ACON effectively reduces context costs, a few limitations remain. First, our empirical evaluation primarily focuses on GPT models due to resource constraints. Second, like many context management frameworks, the compression process itself introduces computational overhead and increased latency.
We include more detailed discussion on limitations and future work in~\autoref{sec:appendix:limitations}.

\section*{Impact Statement}

This paper presents work whose goal is to advance the field of Large Language Models, specifically by addressing the computational costs of long-horizon autonomous agents. By optimizing context compression and enabling effective distillation into smaller models, our work primarily contributes to making advanced agentic capabilities more resource-efficient and accessible. This has positive implications for reducing the computational footprint of AI systems and lowering barriers to entry for researchers with limited resources. We do not foresee specific negative societal consequences beyond the general considerations required when deploying autonomous systems.


\bibliography{reference}

@article{maharana2024evaluating,
  title={Evaluating very long-term conversational memory of llm agents},
  author={Maharana, Adyasha and Lee, Dong-Ho and Tulyakov, Sergey and Bansal, Mohit and Barbieri, Francesco and Fang, Yuwei},
  journal={arXiv preprint arXiv:2402.17753},
  year={2024}
}

@article{wang2025recursively,
  title={Recursively summarizing enables long-term dialogue memory in large language models},
  author={Wang, Qingyue and Fu, Yanhe and Cao, Yanan and Wang, Shuai and Tian, Zhiliang and Ding, Liang},
  journal={Neurocomputing},
  volume={639},
  pages={130193},
  year={2025},
  publisher={Elsevier}
}

@article{packer2023memgpt,
  title={MemGPT: Towards LLMs as Operating Systems.},
  author={Packer, Charles and Fang, Vivian and Patil, Shishir\_G and Lin, Kevin and Wooders, Sarah and Gonzalez, Joseph\_E},
  year={2023},
  publisher={ArXiv}
}

@inproceedings{transformer,
  author    = {Ashish Vaswani and
               Noam Shazeer and
               Niki Parmar and
               Jakob Uszkoreit and
               Llion Jones and
               Aidan N. Gomez and
               Lukasz Kaiser and
               Illia Polosukhin},
  editor    = {Isabelle Guyon and
               Ulrike von Luxburg and
               Samy Bengio and
               Hanna M. Wallach and
               Rob Fergus and
               S. V. N. Vishwanathan and
               Roman Garnett},
  title     = {Attention is All you Need},
  booktitle = {Advances in Neural Information Processing Systems 30: Annual Conference
               on Neural Information Processing Systems 2017, December 4-9, 2017,
               Long Beach, CA, {USA}},
  pages     = {5998--6008},
  year      = {2017},
}

@misc{GPT-5,
  author = {OpenAI},
  title = {Introducing GPT-5},
  howpublished = {\url{https://openai.com/index/introducing-gpt-5/}},
  year={2025},
}

@misc{ChatGPTAgent,
  author = {OpenAI},
  title = {Introducing ChatGPT agent: bridging research and action},
  howpublished = {\url{https://openai.com/index/introducing-chatgpt-agent/}},
  year={2025},
}

@misc{GPT-4.1,
  author = {OpenAI},
  title = {Introducing GPT-4.1 in the API},
  howpublished = {\url{https://openai.com/index/gpt-4-1/}},
  year={2025},
}

@misc{o3,
  author = {OpenAI},
  title = {Introducing OpenAI o3 and o4-mini},
  howpublished = {\url{https://openai.com/index/introducing-o3-and-o4-mini/}},
  year={2025},
}

@inproceedings{
    lora,
    title={Lo{RA}: Low-Rank Adaptation of Large Language Models},
    author={Edward J Hu and Yelong Shen and Phillip Wallis and Zeyuan Allen-Zhu and Yuanzhi Li and Shean Wang and Lu Wang and Weizhu Chen},
    booktitle={International Conference on Learning Representations},
    year={2022},
    url={https://openreview.net/forum?id=nZeVKeeFYf9}
}

@inproceedings{ReAct,
  author       = {Shunyu Yao and
                  Jeffrey Zhao and
                  Dian Yu and
                  Nan Du and
                  Izhak Shafran and
                  Karthik R. Narasimhan and
                  Yuan Cao},
  title        = {ReAct: Synergizing Reasoning and Acting in Language Models},
  booktitle    = {The Eleventh International Conference on Learning Representations,
                  {ICLR} 2023, Kigali, Rwanda, May 1-5, 2023},
  publisher    = {OpenReview.net},
  year         = {2023},
  url          = {https://openreview.net/forum?id=WE\_vluYUL-X},
}

@misc{R1,
      title={DeepSeek-R1: Incentivizing Reasoning Capability in LLMs via Reinforcement Learning}, 
      author={DeepSeek-AI and Daya Guo and Dejian Yang and Haowei Zhang and Junxiao Song and Ruoyu Zhang and Runxin Xu and Qihao Zhu and Shirong Ma and Peiyi Wang and Xiao Bi and Xiaokang Zhang and Xingkai Yu and Yu Wu and Z. F. Wu and Zhibin Gou and Zhihong Shao and Zhuoshu Li and Ziyi Gao and Aixin Liu and Bing Xue and Bingxuan Wang and Bochao Wu and Bei Feng and Chengda Lu and Chenggang Zhao and Chengqi Deng and Chenyu Zhang and Chong Ruan and Damai Dai and Deli Chen and Dongjie Ji and Erhang Li and Fangyun Lin and Fucong Dai and Fuli Luo and Guangbo Hao and Guanting Chen and Guowei Li and H. Zhang and Han Bao and Hanwei Xu and Haocheng Wang and Honghui Ding and Huajian Xin and Huazuo Gao and Hui Qu and Hui Li and Jianzhong Guo and Jiashi Li and Jiawei Wang and Jingchang Chen and Jingyang Yuan and Junjie Qiu and Junlong Li and J. L. Cai and Jiaqi Ni and Jian Liang and Jin Chen and Kai Dong and Kai Hu and Kaige Gao and Kang Guan and Kexin Huang and Kuai Yu and Lean Wang and Lecong Zhang and Liang Zhao and Litong Wang and Liyue Zhang and Lei Xu and Leyi Xia and Mingchuan Zhang and Minghua Zhang and Minghui Tang and Meng Li and Miaojun Wang and Mingming Li and Ning Tian and Panpan Huang and Peng Zhang and Qiancheng Wang and Qinyu Chen and Qiushi Du and Ruiqi Ge and Ruisong Zhang and Ruizhe Pan and Runji Wang and R. J. Chen and R. L. Jin and Ruyi Chen and Shanghao Lu and Shangyan Zhou and Shanhuang Chen and Shengfeng Ye and Shiyu Wang and Shuiping Yu and Shunfeng Zhou and Shuting Pan and S. S. Li and Shuang Zhou and Shaoqing Wu and Shengfeng Ye and Tao Yun and Tian Pei and Tianyu Sun and T. Wang and Wangding Zeng and Wanjia Zhao and Wen Liu and Wenfeng Liang and Wenjun Gao and Wenqin Yu and Wentao Zhang and W. L. Xiao and Wei An and Xiaodong Liu and Xiaohan Wang and Xiaokang Chen and Xiaotao Nie and Xin Cheng and Xin Liu and Xin Xie and Xingchao Liu and Xinyu Yang and Xinyuan Li and Xuecheng Su and Xuheng Lin and X. Q. Li and Xiangyue Jin and Xiaojin Shen and Xiaosha Chen and Xiaowen Sun and Xiaoxiang Wang and Xinnan Song and Xinyi Zhou and Xianzu Wang and Xinxia Shan and Y. K. Li and Y. Q. Wang and Y. X. Wei and Yang Zhang and Yanhong Xu and Yao Li and Yao Zhao and Yaofeng Sun and Yaohui Wang and Yi Yu and Yichao Zhang and Yifan Shi and Yiliang Xiong and Ying He and Yishi Piao and Yisong Wang and Yixuan Tan and Yiyang Ma and Yiyuan Liu and Yongqiang Guo and Yuan Ou and Yuduan Wang and Yue Gong and Yuheng Zou and Yujia He and Yunfan Xiong and Yuxiang Luo and Yuxiang You and Yuxuan Liu and Yuyang Zhou and Y. X. Zhu and Yanhong Xu and Yanping Huang and Yaohui Li and Yi Zheng and Yuchen Zhu and Yunxian Ma and Ying Tang and Yukun Zha and Yuting Yan and Z. Z. Ren and Zehui Ren and Zhangli Sha and Zhe Fu and Zhean Xu and Zhenda Xie and Zhengyan Zhang and Zhewen Hao and Zhicheng Ma and Zhigang Yan and Zhiyu Wu and Zihui Gu and Zijia Zhu and Zijun Liu and Zilin Li and Ziwei Xie and Ziyang Song and Zizheng Pan and Zhen Huang and Zhipeng Xu and Zhongyu Zhang and Zhen Zhang},
      year={2025},
      eprint={2501.12948},
      archivePrefix={arXiv},
      primaryClass={cs.CL},
      url={https://arxiv.org/abs/2501.12948}, 
}

@inproceedings{SeqKD,
  author       = {Yoon Kim and
                  Alexander M. Rush},
  editor       = {Jian Su and
                  Xavier Carreras and
                  Kevin Duh},
  title        = {Sequence-Level Knowledge Distillation},
  booktitle    = {Proceedings of the 2016 Conference on Empirical Methods in Natural
                  Language Processing, {EMNLP} 2016, Austin, Texas, USA, November 1-4,
                  2016},
  pages        = {1317--1327},
  publisher    = {The Association for Computational Linguistics},
  year         = {2016},
  url          = {https://doi.org/10.18653/v1/d16-1139},
}

@inproceedings{CodeAct,
  author       = {Xingyao Wang and
                  Yangyi Chen and
                  Lifan Yuan and
                  Yizhe Zhang and
                  Yunzhu Li and
                  Hao Peng and
                  Heng Ji},
  title        = {Executable Code Actions Elicit Better {LLM} Agents},
  booktitle    = {Forty-first International Conference on Machine Learning, {ICML} 2024,
                  Vienna, Austria, July 21-27, 2024},
  publisher    = {OpenReview.net},
  year         = {2024},
  url          = {https://openreview.net/forum?id=jJ9BoXAfFa}
}

@misc{Kimi-K2,
      title={Kimi K2: Open Agentic Intelligence}, 
      author={Kimi Team and Yifan Bai and Yiping Bao and Guanduo Chen and Jiahao Chen and Ningxin Chen and Ruijue Chen and Yanru Chen and Yuankun Chen and Yutian Chen and Zhuofu Chen and Jialei Cui and Hao Ding and Mengnan Dong and Angang Du and Chenzhuang Du and Dikang Du and Yulun Du and Yu Fan and Yichen Feng and Kelin Fu and Bofei Gao and Hongcheng Gao and Peizhong Gao and Tong Gao and Xinran Gu and Longyu Guan and Haiqing Guo and Jianhang Guo and Hao Hu and Xiaoru Hao and Tianhong He and Weiran He and Wenyang He and Chao Hong and Yangyang Hu and Zhenxing Hu and Weixiao Huang and Zhiqi Huang and Zihao Huang and Tao Jiang and Zhejun Jiang and Xinyi Jin and Yongsheng Kang and Guokun Lai and Cheng Li and Fang Li and Haoyang Li and Ming Li and Wentao Li and Yanhao Li and Yiwei Li and Zhaowei Li and Zheming Li and Hongzhan Lin and Xiaohan Lin and Zongyu Lin and Chengyin Liu and Chenyu Liu and Hongzhang Liu and Jingyuan Liu and Junqi Liu and Liang Liu and Shaowei Liu and T. Y. Liu and Tianwei Liu and Weizhou Liu and Yangyang Liu and Yibo Liu and Yiping Liu and Yue Liu and Zhengying Liu and Enzhe Lu and Lijun Lu and Shengling Ma and Xinyu Ma and Yingwei Ma and Shaoguang Mao and Jie Mei and Xin Men and Yibo Miao and Siyuan Pan and Yebo Peng and Ruoyu Qin and Bowen Qu and Zeyu Shang and Lidong Shi and Shengyuan Shi and Feifan Song and Jianlin Su and Zhengyuan Su and Xinjie Sun and Flood Sung and Heyi Tang and Jiawen Tao and Qifeng Teng and Chensi Wang and Dinglu Wang and Feng Wang and Haiming Wang and Jianzhou Wang and Jiaxing Wang and Jinhong Wang and Shengjie Wang and Shuyi Wang and Yao Wang and Yejie Wang and Yiqin Wang and Yuxin Wang and Yuzhi Wang and Zhaoji Wang and Zhengtao Wang and Zhexu Wang and Chu Wei and Qianqian Wei and Wenhao Wu and Xingzhe Wu and Yuxin Wu and Chenjun Xiao and Xiaotong Xie and Weimin Xiong and Boyu Xu and Jing Xu and Jinjing Xu and L. H. Xu and Lin Xu and Suting Xu and Weixin Xu and Xinran Xu and Yangchuan Xu and Ziyao Xu and Junjie Yan and Yuzi Yan and Xiaofei Yang and Ying Yang and Zhen Yang and Zhilin Yang and Zonghan Yang and Haotian Yao and Xingcheng Yao and Wenjie Ye and Zhuorui Ye and Bohong Yin and Longhui Yu and Enming Yuan and Hongbang Yuan and Mengjie Yuan and Haobing Zhan and Dehao Zhang and Hao Zhang and Wanlu Zhang and Xiaobin Zhang and Yangkun Zhang and Yizhi Zhang and Yongting Zhang and Yu Zhang and Yutao Zhang and Yutong Zhang and Zheng Zhang and Haotian Zhao and Yikai Zhao and Huabin Zheng and Shaojie Zheng and Jianren Zhou and Xinyu Zhou and Zaida Zhou and Zhen Zhu and Weiyu Zhuang and Xinxing Zu},
      year={2025},
      eprint={2507.20534},
      archivePrefix={arXiv},
      primaryClass={cs.LG},
      url={https://arxiv.org/abs/2507.20534}, 
}

@inproceedings{ContextualizeWeb,
  author       = {Dongjun Lee and
                  Juyong Lee and
                  Kyuyoung Kim and
                  Jihoon Tack and
                  Jinwoo Shin and
                  Yee Whye Teh and
                  Kimin Lee},
  title        = {Learning to Contextualize Web Pages for Enhanced Decision Making by
                  {LLM} Agents},
  booktitle    = {The Thirteenth International Conference on Learning Representations,
                  {ICLR} 2025, Singapore, April 24-28, 2025},
  publisher    = {OpenReview.net},
  year         = {2025},
  url          = {https://openreview.net/forum?id=3Gzz7ZQLiz},
}

@inproceedings{Mind2Act,
  author       = {Xiang Deng and
                  Yu Gu and
                  Boyuan Zheng and
                  Shijie Chen and
                  Samual Stevens and
                  Boshi Wang and
                  Huan Sun and
                  Yu Su},
  editor       = {Alice Oh and
                  Tristan Naumann and
                  Amir Globerson and
                  Kate Saenko and
                  Moritz Hardt and
                  Sergey Levine},
  title        = {Mind2Web: Towards a Generalist Agent for the Web},
  booktitle    = {Advances in Neural Information Processing Systems 36: Annual Conference
                  on Neural Information Processing Systems 2023, NeurIPS 2023, New Orleans,
                  LA, USA, December 10 - 16, 2023},
  year         = {2023},
  url          = {http://papers.nips.cc/paper\_files/paper/2023/hash/5950bf290a1570ea401bf98882128160\\-Abstract-Datasets\_and\_Benchmarks.html}
}

@inproceedings{SWE-agent,
  author       = {John Yang and
                  Carlos E. Jimenez and
                  Alexander Wettig and
                  Kilian Lieret and
                  Shunyu Yao and
                  Karthik Narasimhan and
                  Ofir Press},
  editor       = {Amir Globersons and
                  Lester Mackey and
                  Danielle Belgrave and
                  Angela Fan and
                  Ulrich Paquet and
                  Jakub M. Tomczak and
                  Cheng Zhang},
  title        = {SWE-agent: Agent-Computer Interfaces Enable Automated Software Engineering},
  booktitle    = {Advances in Neural Information Processing Systems 38: Annual Conference
                  on Neural Information Processing Systems 2024, NeurIPS 2024, Vancouver,
                  BC, Canada, December 10 - 15, 2024},
  year         = {2024},
  url          = {http://papers.nips.cc/paper\_files/paper/2024/hash/5a7c947568c1b1328ccc5230172e1e7c\\-Abstract-Conference.html}
}

@inproceedings{SWE-bench,
  author       = {Carlos E. Jimenez and
                  John Yang and
                  Alexander Wettig and
                  Shunyu Yao and
                  Kexin Pei and
                  Ofir Press and
                  Karthik R. Narasimhan},
  title        = {SWE-bench: Can Language Models Resolve Real-world Github Issues?},
  booktitle    = {The Twelfth International Conference on Learning Representations,
                  {ICLR} 2024, Vienna, Austria, May 7-11, 2024},
  publisher    = {OpenReview.net},
  year         = {2024},
  url          = {https://openreview.net/forum?id=VTF8yNQM66},
}

@article{openhands-condenser,
  title = {OpenHands Context Condensensation for More Efficient AI Agents},
  author = {Calvin Smith},
  journal = {All Hands AI Blog},
  year = {2025},
  month = {April},
  day = {9},
  url = {https://www.all-hands.dev/blog/openhands-context-condensensation-for\\-more-efficient-ai-agents}
}

@inproceedings{LLMLingua,
  author       = {Huiqiang Jiang and
                  Qianhui Wu and
                  Chin{-}Yew Lin and
                  Yuqing Yang and
                  Lili Qiu},
  editor       = {Houda Bouamor and
                  Juan Pino and
                  Kalika Bali},
  title        = {LLMLingua: Compressing Prompts for Accelerated Inference of Large
                  Language Models},
  booktitle    = {Proceedings of the 2023 Conference on Empirical Methods in Natural
                  Language Processing, {EMNLP} 2023, Singapore, December 6-10, 2023},
  pages        = {13358--13376},
  publisher    = {Association for Computational Linguistics},
  year         = {2023},
  url          = {https://doi.org/10.18653/v1/2023.emnlp-main.825},
}

@inproceedings{LLMLingua2,
  author       = {Zhuoshi Pan and
                  Qianhui Wu and
                  Huiqiang Jiang and
                  Menglin Xia and
                  Xufang Luo and
                  Jue Zhang and
                  Qingwei Lin and
                  Victor R{\"{u}}hle and
                  Yuqing Yang and
                  Chin{-}Yew Lin and
                  H. Vicky Zhao and
                  Lili Qiu and
                  Dongmei Zhang},
  editor       = {Lun{-}Wei Ku and
                  Andre Martins and
                  Vivek Srikumar},
  title        = {LLMLingua-2: Data Distillation for Efficient and Faithful Task-Agnostic
                  Prompt Compression},
  booktitle    = {Findings of the Association for Computational Linguistics, {ACL} 2024,
                  Bangkok, Thailand and virtual meeting, August 11-16, 2024},
  pages        = {963--981},
  publisher    = {Association for Computational Linguistics},
  year         = {2024},
  url          = {https://doi.org/10.18653/v1/2024.findings-acl.57},
}

@inproceedings{irrelevant-context,
  author       = {Freda Shi and
                  Xinyun Chen and
                  Kanishka Misra and
                  Nathan Scales and
                  David Dohan and
                  Ed H. Chi and
                  Nathanael Sch{\"{a}}rli and
                  Denny Zhou},
  editor       = {Andreas Krause and
                  Emma Brunskill and
                  Kyunghyun Cho and
                  Barbara Engelhardt and
                  Sivan Sabato and
                  Jonathan Scarlett},
  title        = {Large Language Models Can Be Easily Distracted by Irrelevant Context},
  booktitle    = {International Conference on Machine Learning, {ICML} 2023, 23-29 July
                  2023, Honolulu, Hawaii, {USA}},
  series       = {Proceedings of Machine Learning Research},
  volume       = {202},
  pages        = {31210--31227},
  publisher    = {{PMLR}},
  year         = {2023},
  url          = {https://proceedings.mlr.press/v202/shi23a.html},
}

@inproceedings{efficient-long-context-lm,
  author       = {Minju Seo and
                  Jinheon Baek and
                  Seongyun Lee and
                  Sung Ju Hwang},
  editor       = {Wanxiang Che and
                  Joyce Nabende and
                  Ekaterina Shutova and
                  Mohammad Taher Pilehvar},
  title        = {Efficient Long Context Language Model Retrieval with Compression},
  booktitle    = {Proceedings of the 63rd Annual Meeting of the Association for Computational
                  Linguistics (Volume 1: Long Papers), {ACL} 2025, Vienna, Austria,
                  July 27 - August 1, 2025},
  pages        = {15251--15268},
  publisher    = {Association for Computational Linguistics},
  year         = {2025},
  url          = {https://aclanthology.org/2025.acl-long.740/}
}

@inproceedings{CompAct,
  author       = {Chanwoong Yoon and
                  Taewhoo Lee and
                  Hyeon Hwang and
                  Minbyul Jeong and
                  Jaewoo Kang},
  editor       = {Yaser Al{-}Onaizan and
                  Mohit Bansal and
                  Yun{-}Nung Chen},
  title        = {CompAct: Compressing Retrieved Documents Actively for Question Answering},
  booktitle    = {Proceedings of the 2024 Conference on Empirical Methods in Natural
                  Language Processing, {EMNLP} 2024, Miami, FL, USA, November 12-16,
                  2024},
  pages        = {21424--21439},
  publisher    = {Association for Computational Linguistics},
  year         = {2024},
  url          = {https://doi.org/10.18653/v1/2024.emnlp-main.1194}
}

@inproceedings{selective-context,
  author       = {Yucheng Li and
                  Bo Dong and
                  Frank Guerin and
                  Chenghua Lin},
  editor       = {Houda Bouamor and
                  Juan Pino and
                  Kalika Bali},
  title        = {Compressing Context to Enhance Inference Efficiency of Large Language
                  Models},
  booktitle    = {Proceedings of the 2023 Conference on Empirical Methods in Natural
                  Language Processing, {EMNLP} 2023, Singapore, December 6-10, 2023},
  pages        = {6342--6353},
  publisher    = {Association for Computational Linguistics},
  year         = {2023},
  url          = {https://doi.org/10.18653/v1/2023.emnlp-main.391}
}

@inproceedings{recomp,
  author       = {Fangyuan Xu and
                  Weijia Shi and
                  Eunsol Choi},
  title        = {{RECOMP:} Improving Retrieval-Augmented LMs with Context Compression
                  and Selective Augmentation},
  booktitle    = {The Twelfth International Conference on Learning Representations,
                  {ICLR} 2024, Vienna, Austria, May 7-11, 2024},
  publisher    = {OpenReview.net},
  year         = {2024},
  url          = {https://openreview.net/forum?id=mlJLVigNHp}
}

@inproceedings{LongLLMLingua,
  author       = {Huiqiang Jiang and
                  Qianhui Wu and
                  Xufang Luo and
                  Dongsheng Li and
                  Chin{-}Yew Lin and
                  Yuqing Yang and
                  Lili Qiu},
  editor       = {Lun{-}Wei Ku and
                  Andre Martins and
                  Vivek Srikumar},
  title        = {LongLLMLingua: Accelerating and Enhancing LLMs in Long Context Scenarios
                  via Prompt Compression},
  booktitle    = {Proceedings of the 62nd Annual Meeting of the Association for Computational
                  Linguistics (Volume 1: Long Papers), {ACL} 2024, Bangkok, Thailand,
                  August 11-16, 2024},
  pages        = {1658--1677},
  publisher    = {Association for Computational Linguistics},
  year         = {2024},
  url          = {https://doi.org/10.18653/v1/2024.acl-long.91}
}

@article{MEM1,
  author       = {Zijian Zhou and
                  Ao Qu and
                  Zhaoxuan Wu and
                  Sunghwan Kim and
                  Alok Prakash and
                  Daniela Rus and
                  Jinhua Zhao and
                  Bryan Kian Hsiang Low and
                  Paul Pu Liang},
  title        = {{MEM1:} Learning to Synergize Memory and Reasoning for Efficient Long-Horizon
                  Agents},
  journal      = {arXiv},
  volume       = {2506.15841},
  year         = {2025},
  url          = {https://doi.org/10.48550/arXiv.2506.15841},
}

@inproceedings{OSWorld,
  author       = {Tianbao Xie and
                  Danyang Zhang and
                  Jixuan Chen and
                  Xiaochuan Li and
                  Siheng Zhao and
                  Ruisheng Cao and
                  Toh Jing Hua and
                  Zhoujun Cheng and
                  Dongchan Shin and
                  Fangyu Lei and
                  Yitao Liu and
                  Yiheng Xu and
                  Shuyan Zhou and
                  Silvio Savarese and
                  Caiming Xiong and
                  Victor Zhong and
                  Tao Yu},
  editor       = {Amir Globersons and
                  Lester Mackey and
                  Danielle Belgrave and
                  Angela Fan and
                  Ulrich Paquet and
                  Jakub M. Tomczak and
                  Cheng Zhang},
  title        = {OSWorld: Benchmarking Multimodal Agents for Open-Ended Tasks in Real
                  Computer Environments},
  booktitle    = {Advances in Neural Information Processing Systems 38: Annual Conference
                  on Neural Information Processing Systems 2024, NeurIPS 2024, Vancouver,
                  BC, Canada, December 10 - 15, 2024},
  year         = {2024},
  url          = {http://papers.nips.cc/paper\_files/paper/2024/hash/5d413e48f84dc61244b6be550f1cd8f5\\-Abstract-Datasets\_and\_Benchmarks\_Track.html},
}

@article{WindowsAgentArena,
  author       = {Rogerio Bonatti and
                  Dan Zhao and
                  Francesco Bonacci and
                  Dillon Dupont and
                  Sara Abdali and
                  Yinheng Li and
                  Yadong Lu and
                  Justin Wagle and
                  Kazuhito Koishida and
                  Arthur Bucker and
                  Lawrence Jang and
                  Zack Hui},
  title        = {Windows Agent Arena: Evaluating Multi-Modal {OS} Agents at Scale},
  journal      = {arXiv},
  volume       = {2409.08264},
  year         = {2024},
  url          = {https://doi.org/10.48550/arXiv.2409.08264},
}

@article{BrowserComp,
  author       = {Jason Wei and
                  Zhiqing Sun and
                  Spencer Papay and
                  Scott McKinney and
                  Jeffrey Han and
                  Isa Fulford and
                  Hyung Won Chung and
                  Alex Tachard Passos and
                  William Fedus and
                  Amelia Glaese},
  title        = {BrowseComp: {A} Simple Yet Challenging Benchmark for Browsing Agents},
  journal      = {arXiv},
  volume       = {2504.12516},
  year         = {2025},
  url          = {https://doi.org/10.48550/arXiv.2504.12516},
}

@inproceedings{WebArena,
  author       = {Shuyan Zhou and
                  Frank F. Xu and
                  Hao Zhu and
                  Xuhui Zhou and
                  Robert Lo and
                  Abishek Sridhar and
                  Xianyi Cheng and
                  Tianyue Ou and
                  Yonatan Bisk and
                  Daniel Fried and
                  Uri Alon and
                  Graham Neubig},
  title        = {WebArena: {A} Realistic Web Environment for Building Autonomous Agents},
  booktitle    = {The Twelfth International Conference on Learning Representations,
                  {ICLR} 2024, Vienna, Austria, May 7-11, 2024},
  publisher    = {OpenReview.net},
  year         = {2024},
  url          = {https://openreview.net/forum?id=oKn9c6ytLx},
}

@inproceedings{ALFWorld,
  author       = {Mohit Shridhar and
                  Xingdi Yuan and
                  Marc{-}Alexandre C{\^{o}}t{\'{e}} and
                  Yonatan Bisk and
                  Adam Trischler and
                  Matthew J. Hausknecht},
  title        = {ALFWorld: Aligning Text and Embodied Environments for Interactive
                  Learning},
  booktitle    = {9th International Conference on Learning Representations, {ICLR} 2021,
                  Virtual Event, Austria, May 3-7, 2021},
  publisher    = {OpenReview.net},
  year         = {2021},
  url          = {https://openreview.net/forum?id=0IOX0YcCdTn},
}

@inproceedings{appworld,
  author       = {Harsh Trivedi and
                  Tushar Khot and
                  Mareike Hartmann and
                  Ruskin Manku and
                  Vinty Dong and
                  Edward Li and
                  Shashank Gupta and
                  Ashish Sabharwal and
                  Niranjan Balasubramanian},
  editor       = {Lun{-}Wei Ku and
                  Andre Martins and
                  Vivek Srikumar},
  title        = {AppWorld: {A} Controllable World of Apps and People for Benchmarking
                  Interactive Coding Agents},
  booktitle    = {Proceedings of the 62nd Annual Meeting of the Association for Computational
                  Linguistics (Volume 1: Long Papers), {ACL} 2024, Bangkok, Thailand,
                  August 11-16, 2024},
  pages        = {16022--16076},
  publisher    = {Association for Computational Linguistics},
  year         = {2024},
  url          = {https://doi.org/10.18653/v1/2024.acl-long.850},
}

@article{OfficeBench,
  author       = {Zilong Wang and
                  Yuedong Cui and
                  Li Zhong and
                  Zimin Zhang and
                  Da Yin and
                  Bill Yuchen Lin and
                  Jingbo Shang},
  title        = {OfficeBench: Benchmarking Language Agents across Multiple Applications
                  for Office Automation},
  journal      = {arXiv},
  volume       = {2407.19056},
  year         = {2024},
  url          = {https://doi.org/10.48550/arXiv.2407.19056}
}

@misc{OdysseyBench,
      title={OdysseyBench: Evaluating LLM Agents on Long-Horizon Complex Office Application Workflows}, 
      author={Weixuan Wang and Dongge Han and Daniel Madrigal Diaz and Jin Xu and Victor Rühle and Saravan Rajmohan},
      year={2025},
      eprint={2508.09124},
      archivePrefix={arXiv},
      primaryClass={cs.CL},
      url={https://arxiv.org/abs/2508.09124}, 
}

@article{METR,
  author       = {Thomas Kwa and
                  Ben West and
                  Joel Becker and
                  Amy Deng and
                  Katharyn Garcia and
                  Max Hasin and
                  Sami Jawhar and
                  Megan Kinniment and
                  Nate Rush and
                  Sydney von Arx and
                  Ryan Bloom and
                  Thomas Broadley and
                  Haoxing Du and
                  Brian Goodrich and
                  Nikola Jurkovic and
                  Luke Harold Miles and
                  Seraphina Nix and
                  Tao Lin and
                  Neev Parikh and
                  David Rein and
                  Lucas Jun Koba Sato and
                  Hjalmar Wijk and
                  Daniel M. Ziegler and
                  Elizabeth Barnes and
                  Lawrence Chan},
  title        = {Measuring {AI} Ability to Complete Long Tasks},
  journal      = {arXiv},
  volume       = {2503.14499},
  year         = {2025},
  url          = {https://doi.org/10.48550/arXiv.2503.14499}
}

@inproceedings{OPRO,
  author       = {Chengrun Yang and
                  Xuezhi Wang and
                  Yifeng Lu and
                  Hanxiao Liu and
                  Quoc V. Le and
                  Denny Zhou and
                  Xinyun Chen},
  title        = {Large Language Models as Optimizers},
  booktitle    = {The Twelfth International Conference on Learning Representations,
                  {ICLR} 2024, Vienna, Austria, May 7-11, 2024},
  publisher    = {OpenReview.net},
  year         = {2024},
  url          = {https://openreview.net/forum?id=Bb4VGOWELI},
}

@inproceedings{
DSPy,
title={{DSP}y: Compiling Declarative Language Model Calls into State-of-the-Art Pipelines},
author={Omar Khattab and Arnav Singhvi and Paridhi Maheshwari and Zhiyuan Zhang and Keshav Santhanam and Sri Vardhamanan A and Saiful Haq and Ashutosh Sharma and Thomas T. Joshi and Hanna Moazam and Heather Miller and Matei Zaharia and Christopher Potts},
booktitle={The Twelfth International Conference on Learning Representations},
year={2024},
url={https://openreview.net/forum?id=sY5N0zY5Od}
}

@article{TextGrad,
  author       = {Mert Y{\"{u}}ksekg{\"{o}}n{\"{u}}l and
                  Federico Bianchi and
                  Joseph Boen and
                  Sheng Liu and
                  Pan Lu and
                  Zhi Huang and
                  Carlos Guestrin and
                  James Zou},
  title        = {Optimizing generative {AI} by backpropagating language model feedback},
  journal      = {Nature},
  volume       = {639},
  number       = {8055},
  pages        = {609--616},
  year         = {2025},
  url          = {https://doi.org/10.1038/s41586-025-08661-4},
}

@inproceedings{TACO-RL,
  author       = {Shivam Shandilya and
                  Menglin Xia and
                  Supriyo Ghosh and
                  Huiqiang Jiang and
                  Jue Zhang and
                  Qianhui Wu and
                  Victor R{\"{u}}hle and
                  Saravan Rajmohan},
  editor       = {Wanxiang Che and
                  Joyce Nabende and
                  Ekaterina Shutova and
                  Mohammad Taher Pilehvar},
  title        = {{TACO-RL:} Task Aware Prompt Compression Optimization with Reinforcement
                  Learning},
  booktitle    = {Findings of the Association for Computational Linguistics, {ACL} 2025,
                  Vienna, Austria, July 27 - August 1, 2025},
  pages        = {1582--1597},
  publisher    = {Association for Computational Linguistics},
  year         = {2025},
  url          = {https://aclanthology.org/2025.findings-acl.81/},
}

@article{blueprint,
  author       = {Dongge Han and
                  Menglin Xia and
                  Daniel Madrigal Diaz and
                  Samuel Kessler and
                  Ankur Mallick and
                  Xuchao Zhang and
                  Mirian del Carmen Hipolito Garcia and
                  Jin Xu and
                  Victor R{\"{u}}hle and
                  Saravan Rajmohan},
  title        = {Enhancing Reasoning Capabilities of Small Language Models with Blueprints
                  and Prompt Template Search},
  journal      = {arXiv},
  volume       = {2506.08669},
  year         = {2025},
  url          = {https://doi.org/10.48550/arXiv.2506.08669},
}

@inproceedings{APO,
  author       = {Reid Pryzant and
                  Dan Iter and
                  Jerry Li and
                  Yin Tat Lee and
                  Chenguang Zhu and
                  Michael Zeng},
  editor       = {Houda Bouamor and
                  Juan Pino and
                  Kalika Bali},
  title        = {Automatic Prompt Optimization with "Gradient Descent" and Beam Search},
  booktitle    = {Proceedings of the 2023 Conference on Empirical Methods in Natural
                  Language Processing, {EMNLP} 2023, Singapore, December 6-10, 2023},
  pages        = {7957--7968},
  publisher    = {Association for Computational Linguistics},
  year         = {2023},
  url          = {https://doi.org/10.18653/v1/2023.emnlp-main.494},
}

@article{NQ,
  author       = {Tom Kwiatkowski and
                  Jennimaria Palomaki and
                  Olivia Redfield and
                  Michael Collins and
                  Ankur P. Parikh and
                  Chris Alberti and
                  Danielle Epstein and
                  Illia Polosukhin and
                  Jacob Devlin and
                  Kenton Lee and
                  Kristina Toutanova and
                  Llion Jones and
                  Matthew Kelcey and
                  Ming{-}Wei Chang and
                  Andrew M. Dai and
                  Jakob Uszkoreit and
                  Quoc Le and
                  Slav Petrov},
  title        = {Natural Questions: a Benchmark for Question Answering Research},
  journal      = {Trans. Assoc. Comput. Linguistics},
  volume       = {7},
  pages        = {452--466},
  year         = {2019},
  url          = {https://doi.org/10.1162/tacl\_a\_00276},
}

@article{LightThinker,
  author       = {Jintian Zhang and
                  Yuqi Zhu and
                  Mengshu Sun and
                  Yujie Luo and
                  Shuofei Qiao and
                  Lun Du and
                  Da Zheng and
                  Huajun Chen and
                  Ningyu Zhang},
  title        = {LightThinker: Thinking Step-by-Step Compression},
  journal      = {arXiv},
  volume       = {2502.15589},
  year         = {2025},
  url          = {https://doi.org/10.48550/arXiv.2502.15589},
}

@article{A-MEM,
  author       = {Wujiang Xu and
                  Zujie Liang and
                  Kai Mei and
                  Hang Gao and
                  Juntao Tan and
                  Yongfeng Zhang},
  title        = {{A-MEM:} Agentic Memory for {LLM} Agents},
  journal      = {arXiv},
  volume       = {2502.12110},
  year         = {2025},
  url          = {https://doi.org/10.48550/arXiv.2502.12110},
}

@article{Phi4,
  author       = {Marah I Abdin and
                  Jyoti Aneja and
                  Harkirat S. Behl and
                  S{\'{e}}bastien Bubeck and
                  Ronen Eldan and
                  Suriya Gunasekar and
                  Michael Harrison and
                  Russell J. Hewett and
                  Mojan Javaheripi and
                  Piero Kauffmann and
                  James R. Lee and
                  Yin Tat Lee and
                  Yuanzhi Li and
                  Weishung Liu and
                  Caio C. T. Mendes and
                  Anh Nguyen and
                  Eric Price and
                  Gustavo de Rosa and
                  Olli Saarikivi and
                  Adil Salim and
                  Shital Shah and
                  Xin Wang and
                  Rachel Ward and
                  Yue Wu and
                  Dingli Yu and
                  Cyril Zhang and
                  Yi Zhang},
  title        = {Phi-4 Technical Report},
  journal      = {arXiv},
  volume       = {2412.08905},
  year         = {2024},
  url          = {https://doi.org/10.48550/arXiv.2412.08905},
}

@article{Qwen3,
  author       = {An Yang and
                  Anfeng Li and
                  Baosong Yang and
                  Beichen Zhang and
                  Binyuan Hui and
                  Bo Zheng and
                  Bowen Yu and
                  Chang Gao and
                  Chengen Huang and
                  Chenxu Lv and
                  Chujie Zheng and
                  Dayiheng Liu and
                  Fan Zhou and
                  Fei Huang and
                  Feng Hu and
                  Hao Ge and
                  Haoran Wei and
                  Huan Lin and
                  Jialong Tang and
                  Jian Yang and
                  Jianhong Tu and
                  Jianwei Zhang and
                  Jian Yang and
                  Jiaxi Yang and
                  Jingren Zhou and
                  Junyang Lin and
                  Kai Dang and
                  Keqin Bao and
                  Kexin Yang and
                  Le Yu and
                  Lianghao Deng and
                  Mei Li and
                  Mingfeng Xue and
                  Mingze Li and
                  Pei Zhang and
                  Peng Wang and
                  Qin Zhu and
                  Rui Men and
                  Ruize Gao and
                  Shixuan Liu and
                  Shuang Luo and
                  Tianhao Li and
                  Tianyi Tang and
                  Wenbiao Yin and
                  Xingzhang Ren and
                  Xinyu Wang and
                  Xinyu Zhang and
                  Xuancheng Ren and
                  Yang Fan and
                  Yang Su and
                  Yichang Zhang and
                  Yinger Zhang and
                  Yu Wan and
                  Yuqiong Liu and
                  Zekun Wang and
                  Zeyu Cui and
                  Zhenru Zhang and
                  Zhipeng Zhou and
                  Zihan Qiu},
  title        = {Qwen3 Technical Report},
  journal      = {arXiv},
  volume       = {2505.09388},
  year         = {2025},
  url          = {https://doi.org/10.48550/arXiv.2505.09388},
}

@book{RL,
	author = {Sutton, Richard S. and Barto, Andrew G.},
	edition = {Second},
	publisher = {The MIT Press},
	title = {Reinforcement Learning: An Introduction},
	year = {2018 }
}

@article{Search-R1,
  author       = {Bowen Jin and
                  Hansi Zeng and
                  Zhenrui Yue and
                  Dong Wang and
                  Hamed Zamani and
                  Jiawei Han},
  title        = {Search-R1: Training LLMs to Reason and Leverage Search Engines with
                  Reinforcement Learning},
  journal      = {arXiv},
  volume       = {2503.09516},
  year         = {2025},
  url          = {https://doi.org/10.48550/arXiv.2503.09516},
}

@misc{EpiCache,
      title={EpiCache: Episodic KV Cache Management for Long Conversational Question Answering}, 
      author={Minsoo Kim and Arnav Kundu and Han-Byul Kim and Richa Dixit and Minsik Cho},
      year={2025},
      eprint={2509.17396},
      archivePrefix={arXiv},
      primaryClass={cs.CL},
      url={https://arxiv.org/abs/2509.17396}, 
}

@inproceedings{CascadeKV,
  author       = {Jeffrey Willette and
                  Heejun Lee and
                  Youngwan Lee and
                  Myeongjae Jeon and
                  Sung Ju Hwang},
  title        = {Training Free Exponential Context Extension via Cascading {KV} Cache},
  booktitle    = {The Thirteenth International Conference on Learning Representations,
                  {ICLR} 2025, Singapore, April 24-28, 2025},
  publisher    = {OpenReview.net},
  year         = {2025},
  url          = {https://openreview.net/forum?id=dSneEp59yX},
}

@inproceedings{claude,
  title={The Claude 3 Model Family: Opus, Sonnet, Haiku},
  author={Anthropic},
  year         = {2024},
  url={https://api.semanticscholar.org/CorpusID:268232499}
}

@article{gemini,
  author       = {Gheorghe Comanici and
                  Eric Bieber and
                  Mike Schaekermann and
                  Ice Pasupat and
                  Noveen Sachdeva and
                  Inderjit S. Dhillon and
                  Marcel Blistein and
                  Ori Ram and
                  Dan Zhang and
                  Evan Rosen and
                  Luke Marris and
                  Sam Petulla and
                  Colin Gaffney and
                  Asaf Aharoni and
                  Nathan Lintz and
                  Tiago Cardal Pais and
                  Henrik Jacobsson and
                  Idan Szpektor and
                  Nan{-}Jiang Jiang and
                  Krishna Haridasan and
                  Ahmed Omran and
                  Nikunj Saunshi and
                  Dara Bahri and
                  Gaurav Mishra and
                  Eric Chu and
                  Toby Boyd and
                  Brad Hekman and
                  Aaron Parisi and
                  Chaoyi Zhang and
                  Kornraphop Kawintiranon and
                  Tania Bedrax{-}Weiss and
                  Oliver Wang and
                  Ya Xu and
                  Ollie Purkiss and
                  Uri Mendlovic and
                  Ila{\"{\i}} Deutel and
                  Nam Nguyen and
                  Adam Langley and
                  Flip Korn and
                  Lucia Rossazza and
                  Alexandre Ram{\'{e}} and
                  Sagar Waghmare and
                  Helen Miller and
                  Nathan Byrd and
                  Ashrith Sheshan and
                  Raia Hadsell Sangnie Bhardwaj and
                  Pawel Janus and
                  Tero Rissa and
                  Dan Horgan and
                  Sharon Silver and
                  Ayzaan Wahid and
                  Sergey Brin and
                  Yves Raimond and
                  Klemen Kloboves and
                  Cindy Wang and
                  Nitesh Bharadwaj Gundavarapu and
                  Ilia Shumailov and
                  Bo Wang and
                  Mantas Pajarskas and
                  Joe Heyward and
                  Martin Nikoltchev and
                  Maciej Kula and
                  Hao Zhou and
                  Zachary Garrett and
                  Sushant Kafle and
                  Sercan Arik and
                  Ankita Goel and
                  Mingyao Yang and
                  Jiho Park and
                  Koji Kojima and
                  Parsa Mahmoudieh and
                  Koray Kavukcuoglu and
                  Grace Chen and
                  Doug Fritz and
                  Anton Bulyenov and
                  Sudeshna Roy and
                  Dimitris Paparas and
                  Hadar Shemtov and
                  Bo{-}Juen Chen and
                  Robin Strudel and
                  David Reitter and
                  Aurko Roy and
                  Andrey Vlasov and
                  Changwan Ryu and
                  Chas Leichner and
                  Haichuan Yang and
                  Zelda Mariet and
                  Denis Vnukov and
                  Tim Sohn and
                  Amy Stuart and
                  Wei Liang and
                  Minmin Chen and
                  Praynaa Rawlani and
                  Christy Koh and
                  JD Co{-}Reyes and
                  Guangda Lai and
                  Praseem Banzal and
                  Dimitrios Vytiniotis and
                  Jieru Mei and
                  Mu Cai},
  title        = {Gemini 2.5: Pushing the Frontier with Advanced Reasoning, Multimodality,
                  Long Context, and Next Generation Agentic Capabilities},
  journal      = {arXiv},
  volume       = {2507.06261},
  year         = {2025},
  url          = {https://doi.org/10.48550/arXiv.2507.06261}
}

@article{GRPO,
  author       = {Zhihong Shao and
                  Peiyi Wang and
                  Qihao Zhu and
                  Runxin Xu and
                  Junxiao Song and
                  Mingchuan Zhang and
                  Y. K. Li and
                  Y. Wu and
                  Daya Guo},
  title        = {DeepSeekMath: Pushing the Limits of Mathematical Reasoning in Open
                  Language Models},
  journal      = {arXiv},
  volume       = {2402.03300},
  year         = {2024},
  url          = {https://doi.org/10.48550/arXiv.2402.03300},
}

@article{Agentfold,
  author       = {Rui Ye and
                  Zhongwang Zhang and
                  Kuan Li and
                  Huifeng Yin and
                  Zhengwei Tao and
                  Yida Zhao and
                  Liangcai Su and
                  Liwen Zhang and
                  Zile Qiao and
                  Xinyu Wang and
                  Pengjun Xie and
                  Fei Huang and
                  Siheng Chen and
                  Jingren Zhou and
                  Yong Jiang},
  title        = {AgentFold: Long-Horizon Web Agents with Proactive Context Management},
  journal      = {arXiv},
  volume       = {2510.24699},
  year         = {2025},
  url          = {https://doi.org/10.48550/arXiv.2510.24699}
}

@article{MemAgent,
  author       = {Hongli Yu and
                  Tinghong Chen and
                  Jiangtao Feng and
                  Jiangjie Chen and
                  Weinan Dai and
                  Qiying Yu and
                  Ya{-}Qin Zhang and
                  Wei{-}Ying Ma and
                  Jingjing Liu and
                  Mingxuan Wang and
                  Hao Zhou},
  title        = {MemAgent: Reshaping Long-Context {LLM} with Multi-Conv RL-based Memory
                  Agent},
  journal      = {arXiv},
  volume       = {2507.02259},
  year         = {2025},
  url          = {https://doi.org/10.48550/arXiv.2507.02259}
}

@article{Contextfolding,
  author       = {Weiwei Sun and
                  Miao Lu and
                  Zhan Ling and
                  Kang Liu and
                  Xuesong Yao and
                  Yiming Yang and
                  Jiecao Chen},
  title        = {Scaling Long-Horizon {LLM} Agent via Context-Folding},
  journal      = {arXiv},
  volume       = {2510.11967},
  year         = {2025},
  url          = {https://doi.org/10.48550/arXiv.2510.11967}
}

@misc{ReSum,
      title={ReSum: Unlocking Long-Horizon Search Intelligence via Context Summarization}, 
      author={Xixi Wu and Kuan Li and Yida Zhao and Liwen Zhang and Litu Ou and Huifeng Yin and Zhongwang Zhang and Xinmiao Yu and Dingchu Zhang and Yong Jiang and Pengjun Xie and Fei Huang and Minhao Cheng and Shuai Wang and Hong Cheng and Jingren Zhou},
      year={2026},
      eprint={2509.13313},
      archivePrefix={arXiv},
      primaryClass={cs.CL},
      url={https://arxiv.org/abs/2509.13313}, 
}

@inproceedings{WebVoyager,
    title = "{W}eb{V}oyager: Building an End-to-End Web Agent with Large Multimodal Models",
    author = "He, Hongliang  and
      Yao, Wenlin  and
      Ma, Kaixin  and
      Yu, Wenhao  and
      Dai, Yong  and
      Zhang, Hongming  and
      Lan, Zhenzhong  and
      Yu, Dong",
    editor = "Ku, Lun-Wei  and
      Martins, Andre  and
      Srikumar, Vivek",
    booktitle = "Proceedings of the 62nd Annual Meeting of the Association for Computational Linguistics (Volume 1: Long Papers)",
    month = aug,
    year = "2024",
    address = "Bangkok, Thailand",
    publisher = "Association for Computational Linguistics",
    url = "https://aclanthology.org/2024.acl-long.371/"
}

@inproceedings{
MetaSPO,
title={System Prompt Optimization with Meta-Learning},
author={Yumin Choi and Jinheon Baek and Sung Ju Hwang},
booktitle={The Thirty-ninth Annual Conference on Neural Information Processing Systems},
year={2026},
url={https://openreview.net/forum?id=IYVknFxsJb}
}
\bibliographystyle{icml2026}

\newpage
\appendix
\onecolumn

\section{Limitations \& Future Works}
\label{sec:appendix:limitations}
Our work addresses the context management problem in long-horizon LLM agents and proposes a framework for optimizing context compression. 
While the method effectively reduces token costs with minimal performance degradation, it also presents several limitations and potentials for future works.

\paragraph{Scope of empirical evaluation.} 
Our work primarily focuses on the GPT models due to computational and budgetary constraints. 
While the proposed framework is designed to be model-agnostic, its empirical generalizability to other foundation models, such as Gemini~\citep{gemini} or Claude~\citep{claude}, has yet to be extensively verified. Furthermore, the inclusion of massive open-source models like DeepSeek-R1~\citep{R1} or Qwen3-235B \citep{Qwen3} was restricted by GPU availability. 
Extending the analysis to these models would provide stronger evidence of robustness and broaden the applicability of our conclusions.

\paragraph{Real-world deployability}
Although our validation on three distinct benchmarks reflects realistic agentic scenarios, these environments remain controlled. Transitioning from benchmark-centric evaluations to in-the-wild deployment---where task complexity is stochastic and environmental constraints are less predictable---remains an open challenge. We identify the integration of our framework into live, multi-agent production systems as a valuable next step.

\paragraph{Convergence of natural language space optimization}
The optimization of our compression guidelines relies on LLM-generated feedback, a strategy aligned with recent prompt optimization work~\citep{TextGrad, DSPy, APO, OPRO}. However, unlike traditional numerical gradient-based optimization, this approach lacks a formal convergence guarantee. While our method of sampling multiple candidates and selecting the best performer partially mitigates this, it does not provide a principled theoretical foundation. 
A deeper analysis of the natural language space optimization process and more principled methods for optimizing the objective
in~\ref{sec:prompt_opt} would be valuable directions for future work.

\paragraph{Distillation gap and data scalability}
A performance gap remains between our distilled models and their teacher models. We hypothesize that our current dataset (100 examples per domain) may not fully capture the reasoning required for complete behavioral cloning. 
Future work will investigate whether increasing the diversity and volume of training data can eliminate the residual performance gap.

\paragraph{Latency and KV-cache dynamics}
A significant, yet often overlooked, challenge is the computational overhead inherent in the compression process itself---a limitation shared by the majority of existing context compression frameworks~\citep{ContextualizeWeb, openhands-condenser, SWE-agent, Contextfolding, MEM1, Agentfold}. 
In transformer-based architectures, history compression typically invalidates the existing KV-cache, necessitating a costly re-computation of the entire compressed sequence.
While observation compression can mitigate this by reducing the context before caching them, the generative overhead required for the compression procedure itself still introduces latency, often slowing down the agent’s response time \citep{ContextualizeWeb}.
Consequently, the field currently faces a trade-off where the gains in long-horizon efficiency are partially offset by the increased computational cost of the compression itself.
To move beyond these constraints, the development of KV-cache eviction or compression strategies---specifically optimized for the dynamic, multi-turn nature of agentic workflows---can be a potential avenue for future research, building upon foundational works in long-context modeling \citep{LightThinker, CascadeKV, EpiCache}.

\section{Experimental Setup Details}
\label{sec:appendix:exp_setup}
\subsection{Datasets}
\label{sec:appendix:datasets}
\paragraph{AppWorld~\citep{appworld}.}
AppWorld is our primary benchmark, providing a high-quality execution environment that integrates nine everyday applications (Spotify, SimpleNote, Amazon, Venmo, Gmail, Splitwise, File system, Todoist, and Phone) through 457 APIs. 
It also includes realistic simulations of approximately 100 functional users. 
This benchmark is particularly suitable for evaluating long-horizon productivity agents, as its multi-step tasks require an average of 42.5 API calls per task. 
We follow the official split, using 90 training tasks for guideline optimization and distillation, and 168 test-normal tasks for evaluation. 
An example trajectory from AppWorld is provided in Example~\ref{ex:appworld}.

\paragraph{OfficeBench~\citep{OfficeBench}.}
OfficeBench is a benchmark for office automation using applications such as Word, Excel, PDF, Calendar, Email, Shell, and Calculator. 
It evaluates the ability of agents to coordinate across multiple apps to complete complex tasks, making it well suited for long-horizon scenarios. 
Tasks are categorized as 1-app, 2-app, or 3-app depending on the number of applications required. 
We restrict our experiments to text-related tasks, excluding those requiring OCR, as OCR quality could confound the evaluation. 
Since no official split is available, we randomly partition the tasks into training and test sets with a 1:1 ratio, resulting in 92 training tasks and 95 test tasks. 
We additionally refine the dataset by removing ambiguous tasks and ensuring that synthetic files (testbeds) are not shared across splits.

\paragraph{8-Objective QA~\citep{MEM1}.}
The 8-objective QA benchmark simulates deep research-style agentic tasks. 
Unlike conventional multi-hop QA, which requires answering a single question using multiple pieces of evidence, this benchmark poses eight distinct questions within one task, and the agent must provide answers to all of them at the end. 
This design creates a more challenging setting for long-horizon agents. 
Following~\citet{MEM1}, we construct each task by grouping eight questions together. 
Questions are drawn from NaturalQuestions~\citep{NQ}, resulting in 100 training tasks (from the train split) and 100 test tasks (from the test split). 
For retrieval, we use a BM25 retriever over the 2018 Wikipedia knowledge base, following~\citet{Search-R1}.

We include the example task of each benchmark in~\autoref{tab:task_examples}.

\begin{table*}[t]
\centering
\caption{Example tasks across benchmarks.}
\label{tab:task_examples}
\renewcommand{\arraystretch}{1.3}
\begin{tabularx}{\textwidth}{lX}
\toprule
\textbf{Benchmark / Difficulty} & \textbf{Example Task} \\
\midrule
\multicolumn{2}{l}{\textbf{AppWorld}} \\
Easy   & Mark ``Taking a solo backpacking trip'' in my Bucket List Simple Note as not done. \\
Medium & Like all the Venmo transactions from today involving any of my roommates on my Venmo social feed. \\
Hard   & Start playing a playlist on Spotify that has enough songs for my workout today. I do not want to have to change the playlist in the middle of my workout. The workout plan is in Simple Note. \\
\midrule
\multicolumn{2}{l}{\textbf{OfficeBench}} \\
1-app  & Create a new Word file called \texttt{random\_paragraph.docx} and add the content in \texttt{random\_paragraph.txt} to it. \\
2-app  & Analyze Excel data of students' grade and generate a teaching report in \texttt{teaching.docx}. \\
3-app  & Read company revenues and send an email with subject \texttt{revenues}, containing data to Bob for reporting, also write a \texttt{revenue.docx} to summarize it. \\
\midrule
\multicolumn{2}{l}{\textbf{8-objective QA}} \\
-- & who wrote the song \emph{Oceans Where Feet May Fail}?; who plays Eddie the Eagle in the movie?; when was the last time England were in the final of World Cup?; who plays Chelsea's mom on \emph{Young and the Restless}?; what is the largest coin in the US?; who sang \emph{Even the Bad Times Are Good}?; who sings \emph{This Is My Town} country song?; which of the Guianas is not an independent country? \\
\bottomrule
\end{tabularx}
\end{table*}

\subsection{Evaluation Metrics}\label{subsec:eval_metrics}
For efficiency evaluation, we adopt two metrics—\emph{peak tokens} and \emph{dependency}—introduced in LightThinker~\citep{LightThinker} and MEM1~\citep{MEM1}.

\paragraph{Peak tokens.} 
Peak tokens are measured as the maximum number of tokens observed in any single sequence throughout the agent’s trajectory, excluding system prompts. 
This metric serves as a proxy for inference-time memory requirements and corresponds to the maximum peak shown in~\autoref{fig:motivation}.

\paragraph{Dependency.} 
Dependency is defined as the area under the curve in~\autoref{fig:motivation}. 
At each step $t$, given the number of input tokens $n_i^{(t)}$ and output tokens $n_o^{(t)}$, it is calculated as:
\begin{equation}
    \sum_{t \in [T]} \frac{(n_i^{(t)} + 2n_o^{(t)}) \times n_o^{(t)}}{2}.
\end{equation}
This metric approximates the cumulative computational cost incurred by action generation across the trajectory.

\paragraph{API Cost.} For the cost analysis, we use the official OpenAI pricing (as of September 2025) for \texttt{gpt-4.1} and \texttt{gpt-4.1-mini}~\citep{GPT-4.1}. Specifically, \texttt{gpt-4.1} is priced at \$3.00 per 1M input tokens and \$12.00 per 1M output tokens.
For \texttt{gpt-4.1-mini}, the costs are \$0.80 per 1M input tokens and \$3.20 per 1M output tokens.
For \texttt{Qwen3-14B}~\citep{Qwen3}, since no official API pricing is available, we approximate the cost using OpenRouter\footnote{\url{https://openrouter.ai/}}
: \$0.06 per 1M input tokens and \$0.24 per 1M output tokens.

\subsection{Implementation Details \& Hyperparameters}
\label{sec:appendix:details}
\paragraph{API Inference.}
We set temperature 0.0 and fix the seed 42.
Note that there is still non-determinism with fixing the seed and setting temperature as 0.
To reduce the instability, we use the API snapshot form Azure OpenAI endpoint \texttt{gpt-4.1-2025-04-14} and \texttt{gpt-4.1-mini-2025-04-14}.

\paragraph{Compression.}
For history compression, we set $T_{\text{hist}} = 4096$ for AppWorld and OfficeBench, and $2048$ for 8-objective QA.
We keep the last action, observation pair to preserve the latest information. This is the same for \ours and all baselines.
For observation compression, we set $T_{\text{obs}} = 1024$ for AppWorld, $512$ for OfficeBench, and $400$ for 8-objective QA.

\paragraph{Prompt Optimization.}
We use the OpenAI \texttt{o3} model~\citep{o3} for both analysis and update of prompts. 
During the update stage, we sample 5 candidate prompts and select the one that performs best on a subset of the training set. 

The prompts used in each stage and step are provided as follows:
\begin{itemize}
    \item Analysis prompt for~\UTIL step: Prompt~\ref{box:prompt_optimizer_analysis}
    \item Update prompt for~\UTIL step: Prompt~\ref{box:prompt_optimizer_update}
    \item Analysis prompt for~\COMP step: Prompt~\ref{box:prompt_optimizer_analysis_compression}
    \item Update prompt for~\COMP step: Prompt~\ref{box:prompt_optimizer_update_compression}
\end{itemize}

We also provide the detailed procedure in Algorithm~\ref{alg:prompt_opt}. 
For the subset used in prompt selection during the~\UTIL step, we consider training tasks in $\gD_{\sf cont}^{(r)}$ where the agent succeeds without compression but fails with compression. 
For the~\COMP step, we use training tasks in $\gD_{\sf succ}^{(r)}$ where the agent succeeds with compression.
We perform one round consisting of a single \UTIL step and a single \COMP step to obtain the guidelines used in our experiments, unless otherwise noted.

\paragraph{Baselines}
For FIFO, we keep last 5 interaction turns which fits in similar compression rate in average with \ours.
For retrieval, we also retrieve 4 interaction turns and keep the last turn. We use OpenAI \texttt{text-embedding-3-large} for embedding.
For LLMLingua, we set keep rate as 30\% for both history and observation.
For naive prompting, we use the similar prompt with~\citet{ContextualizeWeb} and do some human prompt engineering to specialize each prompt to history or observation compression.

\paragraph{Compressor \& Agent Distillation}
Both compressor and agent distillation use LoRA~\citep{lora} with rank 16, $\alpha=32$, learning rate $10^{-4}$, 3 epochs, batch size 4, and maximum sequence length $10{,}000$.  
We adopt linear warmup ($5\%$ ratio), weight decay $0.01$, and AdamW optimizer.  
No hyperparameter tuning was performed; the same setup is applied across all models and benchmarks.  
We sample a single generation from the teacher for fine-tuning, leaving potential improvements from hyperparameter tuning or multi-sample training for future work.
We use 1 A100 80GB GPU for both training and inference.
For inference of fine-tuned models, we use greedy decoding  (temperature 0.0).

\section{Additional Results}
\label{appendix:sec:results}

We provide additional quantitative results to complement the main experiments in~\autoref{sec:exp}.

\subsection{Results with different agent models}
\paragraph{Experiments with gpt-4.1-mini.} 
Results for the smaller variant gpt-4.1-mini~\citep{GPT-4.1} across three benchmarks are reported in AppWorld (\autoref{tab:appendix:1_appworld_mini}), OfficeBench (\autoref{tab:appendix:4_officebench_mini}), and 8-objective QA (\autoref{tab:appendix:3_qa_mini}). 
The trends of \ours are consistent with those for gpt-4.1 in~\autoref{sec:exp}. 
In particular, \autoref{tab:appendix:1_appworld_mini} shows that history compression improves the performance of gpt-4.1-mini compared to the baseline, complementing the findings in~\autoref{sec:exp:agent_distillation} that \ours enhances the effectiveness of smaller LM agents. 
These results highlight the robustness of our method under resource-constrained settings.

\paragraph{Experiments with gpt-5-chat.}
We also evaluate on AppWorld using gpt-5-chat~\citep{GPT-5}, as reported in~\autoref{tab:appendix:1_appworld_gpt5}. 
The improvements follow the same trend as with gpt-4.1, demonstrating that \ours generalizes to the latest stronger proprietary models.

\subsection{Detailed results and analyses}
\label{appendix:subsec:detailed_results}
\paragraph{OfficeBench difficulty breakdowns.}
We further analyze OfficeBench with gpt-4.1 by difficulty level. 
The detailed breakdown in~\autoref{tab:appendix:4_officebench_main} shows that \ours yields the largest gains on the most challenging tasks in Level~3.

\paragraph{Distilled optimizer.}
Additional results for the distilled optimizer in AppWorld are shown in~\autoref{tab:appendix:5_appworld_distill_opt}. 
Beyond the analysis in~\autoref{sec:exp:compressor_distillation}, we also include experiments where the compressor is distilled using guidelines without optimization. 
The results confirm that optimized guidelines consistently yield stronger performance when distilled into smaller models.

\paragraph{History and observation compression.}
In~\autoref{tab:appendix:1_appworld_additional}, we report ablations with gpt-4.1 using both history and observation compression. 
While combining the two compressions achieves larger reductions in peak token usage and dependency, it also leads to substantial performance degradation compared to applying either compression alone.

\paragraph{Additional guideline optimization step.}
We investigate whether running an extra utility maximization step (\UTIL) after the standard sequence of utility maximization and compression maximization (\COMP) is beneficial. 
As shown in~\autoref{tab:appendix:1_appworld_additional}, this additional iteration results in a performance drop, indicating that a single round of optimization is sufficient for effective guideline learning.

\begin{figure*}
    \centering
    \includegraphics[width=1.0\linewidth]{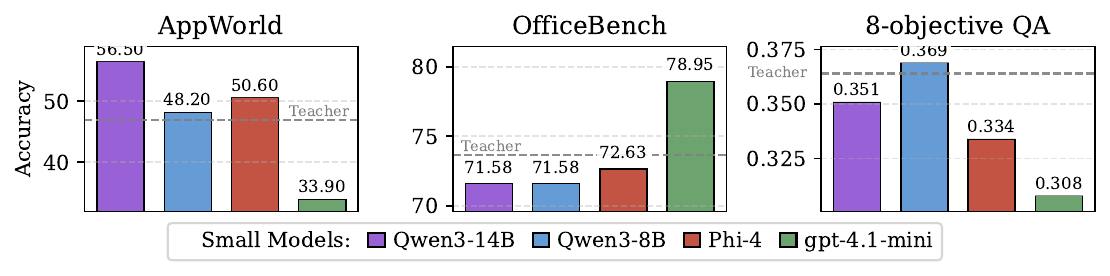}
    \caption{\textbf{Results of distilled compressors on observation compression} with gpt-4.1 as the agent. Student models (Qwen3-14B, Qwen3-8B, Phi-4) are distilled from gpt-4.1 compressor using the optimized compression guideline after~\UTIL step, and evaluated across all benchmarks. We also include result with gpt-4.1-mini without distillation for comparison.}
    \label{fig:exp:obs_distill}
\end{figure*}

\paragraph{Distilled compressor for observation.}
In addition to~\autoref{sec:exp:compressor_distillation}, we report results for observation compressor distillation in~\autoref{fig:exp:obs_distill}. 
Similar to history compression, the performance is largely preserved after distillation, confirming that \ours enables effective transfer of optimized observation compressors to smaller models.

\paragraph{API cost details on compression guideline optimization.}
In this section, we provide a detailed breakdown of the computational costs associated with our framework. All cost estimates are based on the official API pricing as in~\autoref{subsec:eval_metrics}. The total expense is categorized into two phases: trajectory rollout and guideline optimization. \textit{(1) Trajectory Rollout (Data Collection).} This phase accounts for the majority of the budget. We utilize \texttt{gpt-4.1} to collect trajectories on the training set (\eg 90 examples for AppWorld). For each example, we generate two trajectories: one without compression and one with compressed context. The total cost for collecting these rollout trajectories amounts to approximately \$20. It is important to note that this is a one-time data preparation cost which can be amortized across future agent runs or substituted with existing offline logs. \textit{(2) Guideline Optimization.} Despite utilizing the reasoning-intensive \texttt{o3} model, the optimization process is highly cost-efficient. The procedure runs for a single number of iteration. In each iteration, the optimizer generates 5 candidate guidelines and performs one \UTIL step and one \COMP step. The total cost for the entire optimization loop is consistently under \$2 per domain.

\paragraph{Additional Results on WebVoyager}

To further evaluate ACON beyond simulated productivity and question-answering environments, we conduct additional experiments on WebVoyager~\citep{WebVoyager}, a web-agent benchmark that requires agents to interact with webpages and complete user-specified tasks. We use 50 training tasks for compression guideline optimization and evaluate the resulting guidelines on 70 held-out test tasks.
We evaluate WebVoyager using gpt-4.1 as both the agent and compressor, following the main experimental setting.
\begin{wraptable}{r}{0.32\linewidth}
\centering
\caption{Additional results on WebVoyager. We report task accuracy, peak input tokens ($10^3$), and dependency ($10^6$).}
\label{tab}
\resizebox{\linewidth}{!}{
\begin{tabular}{lccc}
\toprule
Method & Acc. $\uparrow$ & Peak $\downarrow$ & Dep. $\downarrow$ \\
\midrule
No compression & 35.7 & 13.28 & 2.53 \\
\midrule
\multicolumn{4}{l}{\textbf{History Compression}} \\
Prompting & 42.9 & \textbf{7.97} & 1.23 \\
\ours & \textbf{48.6} & 8.04 & \textbf{1.19} \\
\midrule
\multicolumn{4}{l}{\textbf{Observation Compression}} \\
Prompting & 45.7 & \textbf{3.91} & 0.85 \\
\ours & \textbf{47.1} & 4.19 & \textbf{0.80} \\
\bottomrule
\end{tabular}
}
\vspace{-1em}
\end{wraptable}
This setting introduces different context dynamics from AppWorld, OfficeBench, and 8-objective QA, as the agent must process verbose webpage observations represented as accessibility trees (AXTrees) while maintaining relevant interaction history across multiple steps~\citep{ContextualizeWeb}.

As reported in Table~\ref{tab}, \ours improves task accuracy over both no compression and prompting baselines while substantially reducing peak tokens and dependency compared to no compression. 
For history compression, \ours improves accuracy from 42.9 to 48.6 compared to prompting, with comparable peak tokens and lower dependency. For observation compression, \ours improves accuracy from 45.7 to 47.1 and reduces dependency from 0.85 to 0.80. 
These results suggest that the optimized compression guidelines learned by \ours are effective not only in structured tool-use environments, but also in web-agent settings where observations are verbose and dynamically changing.

\paragraph{Case study: history compression turns failure into success.}
A notable case study illustrates how history compression enables a smaller agent to succeed on tasks that would otherwise fail.
In the uncompressed trajectory in Example~\ref{ex:appworld_fail_no_compress}, the gpt-4.1-mini agent repeatedly attempted to use the \texttt{file\_system} APIs without managing authentication, leading to persistent 401 Unauthorized errors. 
After compressing the history as in Example~\ref{ex:appworld_success_compress}, however, the compressed history retained only the essential reasoning steps: the need for both username and password, the importance of passing the returned \texttt{access\_token} into subsequent calls, and the absence of proxy APIs in the supervisor app.

This compressed context prevented redundant exploration and guided the agent directly to the correct sequence---login with full credentials, capture the token, and provide it explicitly in \texttt{show\_directory} and \texttt{delete\_file} calls. 
As a result, the agent was able to enumerate and remove all .pdf files in ~/downloads, a task it had previously failed. 
This example highlights how compression does not merely shorten history but clarifies critical dependencies, turning a failure trajectory into a successful one.

\begin{table}[t]
\centering
\caption{Comparison between \ours and MEM1-like methods. The two methods operate under different assumptions regarding model accessibility, training requirements, and architectural coupling.}
\vspace{-0.1in}
\label{tab:appendix_mem1_comparison}
\small
\renewcommand{\arraystretch}{1.1}
\begin{tabular}{@{}%
  >{\raggedright\arraybackslash}p{0.25\linewidth}%
  >{\raggedright\arraybackslash}p{0.34\linewidth}%
  >{\raggedright\arraybackslash}p{0.34\linewidth}@{}}\toprule
\textbf{Dimension} & \textbf{\ours (ours)} & \textbf{MEM1}-like~\citep{MEM1, Contextfolding, Agentfold} \\
\midrule
Is agent model training not required? &
\cmark\ no agent model training or weight updates required &
\xmark\ requires RL training on the agent model \\
\midrule
Can the method work without access to model weights? &
\cmark\ works with both open-source and proprietary API models &
\xmark\ requires full model access and gradients \\
\midrule
Can the agent and compressor be different models? &
\cmark\ supports decoupled design with different model sizes &
\xmark\ reasoning and compression are integrated into a single model \\
\midrule
Is it possible to use a large agent with a small compressor? &
\cmark\ supports combinations like gpt-4.1 agent + Qwen3-14B compressor &
\xmark\ same model must serve as both agent and compressor \\
\midrule
Does optimization avoid GPU-based RL cost? &
\cmark\ under \$2 for guideline optimization, no GPU needed &
\xmark\ RL policy training requires multiple trajectories and GPU computation \\
\bottomrule
\end{tabular}
\end{table}

\subsection{Comparison with MEM1}
\label{appendix:subsec:comparison}
MEM1~\citep{MEM1} and concurrent works~\citep{Contextfolding, Agentfold} propose a learnable context compression policy trained jointly with the agent through reinforcement learning. This design couples reasoning and compression within a single trainable model and requires full access to model weights and gradient updates. In contrast, our method can perform optimization entirely at the prompt-level without any weight updates and enables the agent and compressor to be different models.

This decoupling allows combinations that are not possible in MEM1 and other similar methods. 
For example, one can use a large proprietary model such as gpt-4.1 as the agent while employing a smaller open-source model such as Qwen3-14B as the compressor after distillation, a configuration that other MEM1-like methods cannot support due to its unified training requirement. This flexibility makes \ours applicable to both open-source and proprietary API-based models, including settings where model weights are inaccessible. A detailed comparison is summarized in~\autoref{tab:appendix_mem1_comparison}.

\section{Qualitative Examples}
\label{appendix:sec:examples}

We complement the quantitative results with qualitative illustrations.

\paragraph{Compression guidelines.}
We present examples of compression guidelines before and after optimization in AppWorld. 
The history compression guideline before optimization is shown in Prompt~\ref{box:appworld_history_compression}, the optimized version (\UTIL) in Prompt~\ref{box:appworld_history_optimized}, and the optimized version (\UTIL\COMP) in Prompt~\ref{box:appworld_history_optimized_len}.
Similarly, observation compression guideline examples are provided in Prompt~\ref{box:appworld_obs_compression} and Prompt~\ref{box:appworld_obs_optimized}, and the optimized version (\UTIL\COMP) in Prompt~\ref{box:appworld_obs_optimized_len}.
These comparisons demonstrate that optimization yields more targeted guidelines for compressors.

\paragraph{Compressed histories.}
Compression Example~\ref{compressbox:appworld_history} illustrates history segments before and after guideline optimization in AppWorld with gpt-4.1. 
The optimized guideline retains a more detailed record of task progress, including variable states and guardrails for the environment. 
After the compression maximization step (\COMP), the histories become shorter while still preserving the essential information required for future decision-making. 
This qualitative evidence demonstrates how our framework improves both the efficiency and effectiveness of context compression, complementing the guideline optimization procedure described in~\autoref{sec:prompt_opt}. 

We also present Compression Example~\ref{compressbox:qa_history} for 8-objective QA and Compression Example~\ref{compressbox:officebench_history} for OfficeBench, which confirm that the effects of guideline optimization are consistent across benchmarks.

\paragraph{Compressed observations.}
Compression Example~\ref{compressbox:appworld_obs} shows observations before and after guideline optimization in AppWorld. 
We illustrate the case of printing available APIs for the Spotify app, which produces a lengthy observation. 
The optimized guideline yields a more structured and faithful representation: whereas naive prompting loses the JSON format and omits the crucial ``play\_music'' API, the optimized version preserves both structure and key functionality necessary to complete the task.

\clearpage
\begin{algorithm*}[t]
\caption{Compression Guideline Optimization (\UTIL\ $\leftrightarrow$ \COMP)}
\label{alg:prompt_opt}
\begin{algorithmic}[1]

\STATE {\bfseries Input:} training indices $\mathcal{I}$; fixed agent $\gM(\cdot;\theta,\gP_{\mathrm{agent}})$; compressor $f(\cdot;\phi,\gP)$; initial guideline $\gP^{(0)}$; tradeoff $\lambda \ge 0$; rounds $R$; candidates $K$
\STATE {\bfseries Output:} optimized guideline $\gP^{*}$

\STATE {\bfseries Notation:} for each $i \in \mathcal{I}$ and guideline $\gP$:
\STATE \hspace*{1em} baseline (no compression): context $\mH_i$ with success $r_i^{\mathrm{base}} \in \{0,1\}$
\STATE \hspace*{1em} compressed: $\mH'_i(\gP)$ with success $r_i(\gP) \in \{0,1\}$ and cost
\STATE \hspace*{1em} $C(\mH'_i(\gP)) = \sum_t \gC(\vh_{i,t-1}', o_{i,t}')$

\STATE {\bfseries \# 0) Collect baseline contexts (no compression)}
\FORALL{$i \in \mathcal{I}$}
  \STATE Run $\gM$ without compression to obtain $\mH_i$ and $r_i^{\mathrm{base}}$
\ENDFOR
\STATE $\mathcal{I}^{+} \gets \{ i \in \mathcal{I} \mid r_i^{\mathrm{base}} = 1 \}$ \hfill {\itshape \# indices where baseline succeeds}

\FOR{$r = 0$ {\bfseries to} $2R-2$}
  \STATE {\bfseries \# Note:} $r$ takes even values only (i.e., $0,2,\ldots,2R-2$)

  \STATE {\bfseries \# Stage A: \UTIL\ (reward-first update using $\mH$ vs.\ $\mH'$)}
  \FORALL{$i \in \mathcal{I}$}
    \STATE Run $\gM$ with compression $f(\cdot;\phi,\gP^{(r)})$ to obtain $\mH'_i(\gP^{(r)})$, $r_i(\gP^{(r)})$, $C(\mH'_i(\gP^{(r)}))$
  \ENDFOR

  \STATE $\gD_{\mathrm{cont}}^{(r)} \gets \{ (\mH_i,\mH'_i(\gP^{(r)})) \mid i \in \mathcal{I}^{+},\ r_i(\gP^{(r)}) = 0 \}$
  \FORALL{$(\mH,\mH') \in \gD_{\mathrm{cont}}^{(r)}$}
    \STATE {\itshape \# contrastive feedback: what did $\mH'$ miss vs.\ $\mH$?}
    \STATE $\mathrm{Feedback} \gets \mathrm{LLM}(\mathrm{FeedbackInstr}, \mH, \mH')$
    \STATE Append $\mathrm{Feedback}$ to multiset $\mathcal{F}_{\mathrm{util}}$
  \ENDFOR

  \STATE $\{\gP^{(r+1)}_k\}_{k=1}^{K} \gets \mathrm{LLM}(\mathrm{UpdateInstr}, \gP^{(r)}, \Vert_{f \in \mathcal{F}_{\mathrm{util}}} f)$
  \STATE {\itshape \# where $\Vert$ denotes concatenation}
  \STATE {\bfseries \# Select by reward:} evaluate on a held-out subset of $\mathcal{I}^{+}$ and pick
  \STATE $k^{*}_{\mathrm{util}} \gets \arg\max_{k}\ \mathrm{SuccessRate}\big(\{ r_i(\gP^{(r+1)}_k) \}_{i \in \mathcal{I}^{+}}\big)$
  \STATE $\gP_{\mathrm{util}}^{(r+1)} \gets \gP^{(r+1)}_{k^{*}_{\mathrm{util}}}$

  \STATE {\bfseries \# Stage B: \COMP\ (cost-minimizing refinement using only $\mH'$)}
  \FORALL{$i \in \mathcal{I}$}
    \STATE Using $\gP_{\mathrm{util}}^{(r+1)}$, obtain $\mH'_i$, $r_i$, $C(\mH'_i)$
  \ENDFOR

  \STATE $\gD_{\mathrm{succ}}^{(r)} \gets \{ \mH'_i \mid r_i = 1 \}$
  \FORALL{$\mH' \in \gD_{\mathrm{succ}}^{(r)}$}
    \STATE {\itshape \# find redundant spans within $\mH'$}
    \STATE $\mathrm{CompFeedback} \gets \mathrm{LLM}(\mathrm{CompressInstr}, \mH')$
    \STATE Append $\mathrm{CompFeedback}$ to multiset $\mathcal{F}_{\mathrm{comp}}$
  \ENDFOR

  \STATE $\{\tilde{\gP}^{(r+2)}_k\}_{k=1}^{K} \gets \mathrm{LLM}(\mathrm{UpdateInstr}, \gP_{\mathrm{util}}^{(r+1)}, \Vert_{f \in \mathcal{F}_{\mathrm{comp}}} f)$
  \STATE {\bfseries \# Select by reward-cost:} evaluate on a held-out split of $\mathcal{I}$ and pick
  \STATE $k^{*}_{\mathrm{comp}} \gets \arg\max_{k}\Big(\mathrm{SuccessRate}(\{ r_i(\tilde{\gP}^{(r+2)}_k) )\}) - \lambda \cdot \mathrm{NormCost}(\{ C(\mH'_i(\tilde{\gP}^{(r+2)}_k)) \})\Big)$
  \STATE $\gP^{(r+2)} \gets \tilde{\gP}^{(r+2)}_{k^{*}_{\mathrm{comp}}}$

  \IF{early-stop criterion satisfied}
    \STATE {\bfseries break} \hfill {\itshape \# e.g., success/cost convergence or budget met}
  \ENDIF
\ENDFOR

\STATE $\gP^{*} \gets \gP^{(r+2)}$
\STATE {\bfseries return} $\gP^{*}$

\end{algorithmic}
\end{algorithm*}

\begin{table*}[t]
\centering
\caption{Detailed results on \textbf{OfficeBench} benchmark. We report accuracy (\%), and efficiency metrics: average steps, peak input tokens ($10^3$), and dependency ($10^6$) for Average and each difficulty level. Best values are in bold. Rows in blue background indicate the results from \colorbox{lightblue}{\textbf{ours}}.}
\label{tab:appendix:4_officebench_main}
\resizebox{\textwidth}{!}{
\begin{tabular}{lccccccccccccc}
\toprule
& \multicolumn{4}{c}{\textbf{Average (All)}} & \multicolumn{3}{c}{\textbf{Level 1 (1-app, 42)}} & \multicolumn{3}{c}{\textbf{Level 2 (2-app, 22)}} & \multicolumn{3}{c}{\textbf{Level 3 (3-app, 31)}} \\
\cmidrule(lr){2-5} \cmidrule(lr){6-8} \cmidrule(lr){9-11} \cmidrule(lr){12-14}
\textbf{Method} & Acc. $\uparrow$ & Steps $\downarrow$ & Peak $\downarrow$ & Dep. $\downarrow$
& Acc. $\uparrow$ & Peak $\downarrow$ & Dep. $\downarrow$
& Acc. $\uparrow$ & Peak $\downarrow$ & Dep. $\downarrow$
& Acc. $\uparrow$ & Peak $\downarrow$ & Dep. $\downarrow$ \\
\midrule
\rowcolor{lightgray} \multicolumn{14}{c}{\textbf{Agent:} \texttt{gpt-4.1} / \textbf{Compressor:} \texttt{gpt-4.1}} \\
\midrule
No Compression                     & \bf 76.84 & 11.52 & 7.27 & 4.43 & \bf 92.86 & 6.23 & 4.05 & 77.27 & 6.14 & 1.81 & 54.84 & 8.37 & 6.08 \\
\midrule
\multicolumn{14}{l}{\textbf{History Compression}} \\
FIFO                  & 67.37 & 12.26 & {\bf 4.02} & 2.64 & 83.33 & 4.19 & 0.72 & 63.64 & {\bf 3.51} & {\bf 1.01} & 48.39 & {\bf 4.23} & 4.39 \\
Retrieval             & 65.26 & 16.20 & 4.33 & 2.06 & 85.71 & 4.35 & 0.84 & 63.64 & 3.52 & 1.37 & 38.71 & 4.78 & 2.99 \\
LLMLingua             & 70.53 & 10.89 & 4.65 & 1.85 & 83.33 & 4.17 & {\bf 0.67} & 68.18 & 4.61 & 1.18 & 54.84 & 4.88 & 2.74 \\
Prompting             & 71.58 & {\bf 10.13} & 4.40 & {\bf 1.10} & 85.71 & 4.18 & 0.81 & 77.27 & 4.53 & 1.08 & 48.39 & 4.42 & {\bf 1.23} \\
\rowcolor{lightblue} \ours~\UTIL & 74.74 & 13.13 & 4.93 & 3.85 & 85.71 & 4.71 & 6.89 & 72.73 & 4.64 & 1.44 & 61.29 & 5.19 & 3.89 \\
\rowcolor{lightblue} \ours~\UTIL\COMP & 72.63 & 11.54 & 4.54 & 1.91 & 88.10 & 3.92 & 0.76 & 72.73 & 4.72 & 1.16 & 51.61 & 4.71 & 2.84 \\
\midrule
\multicolumn{14}{l}{\textbf{Observation Compression}} \\
LLMLingua             & 71.58 & 11.89 & 7.38 & 6.14 & 80.95 & 7.35 & 12.40 & 72.73 & 6.31 & 2.11 & {\bf 58.06} & 7.99 & 5.70 \\
Prompting             & 55.79 & 12.24 & 6.44 & 2.68 & 78.57 & 4.51 & 0.98 & 50.00 & 6.98 & 2.61 & 29.03 & 6.98 & 3.46 \\
\rowcolor{lightblue} \ours~\UTIL & 73.68 & 10.83 & 6.55 & 3.85 & 90.48 & 6.57 & 8.02 & 77.27 & 6.11 & 1.97 & 48.39 & 6.80 & 3.10 \\
\rowcolor{lightblue} \ours~\UTIL\COMP & 72.63 & 10.28 & 6.17 & 2.88 & 88.10 & 4.75 & 0.82 & 72.73 & 6.41 & 2.09 & 51.61 & 6.65 & 4.22 \\
\bottomrule
\end{tabular}
}
\end{table*}
\begin{table*}[t]
\centering
\caption{Results across different difficulty levels on \textbf{Appworld} benchmark (test-normal) with gpt-4.1-mini. We adopt the same compression guidelines as those used in the gpt-4.1 experiments. Each block reports accuracy (task goal completion score), average steps, average peak input tokens ($10^3$), and average dependency ($10^6$) for agents. Best results in each column are highlighted in bold. Rows in blue background indicate the results from \colorbox{lightblue}{\textbf{ours}}.}
\label{tab:appendix:1_appworld_mini}
\resizebox{\textwidth}{!}{
\begin{tabular}{lccccccccccccc}
\toprule
\multirow{2}{*}{Method} & 
\multicolumn{4}{c}{Average (168)} & 
\multicolumn{3}{c}{Easy (57)} & 
\multicolumn{3}{c}{Medium (48)} & 
\multicolumn{3}{c}{Hard (63)} \\
\cmidrule(lr){2-5} \cmidrule(lr){6-8} \cmidrule(lr){9-11} \cmidrule(lr){12-14}
& Acc. $\uparrow$ & Steps $\downarrow$ & Peak $\downarrow$ & Dep.$\downarrow$
& Acc. $\uparrow$ & Peak $\downarrow$ & Dep.$\downarrow$
& Acc. $\uparrow$ & Peak $\downarrow$ & Dep.$\downarrow$
& Acc. $\uparrow$ & Peak $\downarrow$ & Dep.$\downarrow$ \\
\midrule
\rowcolor{lightgray}\multicolumn{14}{c}{\textbf{Agent:} \texttt{gpt-4.1-mini} / \textbf{Compressor:} \texttt{gpt-4.1-mini}}\\
\midrule
No compression & 35.7 & 18.14 & 8.55 & 5.07 & 56.1 & 6.45 & 3.72 & 31.2 & 8.31 & 4.79 & 20.6 & 10.64 & 9.18 \\
\midrule[0.1pt]
\multicolumn{14}{l}{\textbf{History Compression}}\\
FIFO           & 39.3 & 30.39 & 6.18 & 5.24 & 75.4 & 4.76 & 2.66 & 35.4 & 5.33 & 4.81 & 9.5 & 8.10 & 7.91 \\
Retrieval & 14.9 & 40.18 & 7.49 & 5.95 & 36.8 & 7.10 & 4.29 & 8.3 & 7.44 & 6.80 & 0.0 & 7.89 & 6.81 \\
LLMLingua & 36.3 & 28.41 & 7.24 & 6.65 & 66.7 & 6.96 & 3.84 & 33.3 & 7.05 & 7.60 & 11.1 & 7.62 & 8.47 \\
Prompting & 35.7 & 24.98 & 6.56 & 4.95 & 64.9 & 5.96 & 2.90 & 27.1 & 6.65 & 5.35 & 15.9 & 6.84 & 6.49 \\
\rowcolor{lightblue} \ours~\UTIL & 42.3 & 22.46 & 6.51 & 5.48 & 64.9 & 5.87 & 2.62 & 37.5 & 7.18 & 5.22 & 25.4 & 7.18 & 8.25 \\
\rowcolor{lightblue}\ours~\UTIL\COMP & 32.7 & 24.27 & 6.99 & 4.97 & 57.9 & 7.50 & 2.77 & 33.3 & 8.45 & 4.99 & 9.5 & 6.95 & 6.97 \\
\midrule[0.1pt]
\multicolumn{14}{l}{\textbf{Observation Compression}}\\
LLMLingua & 25.6 & 20.75 & 8.04 & 8.21 & 38.6 & 6.13 & 3.03 & 27.1 & 8.74 & 13.78 & 12.7 & 9.24 & 8.65 \\
Prompting & 33.9 & 16.71 & 6.04 & 3.87 & 59.7 & 5.21 & 3.41 & 33.3 & 5.99 & 3.27 & 11.1 & 6.83 & 4.74 \\
\rowcolor{lightblue} \ours~\UTIL & 33.9 & 16.78 & 6.86 & 4.58 & 59.7 & 5.44 & 2.93 & 33.3 & 7.13 & 4.26 & 11.1 & 7.97 & 6.38 \\
\rowcolor{lightblue} \ours~\UTIL\COMP & 27.4 & 17.89 & 6.37 & 4.44 & 40.4 & 5.18 & 2.40 & 35.4 & 6.84 & 5.03 & 9.5 & 7.09 & 5.82 \\
\bottomrule
\end{tabular}
}
\end{table*}

\begin{table*}[t]
\centering
\caption{Detailed results on \textbf{OfficeBench} benchmark with gpt-4.1-mini. We adopt the same compression guidelines as those used in the gpt-4.1 experiments. We report accuracy (\%), and efficiency metrics: average steps, peak input tokens ($10^3$), and dependency ($10^6$) for Average and each difficulty level. Rows in blue background indicate the results from \colorbox{lightblue}{\textbf{ours}}.}
\label{tab:appendix:4_officebench_mini}
\resizebox{\textwidth}{!}{
\begin{tabular}{lccccccccccccc}
\toprule
& \multicolumn{4}{c}{\textbf{Average (All)}} & \multicolumn{3}{c}{\textbf{Level 1 (1-app, 42)}} & \multicolumn{3}{c}{\textbf{Level 2 (2-app, 22)}} & \multicolumn{3}{c}{\textbf{Level 3 (3-app, 31)}} \\
\cmidrule(lr){2-5} \cmidrule(lr){6-8} \cmidrule(lr){9-11} \cmidrule(lr){12-14}
\textbf{Method} & Acc. $\uparrow$ & Steps $\downarrow$ & Peak $\downarrow$ & Dep. $\downarrow$
& Acc. $\uparrow$ & Peak $\downarrow$ & Dep. $\downarrow$
& Acc. $\uparrow$ & Peak $\downarrow$ & Dep. $\downarrow$
& Acc. $\uparrow$ & Peak $\downarrow$ & Dep. $\downarrow$ \\
\midrule
\rowcolor{lightgray} \multicolumn{14}{c}{\textbf{Agent:} \texttt{gpt-4.1-mini} / \textbf{Compressor:} \texttt{gpt-4.1-mini}} \\
\midrule
No Compression                     & 72.63 & 11.96 & 7.36 & 3.92 & 88.10 & 6.66 & 4.29 & 68.18 & 4.97 & 1.01 & 54.84 & 9.02 & 5.40 \\
\midrule
\multicolumn{14}{l}{\textbf{History Compression}} \\
FIFO                  & 65.26 & 10.91 & 4.03 & 1.46 & 83.33 & 4.10 & 0.78 & 59.09 & 3.69 & 0.96 & 45.16 & 4.19 & 2.03 \\
Retrieval             & 67.37 & 14.46 & 4.55 & 2.74 & 85.71 & 5.85 & 5.86 & 59.09 & 3.47 & 0.87 & 48.39 & 4.59 & 2.45 \\
LLMLingua             & 67.39 & 11.59 & 4.90 & 2.18 & 87.18 & 4.31 & 3.87 & 59.09 & 4.58 & 0.92 & 48.39 & 5.34 & 2.17 \\
Prompting             & 71.58 & 11.78 & 4.93 & 3.10 & 85.71 & 4.73 & 4.75 & 72.73 & 4.40 & 0.86 & 51.61 & 5.32 & 3.06 \\
\rowcolor{lightblue} \ours & 73.68 & 12.41 & 4.82 & 1.96 & 88.10 & 4.12 & 0.83 & 68.18 & 4.39 & 0.86 & 58.06 & 5.37 & 3.07 \\
\midrule
\multicolumn{14}{l}{\textbf{Observation Compression}} \\
LLMLingua             & 66.32 & 11.02 & 6.34 & 2.40 & 78.57 & 6.09 & 2.12 & 63.64 & 4.82 & 0.97 & 51.61 & 7.30 & 3.34 \\
Prompting             & 73.68 & 11.43 & 6.45 & 2.62 & 88.10 & 4.82 & 1.44 & 72.73 & 4.95 & 1.06 & 54.84 & 8.01 & 4.01 \\
\rowcolor{lightblue} \ours & 71.58 & 10.96 & 6.00 & 2.19 & 88.10 & 4.45 & 1.06 & 63.64 & 4.89 & 1.00 & 54.84 & 7.30 & 3.36 \\
\bottomrule
\end{tabular}
}
\end{table*}

\begin{table*}[t]
\centering
\caption{Results on \textbf{8-objective QA} benchmark with gpt-4.1-mini. We adopt the same compression guidelines as those used in the gpt-4.1 experiments. We report EM/F1 and efficiency metrics (Steps, Peak input tokens ($10^3$), and Dependency ($10^6$)).}
\label{tab:appendix:3_qa_mini}
\centering
\resizebox{0.5\textwidth}{!}{
\setlength{\tabcolsep}{7pt} 
\begin{tabular}{lccccc}
\toprule
Method & EM $\uparrow$ & F1 $\uparrow$ & Steps $\downarrow$ & Peak $\downarrow$ & Dep. $\downarrow$ \\
\midrule
\rowcolor{lightgray} \multicolumn{6}{c}{\textbf{Agent:} \texttt{gpt-4.1-mini} / \textbf{Compressor:} \texttt{gpt-4.1-mini}} \\
\midrule
No compression                 & 0.330 & 0.436 & 19.80 & 12.93 & 5.63 \\
\midrule
\multicolumn{6}{l}{\bf History Compression} \\
FIFO                           & 0.024 & 0.031 & 28.45 & 5.33  & 3.89 \\
Retrieval                      & 0.143 & 0.190 & 26.90 & 5.34  & 3.55 \\
LLMLingua                      & 0.140 & 0.194 & 25.24 & 6.69  & 3.92 \\
Prompting                      & 0.149 & 0.207 & 25.27 & 4.85  & 2.44 \\
\rowcolor{lightblue} \ours      & 0.238 & 0.325 & 21.05 & 4.78  & 2.03 \\
\rowcolor{lightblue} \ours (iter2) & 0.248 & 0.353 & 19.18 & 4.79 & 1.79 \\
\midrule
\multicolumn{6}{l}{\bf Observation Compression} \\
LLMLingua                      & 0.316 & 0.430 & 15.96 & 5.54  & 1.60 \\
Prompting                      & 0.282 & 0.402 & 11.71 & 3.91  & 0.65 \\
\rowcolor{lightblue} \ours      & 0.323 & 0.434 & 14.42 & 4.71  & 1.10 \\
\rowcolor{lightblue} \ours (iter2) & 0.316 & 0.443 & 11.69 & 3.97 & 0.63 \\
\bottomrule
\end{tabular}
}
\end{table*}

\begin{table*}[t]
\centering
\caption{Results across different difficulty levels on \textbf{AppWorld} benchmark (test-normal) with gpt-5-chat. We adopt the same compression guidelines as those used in the gpt-4.1 experiments. Each block reports accuracy (task goal completion score), steps, peak input tokens ($10^3$), and dependency ($10^6$) for agents. Best results in each column are highlighted in bold. Rows in blue background indicate the results from \colorbox{lightblue}{\textbf{ours}}.}
\label{tab:appendix:1_appworld_gpt5}
\resizebox{\textwidth}{!}{
\begin{tabular}{lccccccccccccc}
\toprule
\multirow{2}{*}{Method} & 
\multicolumn{4}{c}{Average (168)} & 
\multicolumn{3}{c}{Easy (57)} & 
\multicolumn{3}{c}{Medium (48)} & 
\multicolumn{3}{c}{Hard (63)} \\
\cmidrule(lr){2-5} \cmidrule(lr){6-8} \cmidrule(lr){9-11} \cmidrule(lr){12-14}
& Acc. $\uparrow$ & Steps $\downarrow$ & Peak $\downarrow$ & Dep.$\downarrow$
& Acc. $\uparrow$ & Peak $\downarrow$ & Dep.$\downarrow$
& Acc. $\uparrow$ & Peak $\downarrow$ & Dep.$\downarrow$
& Acc. $\uparrow$ & Peak $\downarrow$ & Dep.$\downarrow$ \\
\midrule
\rowcolor{lightgray}\multicolumn{14}{c}{\textbf{Agent:} \texttt{gpt-5-chat} / \textbf{Compressor:} \texttt{gpt-5-chat}}\\
\midrule
No compression & 66.7 & 16.45 & 9.67 & 4.78 & 89.5 & 7.55 & 2.31 & 64.6 & 9.58 & 4.13 & 47.6 & 11.67 & 7.51 \\
\midrule[0.1pt]
\multicolumn{14}{l}{\textbf{History Compression}}\\
FIFO (last-5) & 46.4 & 30.61 & 6.81 & 4.85 & 79.0 & 5.21 & 2.10 & 43.8 & 6.82 & 5.50 & 19.1 & 8.24 & 6.84 \\
\rowcolor{lightblue} Prompting & 58.9 & 22.24 & 7.46 & 4.02 & 82.5 & 7.15 & 2.13 & 66.7 & 7.19 & 3.69 & 31.8 & 7.93 & 5.97 \\
\rowcolor{lightblue} \ours~\UTIL & 58.3 & 20.15 & 6.97 & 3.74 & 80.7 & 6.66 & 2.04 & 66.7 & 7.08 & 3.40 & 31.8 & 7.16 & 5.54 \\
\rowcolor{lightblue} \ours~\UTIL\COMP & 62.5 & 22.29 & 7.26 & 3.85 & 86.0 & 6.44 & 2.04 & 72.9 & 6.98 & 3.93 & 33.3 & 8.20 & 5.42 \\
\midrule[0.1pt]
\multicolumn{14}{l}{\textbf{Observation Compression}}\\
\rowcolor{lightblue} Prompting & 60.1 & 17.39 & 6.50 & 3.72 & 80.7 & 4.98 & 1.72 & 68.8 & 6.40 & 3.48 & 34.9 & 7.96 & 5.70 \\
\rowcolor{lightblue} \ours~\UTIL & 65.5 & 17.16 & 7.58 & 3.96 & 84.2 & 5.62 & 1.94 & 68.8 & 7.49 & 3.46 & 46.0 & 9.41 & 6.16 \\
\rowcolor{lightblue} \ours~\UTIL\COMP & 62.5 & 18.21 & 7.21 & 4.24 & 80.7 & 5.52 & 2.02 & 70.8 & 7.18 & 3.69 & 39.7 & 8.76 & 6.67 \\
\midrule[0.1pt]
\multicolumn{14}{l}{\textbf{History + Observation Compression}}\\
\rowcolor{lightblue} \ours~\UTIL & 63.1 & 20.02 & 5.89 & 3.63 & 77.2 & 5.27 & 1.92 & 77.1 & 6.03 & 3.52 & 39.7 & 6.35 & 5.28 \\
\rowcolor{lightblue} \ours~\UTIL\COMP & 58.9 & 22.90 & 5.83 & 4.07 & 80.7 & 5.35 & 1.94 & 77.1 & 5.94 & 3.56 & 25.4 & 6.17 & 6.39 \\
\bottomrule
\end{tabular}
}
\end{table*}

\begin{table*}[t]
\centering
\caption{Results across different difficulty levels on \textbf{AppWorld} with \textbf{distilled compressors}. We report accuracy (task goal completion score), average steps, peak input tokens ($10^3$), and dependency ($10^6$). For all compressors, we use the optimized compression guideline after the utilization maximization~\UTIL step. `Fine-tune' means that we fine-tune small models with outputs from naive prompt before compression guideline optimization.}
\label{tab:appendix:5_appworld_distill_opt}
\resizebox{\textwidth}{!}{
\begin{tabular}{lccccccccccccc}
\toprule
\multirow{2}{*}{Method} & 
\multicolumn{4}{c}{Average} & 
\multicolumn{3}{c}{Easy} & 
\multicolumn{3}{c}{Medium} & 
\multicolumn{3}{c}{Hard} \\
\cmidrule(lr){2-5} \cmidrule(lr){6-8} \cmidrule(lr){9-11} \cmidrule(lr){12-14}
& Acc. $\uparrow$ & Steps $\downarrow$ & Peak $\downarrow$ & Dep.$\downarrow$
& Acc. $\uparrow$ & Peak $\downarrow$ & Dep.$\downarrow$
& Acc. $\uparrow$ & Peak $\downarrow$ & Dep.$\downarrow$
& Acc. $\uparrow$ & Peak $\downarrow$ & Dep.$\downarrow$ \\
\midrule
\rowcolor{lightgray} \multicolumn{14}{c}{\textbf{Agent:} \texttt{gpt-4.1} / \textbf{Compressor:} \texttt{gpt-4.1-mini} or Distilled models (\texttt{Qwen3}, \texttt{Phi-4})} \\
\midrule
\multicolumn{14}{l}{\textbf{History Compression}}\\
Prompting (\texttt{gpt-4.1-mini}) & 39.3 & 23.61 & 7.03 & 5.19 & 64.9 & 6.64 & 3.17 & 35.4 & 7.63 & 5.42 & 19.1 & 6.93 & 6.84 \\
\ours (\texttt{gpt-4.1-mini}) & 47.6 & 21.46 & 7.25 & 5.24 & 75.4 & 6.75 & 2.84 & 35.4 & 7.25 & 5.36 & 31.8 & 7.70 & 7.32 \\
Fine-tune (\texttt{Qwen3-14B}) & 44.6 & 24.16 & 7.16 & 4.95 & 71.9 & 6.79 & 2.88 & 43.8 & 7.39 & 4.88 & 20.6 & 7.33 & 6.88 \\
\ours (\texttt{Qwen3-14B}) & 50.0 & 21.72 & 6.83 & 4.80 & 79.0 & 6.42 & 2.54 & 50.0 & 6.87 & 4.89 & 23.8 & 7.17 & 6.79 \\
\ours (\texttt{Qwen3-8B}) & 47.0 & 21.58 & 6.98 & 4.76 & 71.9 & 6.64 & 2.93 & 37.5 & 7.24 & 4.67 & 31.8 & 7.09 & 6.48 \\
\ours (\texttt{Phi-4}) & 44.6 & 21.19 & 7.24 & 4.76 & 68.4 & 7.33 & 2.75 & 39.6 & 7.12 & 4.16 & 27.0 & 7.26 & 7.04 \\
\midrule
\multicolumn{14}{l}{\textbf{Observation Compression}}\\
Prompting (\texttt{gpt-4.1-mini}) & 44.0 & 16.67 & 6.84 & 4.30 & 71.9 & 5.08 & 2.19 & 35.4 & 6.72 & 3.77 & 25.4 & 8.53 & 6.61 \\
\ours (\texttt{gpt-4.1-mini}) & 48.2 & 18.00 & 8.66 & 6.62 & 71.9 & 6.05 & 2.60 & 37.5 & 9.23 & 7.41 & 34.9 & 10.60 & 9.65 \\
Fine-tune (\texttt{Qwen3-14B}) & 40.5 & 17.71 & 6.64 & 4.38 & 64.9 & 4.91 & 1.97 & 31.2 & 6.72 & 4.05 & 25.4 & 8.16 & 6.81 \\
\ours (\texttt{Qwen3-14B}) & 56.5 & 16.78 & 7.57 & 5.06 & 82.5 & 5.69 & 2.20 & 54.2 & 7.39 & 4.46 & 34.9 & 9.40 & 8.10 \\
\ours (\texttt{Qwen3-8B}) & 48.2 & 16.10 & 7.33 & 4.82 & 71.9 & 5.49 & 2.03 & 50.0 & 7.20 & 4.20 & 25.4 & 9.10 & 7.82 \\
\ours (\texttt{Phi-4}) & 50.6 & 16.88 & 7.88 & 5.41 & 77.2 & 5.85 & 2.88 & 52.1 & 7.75 & 4.77 & 25.4 & 9.83 & 8.18 \\
\bottomrule
\end{tabular}
}
\end{table*}

\begin{table*}[t]
\centering
\caption{Additional results for additional guideline optimization step and unified compression on \textbf{Appworld} benchmark (test-normal). Each block reports accuracy (task goal completion score), steps, peak input tokens ($10^3$), and dependency ($10^6$) for agents. Best results in each column are highlighted in bold. Rows in blue background indicate the results from \colorbox{lightblue}{\textbf{ours}}.}
\label{tab:appendix:1_appworld_additional}
\resizebox{\textwidth}{!}{
\begin{tabular}{lccccccccccccc}
\toprule
\multirow{2}{*}{Method} & 
\multicolumn{4}{c}{Average (168)} & 
\multicolumn{3}{c}{Easy (57)} & 
\multicolumn{3}{c}{Medium (48)} & 
\multicolumn{3}{c}{Hard (63)} \\
\cmidrule(lr){2-5} \cmidrule(lr){6-8} \cmidrule(lr){9-11} \cmidrule(lr){12-14}
& Acc. $\uparrow$ & Steps $\downarrow$ & Peak $\downarrow$ & Dep.$\downarrow$
& Acc. $\uparrow$ & Peak $\downarrow$ & Dep.$\downarrow$
& Acc. $\uparrow$ & Peak $\downarrow$ & Dep.$\downarrow$
& Acc. $\uparrow$ & Peak $\downarrow$ & Dep.$\downarrow$ \\
\midrule
\rowcolor{lightgray}\multicolumn{14}{c}{\textbf{Agent:} \texttt{gpt-4.1} / \textbf{Compressor:} \texttt{gpt-4.1}}\\
\midrule
\multicolumn{14}{l}{\textbf{History Compression}} \\
\rowcolor{lightblue} \ours~\UTIL\COMP\UTIL & 47.0 & 22.28 & 7.22 & 4.66 
& 68.4 & 7.01 & 2.69 & 58.3 & 7.16 & 4.39  & 19.1 & 7.45 & 6.65 \\
\midrule[0.1pt]
\multicolumn{14}{l}{\textbf{History + Observation Compression}}\\
Prompting & 36.3 & 19.33 & 5.38 & 3.44 & 71.9 & 4.87 & 1.80 & 21.6 & 5.63 & 3.60 & 14.3 & \bf 5.64 & 4.79 \\
\rowcolor{lightblue} \ours & 45.8 & 20.32 & 5.85 & 4.26 & 75.4 & 5.29 & 2.07 & 39.6 & 6.15 & 4.29 & 23.8 & 6.12 & 6.21 \\
\rowcolor{lightblue} \ours~\UTIL\COMP & 44.6 & 21.75 & 5.90 & 4.98 & 77.2 & 5.50 & 2.33 & 39.6 & 6.18 & 3.80 & 19.1 & 6.18 & 8.28 \\
\bottomrule
\end{tabular}
}
\end{table*}

\clearpage

\newtcblisting[auto counter, number within=section]{promptbox}[2][]{
  colback=white,
  colframe=blue!80!black,
  coltitle=white,
  title=Prompt~\thetcbcounter: #2,
  fonttitle=\bfseries\normalsize,
  boxrule=1pt,
  arc=2mm,
  top=2mm,
  bottom=2mm,
  width=0.95\linewidth,
  breakable,
  listing only,
  listing engine=listings,
  listing options={
    basicstyle=\ttfamily\small,
    breaklines=true,
    columns=fullflexible,
    keepspaces=true
  },
  #1
}

\begin{promptbox}[label={box:prompt_optimizer_analysis}]{Prompt for analysis before prompt optimization (utility step)}
You are an expert agent trajectory auditor.

Analyze why the HISTORY-OPTIMIZED agent failed OR became significantly less efficient while the BASELINE succeeded.
You are given:
- task_name: {{ task_name }}
- Baseline full history (single continuous session)
- Optimized history split into multiple sessions where each new session starts with a fresh system + user prompt and an injected <HISTORY_SUMMARY> summarizing earlier interactions.
- baseline_success={{ baseline_success }} optimized_success={{ optimized_success }}
- baseline_env_steps={{ baseline_env_steps | default('null') }} optimized_env_steps={{ optimized_env_steps | default('null') }} step_ratio={{ step_ratio | default('null') }} performance_regression={{ performance_regression | default('false') }}

Goals:
1. Determine whether summarization / session resetting removed, distorted, delayed, or bloated reasoning causing failure OR inflated step count (> threshold factor of baseline).
2. Identify the FIRST divergence point where the optimized trajectory meaningfully deviates from the successful & efficient baseline path.
3. Categorize root causes (e.g., Missing Critical Fact, Incorrect Summary, Lost Variable/State, Unnecessary Re-discovery, Instruction Drift, API Misuse, Premature Completion, Token Truncation, Inefficient Looping, Redundant API Calls, Over-Exploration, Other).
4. Extract concrete evidence snippets (quote exact lines) from baseline vs optimized showing:
   - Critical facts present in baseline but absent/altered in optimized (esp. after a session boundary)
   - Summary inaccuracies (baseline ground truth vs summary text)
   - Redundant or looping action patterns causing step inflation.
5. Suggest precise remediation strategies: summary style changes, retain variable/value tables, move session boundaries, guardrail prompts, caching, early-exit heuristics, loop detection, etc.
6. Provide a reliability_score (0.0-1.0) reflecting confidence in your causal attribution.
7. If performance_regression==true, analyze efficiency degradation even if optimized_success==true.

Output STRICTLY valid JSON object with keys:
{
  "task_name": str,
  "divergence_step_description": str,
  "root_cause_categories": [str, ...],
  "missing_or_distorted_facts": [ {"baseline": str, "optimized_context_absent_or_changed": str, "impact": str} ],
  "summary_inaccuracies": [ {"summary_excerpt": str, "issue_type": str, "correct_baseline_reference": str, "impact": str} ],
  "lost_state_variables": [ {"name_or_pattern": str, "baseline_evidence": str, "optimized_issue": str} ],
  "api_or_action_errors": [ {"optimized_step_excerpt": str, "error_type": str, "improvement": str} ],
  "inefficiency_patterns": [ {"pattern": str, "evidence_excerpt": str, "excess_steps": int, "cause": str, "remediation": str} ],
  "timeline_of_divergence": [ {"phase": str, "optimized_excerpt": str, "baseline_contrast": str, "effect": str} ],
  "performance_regression": bool,
  "baseline_env_steps": int | null,
  "optimized_env_steps": int | null,
  "step_ratio": float | null,
  "remediation_recommendations": [ str, ... ],
  "recovery_opportunities_missed": [ {"optimized_excerpt": str, "missed_fix_action": str} ],
  "reliability_score": float,
  "concise_failure_mechanism_summary": str
}

If some sections have no data, use an empty list. For non-applicable numeric fields use null.
Do NOT include any extra commentary outside JSON.

---
BASELINE_HISTORY_START
{{ baseline_history }}
BASELINE_HISTORY_END

OPTIMIZED_MULTI_SESSION_HISTORY_START
{{ optimized_history }}
OPTIMIZED_MULTI_SESSION_HISTORY_END

Failure or performance report / metadata (may be null):
{{ failure_report }}

Proceed with rigorous comparison.
\end{promptbox}
\begin{promptbox}[label={box:prompt_optimizer_update}]{Prompt for prompt optimization after analysis (utility step)}
You are an expert prompt engineer tasked with refining a HISTORY SUMMARIZATION prompt.
Rewrite the ORIGINAL PROMPT to reduce length of the HISTORY SUMMARY while preserving factual continuity for the next session.
Ground all changes in the PER-SAMPLE REDUCTION SIGNALS below. Do not aggregate across samples; use the patterns and rules as-is.

Constraints:
- Keep all Jinja placeholders, variable names, and structure intact where possible.
- Add explicit, concrete rules that prevent verbosity and retain essential state.
- Do not include literal values from prior content; refer to variable names only.
- Output ONLY the improved prompt template (no extra commentary).

Context (samples below are the only ground truth signals to use):
- Average original summary size (chars) across sampled set: {{ avg_orig_chars }}

{% for s in samples %}
===== SAMPLE {{ loop.index0 }} =====
- Task/Session: {{ s.task_label }} / {{ s.session or 'unknown-session' }}
- Analysis Overview:
{% if s.overview %}
{% for k, v in s.overview.items() %}  - {{ k }}: {{ v }}
{% endfor %}
{% else %}  - (none provided)
{% endif %}

- Removals (patterns -> action):
{% for r in s.removals %}  - [{{ r.category | default('unknown') }}] {{ r.pattern | default('') }} -> {{ r.action | default('drop') }}
{% endfor %}

- KEEP examples (evidence-driven essentials):
{% for k in s.keeps %}  - Reason: {{ k.reason | default('') }} | Evidence: {{ k.evidence_spans | default([]) | join('; ') }}
{% endfor %}

- Summary Rules:
{% for rule in s.rules %}  - {{ rule }}
{% endfor %}

{% endfor %}

Original Prompt Template (verbatim between markers):
<<<ORIGINAL_PROMPT>>>
{{ original_prompt }}
<<<ORIGINAL_PROMPT>>>

Output only the improved prompt template text, ready to be used as a Jinja template.
\end{promptbox}

\begin{promptbox}[label={box:prompt_optimizer_analysis_compression}]{Prompt for analysis before prompt optimization (compression step)}
You are an expert prompt engineer tasked with refining a HISTORY SUMMARIZATION prompt.
Rewrite the ORIGINAL PROMPT to reduce length of the HISTORY SUMMARY while preserving factual continuity for the next session.
Ground all changes in the PER-SAMPLE REDUCTION SIGNALS below. Do not aggregate across samples; use the patterns and rules as-is.

Constraints:
- Keep all Jinja placeholders, variable names, and structure intact where possible.
- Add explicit, concrete rules that prevent verbosity and retain essential state.
- Do not include literal values from prior content; refer to variable names only.
- Output ONLY the improved prompt template (no extra commentary).

Context (samples below are the only ground truth signals to use):
- Average original summary size (chars) across sampled set: {{ avg_orig_chars }}

{% for s in samples %}
===== SAMPLE {{ loop.index0 }} =====
- Task/Session: {{ s.task_label }} / {{ s.session or 'unknown-session' }}
- Analysis Overview:
{% if s.overview %}
{% for k, v in s.overview.items() %}  - {{ k }}: {{ v }}
{% endfor %}
{% else %}  - (none provided)
{% endif %}

- Removals (patterns -> action):
{% for r in s.removals %}  - [{{ r.category | default('unknown') }}] {{ r.pattern | default('') }} -> {{ r.action | default('drop') }}
{% endfor %}

- KEEP examples (evidence-driven essentials):
{% for k in s.keeps %}  - Reason: {{ k.reason | default('') }} | Evidence: {{ k.evidence_spans | default([]) | join('; ') }}
{% endfor %}

- Summary Rules:
{% for rule in s.rules %}  - {{ rule }}
{% endfor %}

{% endfor %}

Original Prompt Template (verbatim between markers):
<<<ORIGINAL_PROMPT>>>
{{ original_prompt }}
<<<ORIGINAL_PROMPT>>>

Output only the improved prompt template text, ready to be used as a Jinja template.
\end{promptbox}
\begin{promptbox}[label={box:prompt_optimizer_update_compression}]{Prompt for analysis before prompt optimization (compression step)}
You are an expert prompt engineer tasked with refining a HISTORY SUMMARIZATION prompt.
Rewrite the ORIGINAL PROMPT to reduce length of the HISTORY SUMMARY while preserving factual continuity for the next session.
Ground all changes in the PER-SAMPLE REDUCTION SIGNALS below. Do not aggregate across samples; use the patterns and rules as-is.

Constraints:
- Keep all Jinja placeholders, variable names, and structure intact where possible.
- Add explicit, concrete rules that prevent verbosity and retain essential state.
- Do not include literal values from prior content; refer to variable names only.
- Output ONLY the improved prompt template (no extra commentary).

Context (samples below are the only ground truth signals to use):
- Average original summary size (chars) across sampled set: {{ avg_orig_chars }}

{% for s in samples %}
===== SAMPLE {{ loop.index0 }} =====
- Task/Session: {{ s.task_label }} / {{ s.session or 'unknown-session' }}
- Analysis Overview:
{% if s.overview %}
{% for k, v in s.overview.items() %}  - {{ k }}: {{ v }}
{% endfor %}
{% else %}  - (none provided)
{% endif %}

- Removals (patterns -> action):
{% for r in s.removals %}  - [{{ r.category | default('unknown') }}] {{ r.pattern | default('') }} -> {{ r.action | default('drop') }}
{% endfor %}

- KEEP examples (evidence-driven essentials):
{% for k in s.keeps %}  - Reason: {{ k.reason | default('') }} | Evidence: {{ k.evidence_spans | default([]) | join('; ') }}
{% endfor %}

- Summary Rules:
{% for rule in s.rules %}  - {{ rule }}
{% endfor %}

{% endfor %}

Original Prompt Template (verbatim between markers):
<<<ORIGINAL_PROMPT>>>
{{ original_prompt }}
<<<ORIGINAL_PROMPT>>>

Output only the improved prompt template text, ready to be used as a Jinja template.

\end{promptbox}
\begin{promptbox}[label={box:appworld_history_compression}]{AppWorld Prompt for history compression before optimization}
You are maintaining a structured context-aware summary for a productivity agent. You will be given the user instruction for the agent, a list of interactions corresponding to actions taken by the agent, and the most recent previous summary if one exists. Produce the following:

### REASONING
Summarize key progress, decisions made, important observed outcomes, and rationale behind actions taken so far. Include how earlier steps influenced later ones and why certain data is retained in the summary.

### COMPLETED
List completed subtasks or successful outcomes, with brief results if applicable.

---

## [Information Source]

### USER INSTRUCTION

{{ task }}

## [PREVIOUS SUMMARY] (if any)

{{ prev_summary }}

## [HISTORY OF INTERACTIONS]

{{ history }}

---

## PRIORITIZE

1. Keep all sections relevant and concise.  
2. Use reusable structured formats when summarizing artifacts.  
3. Ensure agent can resume task with no loss of information.
4. Include key info from errors or failed attempts to prevent repeated mistakes.
5. Preserve all essential artifacts and data needed to complete the task.

---

### [Output Format]

Do **not** include the input or any additional explanation. Only return the formatted summary.
\end{promptbox}
\begin{promptbox}[label={box:appworld_history_optimized}]{AppWorld Prompt for history compression after optimization (\UTIL)}
You maintain a compact, state-preserving HISTORY_SUMMARY for a multi-session agent.

Input:
[USER INSTRUCTION] {{ task }}
[PREVIOUS SUMMARY] {{ prev_summary }}
[HISTORY OF INTERACTIONS] {{ history }}

Create the following sections-use the exact headings and order:

<HISTORY_SUMMARY>

1. REASONING  
   - Key progress, decisions, outcomes, and their rationale.  
   - Note how earlier steps influence later ones.

2. VARS  
   | name | value | purpose |  
   |------|-------|---------|  
   Record every runtime value the next session must re-declare (tokens, ids, lists, last page_index/page_limit, etc.).  

3. TODO  
   List pending actions with enough detail to execute directly.

4. COMPLETED  
   Bullet list of finished subtasks with brief results.

5. GUARDRAILS  
   Short reminders that prevent repeat errors, e.g.  
   - Memory resets; re-create VARS before use.  
   - Paginate until empty page.  
   - Validate API parameters against spec.  
   - Avoid redundant logins or doc look-ups.

Requirements:
- Be concise-bullets and tables preferred; no extraneous prose.  
- Preserve all essential facts, parameters, and artifacts; omit nothing critical.  
- Include errors only if they inform future avoidance.  
- Do not output the input or any commentary-return only <HISTORY_SUMMARY>.
\end{promptbox}
\begin{promptbox}[label={box:appworld_history_optimized_len}]{AppWorld Prompt for history compression after optimization\\(\UTIL\COMP)}
You maintain a compact, state-preserving HISTORY_SUMMARY for a multi-session agent.

Input:
[USER INSTRUCTION] {{ task }}
[PREVIOUS SUMMARY] {{ prev_summary }}
[HISTORY OF INTERACTIONS] {{ history }}

Summary Compression Rules:
- Collapse multi-bullet narratives into <=2 concise sentences.
- Replace repetitive step logs with one summarizing phrase.
- Truncate long token/credential strings to "<token>" unless verbatim reuse is required.
- Remove unused/expired credentials, page_index/page_limit, verbose API dumps, and table borders.
- Shrink GUARDRAILS to one bullet unless multiple items are still critical.
- Delete tool/API log output, greetings, meta prose, and section headers that no longer contain content.
- Keep only variables actively referenced in upcoming steps; list each once in VARS.
- Reference removal categories [repetition], [tool-logs], [meta], [formatting] to prune similar lines.
- Preserve factual continuity; never invent or alter state variables.
- Target summaries well under {{ max_chars | default(1500) }} characters.

Critical Essentials:
Always keep evidence-driven items required next session (e.g., tokens, ids, emails, amounts, lists, paths, description strings, brief task status).

Output EXACTLY the following structure---nothing more:

<HISTORY_SUMMARY>

1. REASONING
   One brief paragraph on key progress and rationale.

2. VARS
   key=value pairs, comma-separated; only still-needed runtime values.

3. TODO
   Bulleted next actions (<=5).

4. COMPLETED
   Bulleted finished subtasks (<=5).

5. GUARDRAILS
   Single concise bullet, or omit if none.

Return only the <HISTORY_SUMMARY> block---no additional commentary or input echoes.
\end{promptbox}
\begin{promptbox}[label={box:appworld_obs_compression}]{AppWorld Prompt for observation compression before optimization}
Your task is to generate a "Reasoning" and a "Refined Observation" based on the inputs below.

In the "Reasoning", analyze the user instruction and history to identify what information from the current observation is necessary to complete the remaining steps.  
Think about what parts can be summarized or transformed to reduce length, while ensuring that future actions can still be executed based on the refined observation alone.

In the "Refined Observation", include only the information that is minimal but sufficient for the next steps.

[Information source]
# User Instruction
{{ task }}

# History of interactions
{{ history }}

# Observation at the current time step
{{ observation }}

[Output format]
# Reasoning
... your reasoning for what matters and how to optimize it ...
# Refined Observation
... reduced and actionable observation ...
\end{promptbox}
\begin{promptbox}[label={box:appworld_obs_optimized}]{AppWorld Prompt for observation compression after optimization (\UTIL)}
Your task: write two sections---"Reasoning" and "Refined Observation".

1. Reasoning
   - Examine task, history, and observation.
   - Decide exactly which parts of the observation must be kept so the next agent step can succeed.
   - Note any need to paginate (page_limit default = 5, page_index).
   - Justify any data you drop.

2. Refined Observation
   - Contain only the minimal yet sufficient info for the next step.
   - Always preserve:
     - Every endpoint that may be called, plus its full parameter list and defaults (especially page_limit/page_index, auth tokens).
     - Response-schema fields referenced or likely needed later (e.g., play_count, release_date, like_count, position, ids).
     - Raw data rows required for future comparisons or loops; if summarising, keep at least all positive-match examples.
   - Never:
     - Omit defaults that affect behaviour.
     - Declare parameters "not critical" without proof.
     - Hallucinate endpoints or fields.
     - Replace machine-readable data with vague prose.

[Information source]
# User Instruction
{{ task }}

# History of interactions
{{ history }}

# Observation at the current time step
{{ observation }}

[Output format]
# Reasoning
...concise analysis explaining what is kept/removed...

# Refined Observation
...trimmed yet complete observation ensuring future steps remain possible...

\end{promptbox}
\begin{promptbox}[label={box:appworld_obs_optimized_len}]{AppWorld Prompt for observation compression after optimization\\(\UTIL\COMP)}
Your task: create two sections---"Reasoning" and "Refined Observation".

1. Reasoning (<=40 words)
   - Briefly state what was kept and why; note dropped categories and any pagination needs.

2. Refined Observation (use ONLY the current observation)
   - Keep strictly necessary data for the next step.
   - Always preserve, when present:
     - access_token or other auth values
     - page_limit, page_index and other defaults that affect calls
     - every endpoint name; include parameters only if required; description <= 4 words
     - identifiers/fields needed for comparisons, loops, or API calls (ids, titles, counts, paths, etc.)
   - Minimise length:
     - Delete unused fields and narrative text.
     - Compress JSON/arrays (one object per line, no extra spaces).
     - Summarise long uniform lists with a range/pattern when individual rows are not needed.
   - Never invent, alter, or omit a required literal.
   - Exclude history, prior summaries, and meta comments.

[Information source]
# User Instruction
{{ task }}

# History of interactions
{{ history }}

# Observation at the current time step
{{ observation }}

[Output format]
# Reasoning
...

# Refined Observation
...
\end{promptbox}

\clearpage
\begin{examplebox}[label={ex:appworld}]{AppWorld task trajectory example}
\textbf{Problem:} Name the artist least recommended to me on Spotify.

\rule{\linewidth}{0.4pt}

\begin{lstlisting}{python}
# First, let's check what APIs are available under the Spotify app to see 
if there is an API related to recommendations or artists.
print(apis.api_docs.show_api_descriptions(app_name='spotify'))
\end{lstlisting}

\begin{observationbox}
\begin{lstlisting}{json}
[{"name":"login",...},{"name":"show_recommendations","description":
"Get personalized song recommendations for the user."}, ...]
\end{lstlisting}
\end{observationbox}

\begin{lstlisting}{python}
# There is an API called "show_recommendations" which provides personalized 
song recommendations. Let's check its detailed specification...
print(apis.api_docs.show_api_doc(app_name='spotify',
api_name='show_recommendations'))
\end{lstlisting}

\begin{observationbox}
\begin{lstlisting}{json}
{"path":"/recommendations","method":"GET","parameters":
[{"name":"access_token","required":true}, 
{"name":"page_index",...},{"name":"page_limit",...}], ...}
\end{lstlisting}
\end{observationbox}

\begin{lstlisting}{python}
# To call `show_recommendations`, I need an access token. Check the login spec.
print(apis.api_docs.show_api_doc(app_name='spotify', api_name='login'))
\end{lstlisting}

\begin{observationbox}
\begin{lstlisting}{json}
{"path":"/auth/token","method":"POST","parameters":[{"name":"username"},
{"name":"password"}], ...}
\end{lstlisting}
\end{observationbox}

\begin{lstlisting}{python}
# I need Timothy's Spotify account password to log in. 
# Let's get it from supervisor.
spotify_passwords = apis.supervisor.show_account_passwords()
spotify_password = next((acc["password"] for acc in spotify_passwords 
if acc["account_name"]=="spotify"), None)
print(spotify_password)
\end{lstlisting}

\begin{observationbox}
\begin{lstlisting}{text}
<REDACTED>
\end{lstlisting}
\end{observationbox}

\begin{lstlisting}{python}
# Now log in to get the access token.
spotify_login = apis.spotify.login(username="timothy.whit@gmail.com",
                                   password="<REDACTED>")
access_token = spotify_login.get("access_token")
print(access_token)
\end{lstlisting}

\begin{observationbox}
\begin{lstlisting}{text}
eyJhbGciOiJIUzI1NiIsInR5cCI6IkpXVCJ9...<REDACTED>...
\end{lstlisting}
\end{observationbox}

\begin{lstlisting}{python}
# Now get the first page of personalized song recommendations.
recommendations_page_0 = apis.spotify.show_recommendations(
    access_token=access_token, page_index=0, page_limit=20
)
print(recommendations_page_0[:4])  # show a few items only
\end{lstlisting}

\begin{observationbox}
\begin{lstlisting}{json}
[
 {"song_id":10,"title":"The Curse of Loving You",
 "artists":[{"name":"Lucas Grey"}]},
 {"song_id":11,"title":"Lost in a Moment's Grace",
 "artists":[{"name":"Apollo Serenade"},{"name":"Phoenix Rivers"}]},
 {"song_id":40,"title":"Burning Like a Distant Star",
 "artists":[{"name":"Aria Sterling"}]},
 ...
]
\end{lstlisting}
\end{observationbox}

\begin{lstlisting}{python}
# To find the artist least recommended to Timothy, I will count how many times
each artist appears in the recommendations. The artist with the fewest 
appearances is the least recommended.
from collections import Counter

artist_names = [a["name"] for r in recommendations_page_0 for a in r["artists"]]
artist_counts = Counter(artist_names)
least_recommended_artist = min(artist_counts, key=artist_counts.get)
print(least_recommended_artist)
\end{lstlisting}

\begin{observationbox}
\begin{lstlisting}{text}
Evelyn Rose
\end{lstlisting}
\end{observationbox}

\begin{lstlisting}{python}
# Task complete. I will submit "Evelyn Rose" as the answer.
apis.supervisor.complete_task(answer="Evelyn Rose")
\end{lstlisting}

\end{examplebox}

\begin{examplebox}[label={ex:appworld_fail_no_compress}]{AppWorld trajectory example - gpt-4.1-mini agent fails without history compression}
\textbf{Problem:} Delete all \texttt{.pdf} files from \texttt{\textasciitilde{}/downloads}.

\rule{\linewidth}{0.4pt}

\begin{lstlisting}{python}

# Understand available APIs in the file_system app.

print(apis.api_docs.show_api_descriptions(app_name='file_system'))
\end{lstlisting}

\begin{observationbox}
\begin{lstlisting}{json}
[
{"name":"show_account","description":"Show your account information. 
Unlike show_profile, this includes private information."},
{"name":"signup","description":"Sign up to create account."},
{"name":"delete_account","description":"Delete your account."},
{"name":"update_account_name","description":"Update your first or last name 
in the account profile."},
...
{"name":"copy_directory","description":"Copy a directory to another 
location."},
{"name":"move_directory","description":"Move a directory to another 
location."},
{"name":"compress_directory","description":"Compress a directory."},
{"name":"decompress_file","description":"Decompress a compressed file."}
]
\end{lstlisting}
\end{observationbox}

\begin{lstlisting}{python}

# List files in ~/downloads to find PDFs.

downloads_path = '~/downloads'
downloads_files = apis.file_system.show_directory(path=downloads_path)
print(downloads_files)
\end{lstlisting}

\begin{observationbox}
\begin{lstlisting}{text}
Execution failed. Traceback:
...
Exception: Response status code is 401:
{"message":"You are either not authorized to access this file_system API 
endpoint or your access token is missing, invalid or expired."}
\end{lstlisting}
\end{observationbox}

\begin{lstlisting}{python}

# Need authentication. Check supervisor for stored credentials.

print(apis.supervisor.show_account_passwords())
\end{lstlisting}

\begin{observationbox}
\begin{lstlisting}{json}
[
{"account_name":"amazon","password":"Gt$!_*W"},
{"account_name":"file_system","password":"8nNw!jZ"},
{"account_name":"gmail","password":"qu4Y7}s"},
{"account_name":"phone","password":"C4n&I40"},
{"account_name":"simple_note","password":"e+QwbmV"},
{"account_name":"splitwise","password":"mSqG}QU"},
{"account_name":"spotify","password":"%CCvl8v"},
{"account_name":"todoist","password":"jHZ#RPM"},
{"account_name":"venmo","password":"Wq8!RAU"}
]
\end{lstlisting}
\end{observationbox}

\begin{lstlisting}{python}

# Try logging in with just the password (no username).

file_system_password = [a["password"] for a in 
apis.supervisor.show_account_passwords()
if a["account_name"] == "file_system"][0]
file_system_login_response = apis.file_system.login(password=
file_system_password)
print(file_system_login_response)
\end{lstlisting}

\begin{observationbox}
\begin{lstlisting}{text}
Execution failed. Traceback:
...
Exception: Response status code is 422:
{"message":"Validation error. Reason: \nusername: Field required"}
\end{lstlisting}
\end{observationbox}

\begin{lstlisting}{python}

# Need username. Check supervisor profile for email/username.

print(apis.supervisor.show_profile())
\end{lstlisting}

\begin{observationbox}
\begin{lstlisting}{json}
{"first_name":"Susan","last_name":"Burton","email":"[susanmiller@gmail.com]
(mailto:susanmiller@gmail.com)",
"phone_number":"3296062648","birthday":"1994-04-30","sex":"female"}
\end{lstlisting}
\end{observationbox}

\begin{lstlisting}{python}

# Attempt login with email as username.

file_system_username = "[susanmiller@gmail.com](mailto:susanmiller@gmail.com)"
file_system_password = [a["password"] for a in 
apis.supervisor.show_account_passwords()
if a["account_name"] == "file_system"][0]
file_system_login_response = apis.file_system.login(
username=file_system_username, password=file_system_password)
print(file_system_login_response)
\end{lstlisting}

\begin{observationbox}
\begin{lstlisting}{json}
{"access_token":"eyJhbGciOiJIUzI1NiIsInR5cCI6IkpXVCJ9...<REDACTED>...",
"token_type":"Bearer"}
\end{lstlisting}
\end{observationbox}

\begin{lstlisting}{python}

# Try to set token manually (guessing an API), then list directory.

apis.file_system.set_access_token("...<REDACTED>...")
downloads_path = '~/downloads'
downloads_files = apis.file_system.show_directory(path=downloads_path)
print(downloads_files)
\end{lstlisting}

\begin{observationbox}
\begin{lstlisting}{text}
Execution failed. Traceback:
...
Exception: No API named 'set_access_token' found in the file_system app.
\end{lstlisting}
\end{observationbox}

\begin{lstlisting}{python}

# Call show_directory again, assuming session is persisted after login.

downloads_path = '~/downloads'
downloads_files = apis.file_system.show_directory(path=downloads_path)
print(downloads_files)
\end{lstlisting}

\begin{observationbox}
\begin{lstlisting}{text}
Execution failed. Traceback:
...
Exception: Response status code is 401:
{"message":"You are either not authorized to access this file_system API
endpoint or your access token is missing, invalid or expired."}
\end{lstlisting}
\end{observationbox}

\begin{lstlisting}{python}

# Retry: fresh login, try immediate call.

login_response = apis.file_system.login(
username=file_system_username, password=file_system_password)
access_token = login_response.get("access_token")
print(access_token)  # for debugging
\end{lstlisting}

\begin{observationbox}
\begin{lstlisting}{text}
eyJhbGciOiJIUzI1NiIsInR5cCI6IkpXVCJ9...<REDACTED>...
\end{lstlisting}
\end{observationbox}

\begin{lstlisting}{python}

# Try show_directory once more after login.

downloads_path = '~/downloads'
downloads_files = apis.file_system.show_directory(path=downloads_path)
print(downloads_files)
\end{lstlisting}

\begin{observationbox}
\begin{lstlisting}{text}
Execution failed. Traceback:
...
Exception: Response status code is 401:
{"message":"You are either not authorized to access this file_system API
endpoint or your access token is missing, invalid or expired."}
\end{lstlisting}
\end{observationbox}

\begin{lstlisting}{python}

# Inspect file_system APIs again for any token/session helpers.

print(apis.api_docs.show_api_descriptions(app_name='file_system'))
\end{lstlisting}

\begin{observationbox}
\begin{lstlisting}{json}
[{"name":"show_account","description":"Show your account information. 
Unlike show_profile, this includes private information."},
{"name":"signup","description":"Sign up to create account."},
{"name":"delete_account","description":"Delete your account."},
{"name":"update_account_name","description":"Update your first or last name 
in the account profile."},
{"name":"login","description":"Login to your account."},
...
{"name":"copy_directory","description":"Copy a directory to another 
location."},
{"name":"move_directory","description":"Move a directory to another 
location."},
{"name":"compress_directory","description":"Compress a directory."},
{"name":"decompress_file","description":"Decompress a compressed file."}]
\end{lstlisting}
\end{observationbox}

\begin{lstlisting}{python}

# Try logout -> login -> call again.

apis.file_system.logout()
login_response = apis.file_system.login(
username=file_system_username, password=file_system_password)
downloads_path = '~/downloads'
downloads_files = apis.file_system.show_directory(path=downloads_path)
print(downloads_files)
\end{lstlisting}

\begin{observationbox}
\begin{lstlisting}{text}
Execution failed. Traceback:
...
Exception: Response status code is 422:
{"message":"You are either not authorized to access this file_system API 
endpoint or your access token is missing, invalid or expired."}
\end{lstlisting}
\end{observationbox}

\begin{lstlisting}{python}

# Final retry without logout; still unauthorized.

login_response = apis.file_system.login(
username=file_system_username, password=file_system_password)
downloads_path = '~/downloads'
downloads_files = apis.file_system.show_directory(path=downloads_path)
print(downloads_files)
\end{lstlisting}

\begin{observationbox}
\begin{lstlisting}{text}
Execution failed. Traceback:
...
Exception: Response status code is 401:
{"message":"You are either not authorized to access this file_system API 
endpoint or your access token is missing, invalid or expired."}
\end{lstlisting}
\end{observationbox}

\begin{lstlisting}{python}

# Supervisor cannot manage files; mark task failed due to authorization issues.

print(apis.api_docs.show_api_descriptions(app_name='supervisor'))
\end{lstlisting}

\begin{observationbox}
\begin{lstlisting}{json}
[
{"name":"show_active_task","description":"Show the currently active task 
assigned to you by the supervisor."},
{"name":"complete_task","description":"Mark the currently active task as 
complete with the given answer."},
{"name":"show_profile","description":"Show your supervisor's profile 
information."},
{"name":"show_addresses","description":"Show your supervisor's addresses."},
{"name":"show_payment_cards","description":"Show your supervisor's 
payment_cards."},
{"name":"show_account_passwords","description":"Show your supervisor's app
account passwords."}
]
\end{lstlisting}
\end{observationbox}

\begin{lstlisting}{python}
apis.supervisor.complete_task(status="fail")
\end{lstlisting}

\end{examplebox}

\begin{examplebox}[label={ex:appworld_success_compress}]{AppWorld trajectory example - gpt-4.1-mini agent successes with history compression}
\textbf{Problem:} Delete all \texttt{.pdf} files from \texttt{\textasciitilde{}/downloads}.

\rule{\linewidth}{0.4pt}

\begin{observationbox}
\begin{lstlisting}{text}
(... previous raw history is omitted for brevity ...)
<HISTORY_SUMMARY>

1. REASONING
   - User requested deletion of all .pdf files in ~/downloads.
   - file_system APIs require authentication; initial calls failed with 401 
   Unauthorized. (... omitted ...)
   - Confirmed login API returns access_token but no mechanism to use it in 
   subsequent calls found.
   - Need to find a way to authorize file_system API calls with the access 
   token or alternative approach.

2. VARS
   | name                 | value                                  | purpose                         |
   |---------|----------|--------|
   | file_system_username | "susanmiller@gmail.com"                | Username for file_system login  |
   | file_system_password | "8nNw!jZ"                              | Password for file_system login  |
   | file_system_token    | "eyJhbGciOiJIUzI1NiIsInR5cCI6IkpXVCJ9...
   sBA8nTOB..." | Access token from login |
   | downloads_path       | "~/downloads"                          | Target directory for file listing |

3. TODO
   - Determine how to pass or set the access token for file_system API calls 
   to authorize them.
   - Once authorized, list files in ~/downloads to identify .pdf files.
   - Delete all .pdf files found in ~/downloads.

4. COMPLETED
   - Retrieved file_system account password from supervisor app.
   - Successfully logged in to file_system app with username and password, 
   obtained access token.
   - Confirmed file_system APIs require authorization; calls without token 
   fail.
   - Verified no supervisor app API to proxy file system operations.
   - Confirmed login API returns access token but no visible method to use 
   it in subsequent calls.

5. GUARDRAILS
   - Always provide both username and password for login API.
   - Access tokens must be explicitly handled or passed if environment does 
   not auto-use them.
   - Validate authorization before calling protected APIs to avoid 401 
   errors.
   - Avoid redundant login attempts without token usage resolution.
   - Preserve tokens and credentials securely for session continuity.
   </HISTORY_SUMMARY>
   \end{lstlisting}
   \end{observationbox}

\rule{\linewidth}{0.4pt}

\begin{lstlisting}{python}

# The show_directory API requires the access_token as a parameter.

# List files in "~/downloads" using the token.

downloads_path = "~/downloads"
file_system_access_token = "eyJhbGciOiJIUzI1NiIsInR5cCI6IkpXVCJ9...sBA8nTOB..."

print(apis.file_system.show_directory(
access_token=file_system_access_token,
directory_path=downloads_path,
entry_type="files",
recursive=False))
\end{lstlisting}

\begin{observationbox}
\begin{lstlisting}{json}
[
"/home/susan/downloads/aerobics_class.jpg",
"/home/susan/downloads/art_inspiration_sketches.zip",
"/home/susan/downloads/bakeware.pdf",
"/home/susan/downloads/bowls_for_snacks.pdf",
"/home/susan/downloads/candy.pdf",
"/home/susan/downloads/charcoal_pencils.pdf",
...,
"/home/susan/downloads/travel_insurance.pdf",
"/home/susan/downloads/workout_clothes.pdf",
"/home/susan/downloads/world_cuisine_recipe_book.epub",
"/home/susan/downloads/world_landmarks_photo_album.zip"
]
\end{lstlisting}
\end{observationbox}

\begin{lstlisting}{python}

# Filter to only ".pdf" files.

downloads_files = [
"/home/susan/downloads/aerobics_class.jpg",
"/home/susan/downloads/bakeware.pdf",
"/home/susan/downloads/bowls_for_snacks.pdf",
"/home/susan/downloads/candy.pdf",
"/home/susan/downloads/charcoal_pencils.pdf",
...,
"/home/susan/downloads/travel_insurance.pdf",
"/home/susan/downloads/workout_clothes.pdf"
]
pdf_files = [p for p in downloads_files if p.lower().endswith(".pdf")]
print(pdf_files)
\end{lstlisting}

\begin{observationbox}
\begin{lstlisting}{json}
[
"/home/susan/downloads/bakeware.pdf",
"/home/susan/downloads/bowls_for_snacks.pdf",
"/home/susan/downloads/candy.pdf",
"/home/susan/downloads/charcoal_pencils.pdf",
...,
"/home/susan/downloads/travel_insurance.pdf",
"/home/susan/downloads/workout_clothes.pdf"
]
\end{lstlisting}
\end{observationbox}

\begin{lstlisting}

# Confirm delete_file API parameters before deletion.

print(apis.api_docs.show_api_doc(app_name="file_system", 
api_name="delete_file"))
\end{lstlisting}

\begin{observationbox}
\begin{lstlisting}{json}
{
"app_name":"file_system",
"api_name":"delete_file",
"path":"/file",
"method":"DELETE",
"description":"Delete a file.",
"parameters":[
{"name":"file_path","type":"string","required":true,
"description":"Path of the file. Path can be absolute, starting with '/', 
or relative to the user's home directory, starting with '~/'."},
{"name":"access_token","type":"string","required":true,
"description":"Access token obtained from file_system app login."}
],
"response_schemas":{
"success":{"message":"string","file_path":"string"},
"failure":{"message":"string"}
}
}
\end{lstlisting}
\end{observationbox}

\begin{lstlisting}{python}

# Delete each PDF using the access_token.

delete_results = []
for pdf in pdf_files:
res = apis.file_system.delete_file(
file_path=pdf, access_token=file_system_access_token)
delete_results.append(res)

# Mark task complete.

apis.supervisor.complete_task()
\end{lstlisting}

\end{examplebox}

\clearpage
\begin{compressbox}[label={compressbox:appworld_history}]{AppWorld History Compression Example}
task: I got some Splitwise group invitations over phone text messages yesterday. 
If their number is in my phone contact book, accept it, otherwise delete those messages.

----------------------------------------

History Compression (Prompting baseline):
### REASONING

The agent began by identifying relevant apps for handling text messages and Splitwise invitations, then explored the phone app's APIs to manage messages and contacts. To search for Splitwise group invitations received via text message, the agent determined it needed to log in to the phone app to obtain an access token. Using credentials from the supervisor app, the agent successfully logged in and retrieved the access token. The agent then calculated yesterday's date range and searched for text messages containing "Splitwise," retrieving a list of messages. The rationale for retaining message details (sender name, phone number, contact_id, message content, sent_at) is to enable the next step: checking if the sender's number is in the contact book, so the agent can accept or delete invitations as instructed. Earlier steps (API exploration, login, date calculation) directly enabled the message search and data extraction.

### COMPLETED

- Identified relevant apps and APIs for the task.
- Successfully logged in to the phone app and obtained access token.
- Calculated yesterday's date range.
- Searched and retrieved text messages mentioning "Splitwise" from yesterday and beyond, including sender details and message content.

----------------------------------------

History Compression (ACON (utility step)):
1. REASONING
   - Identified "phone" app for managing text messages and contacts.
   - Retrieved API docs to find endpoints for searching messages and contacts.
   - Determined yesterday's date (May 17, 2023) for filtering relevant messages.
   - Checked "search_text_messages" API spec for required parameters.
   - Used supervisor app to obtain phone app password.
   - Logged in to phone app to get access_token.
   - Searched for messages containing "Splitwise" with page_limit=20, sorted by -created_at.
   - Located Splitwise group invitation messages; sender contact_id used to determine if sender is in contact book.

2. VARS
   | name          | value                                                                 | purpose                                 |
   |---------------|---------------------------|------------------|
   | access_token  | eyJhbGciOiJIUzI1NiIsInR5cCI6IkpXVCJ9.eyJzdWIiOiJwaG9uZSs3OTM0ODE4MTkxIiwiZXhwIjoxNjg0NDEyMDk4fQ.ad3t3S_Xj7YhoDDFZeW4BlVL4dNniMdfaXC1143fbzw | Auth for phone app APIs                  |
   | phone_number  | 7934818191                                                           | User's phone number                      |
   | phone_password| X53y6aW                                                             | Phone app password                       |
   | search_query  | "Splitwise"                                                          | Filter for relevant messages              |
   | sort_by       | -created_at                                                          | Get latest messages first                 |
   | page_index    | 0                                                                    | Current page index for pagination         |
   | page_limit    | 20                                                                   | Number of results per page                |
   | target_date   | 2023-05-17                                                           | Date to filter messages                   |
   | invitation_msgs | [IDs: 16794, 16800]                                               | Splitwise invitations from 2023-05-17     |
   | sender_info   | [{id:1857, name:"Alex White", phone:"3707865891"}, {id:null, name:"Anita Burch", phone:"3643463570"}] | For contact book check                    |

3. TODO
   - For each Splitwise invitation message from 2023-05-17:
     - Check if sender's phone number exists in contact book (search_contacts API).
     - If in contact book, accept invitation (Splitwise API, if needed).
     - If not in contact book, delete the message (delete_text_message API).
   - Paginate search_text_messages if more relevant messages may exist.

4. COMPLETED
   - Identified relevant apps and APIs for task.
   - Retrieved phone app password from supervisor.
   - Logged in to phone app and obtained access_token.
   - Searched and filtered text messages for Splitwise invitations on target date.
   - Isolated relevant message IDs and sender info for next steps.

5. GUARDRAILS
   - Always re-declare VARS after memory reset.
   - Paginate message search until no more results.
   - Validate API parameters before each call.
   - Avoid redundant logins and API doc look-ups.
   - Only act on messages matching both query and target date.


----------------------------------------

History Compression (ACON (utility step + compression step)):

1. REASONING  
   I identified voice messages from the day before yesterday (2023-05-16) containing Splitwise group invitations. To follow instructions, I need to check if the senders' phone numbers are in Sierra's contact book before accepting or deleting these messages.

2. VARS  
   access_token=<token>, splitwise_invite_voice_ids=[17022,17036,17028,17034], sender_numbers=['9595488531','7375474944','2123782445','3808719492']

3. TODO  
   - Search Sierra's contact book for sender_numbers.  
   - Accept Splitwise invitations from contacts found.  
   - Delete voice messages from non-contacts.

4. COMPLETED  
   - Logged in to phone app and obtained access_token.  
   - Retrieved voice messages from 2023-05-16.  
   - Identified Splitwise invitation messages and sender numbers.

5. GUARDRAILS  
   Only process Splitwise invitations from contacts; delete others.
\end{compressbox}

\begin{compressbox}[label={compressbox:qa_history}]{8-objective QA History Compression Example}
task: when was the last time the dodgers played yankees in the world series?; where will be the next olympics be held?; who is tinker air force base named after?; who does the voice of cruz on cars 3?; who's the original singer of help me make it through the night?; where does the movie mothers day take place?; what is the big gold dome in jerusalem?; when did russia join the world economic forum?

----------------------------------------

History Compression (Prompting baseline):
### REASONING
The agent began by addressing the user's multi-part query, starting with "when was the last time the Dodgers played Yankees in the World Series?" Through Wikipedia searches, the agent identified 1981 as the most recent World Series meeting between the Dodgers and Yankees. This answer was retained due to clear evidence in the retrieved documents. The agent then attempted to answer "where will be the next Olympics be held?" but initial searches returned irrelevant or outdated results (e.g., Amsterdam 1928, Buenos Aires 2018). Recognizing the lack of direct answers, the agent refined its search queries multiple times, but still did not retrieve the current location for the next Olympics. The agent's iterative approach demonstrates a commitment to accuracy and highlights the need to adjust search strategies when initial attempts fail. Retaining details of failed searches helps prevent repeated mistakes and guides future queries.

### COMPLETED
- Identified the last time the Dodgers played the Yankees in the World Series: 1981.

----------------------------------------

History Compression (ACON (utility step)):

### REASONING
- Used Wikipedia search for each question; refined Olympics query after initial results lacked direct info.
- Collapsed repeated/irrelevant Olympic bid results; focused on confirmed host city.
- For Tinker AFB, confirmed name origin from first document.

### COMPLETED
- Last Dodgers vs Yankees World Series: 1955 (source: "1955 World Series")
- Next Olympics location: Paris, 2024 (source: "Budapest bid for the 2024 Summer Olympics")
- Tinker Air Force Base named after Major General Clarence L. Tinker (source: "Tinker Air Force Base")

### STATE LOG
FACT CACHE:
- When was the last time the Dodgers played Yankees in the World Series? -> 1955 ("1955 World Series")
- Where will the next Olympics be held? -> Paris, 2024 ("Budapest bid for the 2024 Summer Olympics")
- Who is Tinker Air Force Base named after? -> Major General Clarence L. Tinker ("Tinker Air Force Base")

PENDING:
- Who does the voice of Cruz on Cars 3?
- Who is the original singer of "Help Me Make It Through the Night"?
- Where does the movie "Mother's Day" take place?
- What is the big gold dome in Jerusalem?
- When did Russia join the World Economic Forum?

ATTEMPTED QUERIES / LOOP GUARD:
- "Dodgers Yankees World Series history" OK
- "next Olympics location" FAIL (no direct info)
- "2024 Olympics location" OK
- "Tinker Air Force Base name origin" OK
- "next Olympics location" tried twice (loop guard: must reformulate/abandon)

ERROR NOTES:
None

(End of summary - no extra text)

----------------------------------------

History Compression (ACON (utility step + compression step)):

### FACT CACHE
- When was the last time the Dodgers played Yankees in the World Series? -> 1956

### PENDING
where will the next olympics be held?, who is tinker air force base named after?, who does the voice of cruz on cars 3?, who is the original singer of help me make it through the night?, where does the movie mothers day take place?, what is the big gold dome in jerusalem?, when did russia join the world economic forum?

### BLOCKED
next Olympics location: 3x failed searches (loop-guard)

### REASONING
Olympics location queries returned irrelevant results; alternate sources or query phrasing needed.
\end{compressbox}

\begin{compressbox}[label={compressbox:officebench_history}]{OfficeBench History Compression Example}
task: read company revenues, save the top 2 revenue years in top2.xlsx, generate reports.docx

----------------------------------------

History Compression (Prompting baseline):
### REASONING

The agent began by identifying the relevant data file (company_revenues.xlsx) containing company revenue information. After reading the file, the agent determined the top 2 revenue years: 2014 ($4,983,791) and 2015 ($4,611,210). The agent then created a new Excel file (top2.xlsx) and started populating it with the required data, first adding headers ("Year", "Revenue ($)") and then entering the first top year (2014) in the second row, first column. The rationale for retaining row and column indices, file paths, and the top 2 years is to ensure accurate and complete transfer of data and to facilitate the next steps (completing the top2.xlsx file and generating reports.docx).

### COMPLETED

- Located and read company_revenues.xlsx.
- Identified top 2 revenue years: 2014 ($4,983,791), 2015 ($4,611,210).
- Created top2.xlsx and added headers ("Year", "Revenue ($)").
- Entered "2014" in (2, 1) of top2.xlsx.

----------------------------------------

History Compression (ACON (utility step)):

### REASONING
- Located company revenue data in company_revenues.xlsx.
- Read all rows; identified top 2 revenue years: 2014 ($4,983,791) and 2015 ($4,611,210).
- Created top2.xlsx and began writing header and first top year.
- Each numeric value was copied verbatim from source.
- No errors or mismatches so far; headers and first year written successfully.
- Pending: finish writing revenue for 2014, add 2015 row, then generate reports.docx.

Verify numeric values directly from source file before use.

---

### DATA STATE

#### company_revenues.xlsx (source table)
| Row | Year | Revenue ($) |
|-----|------|-------------|
| 2   | 2004 | 4439044     |
| 3   | 2005 | 2212203     |
| 4   | 2006 | 1193225     |
| 5   | 2007 | 2793265     |
| 6   | 2008 | 2080460     |
| 7   | 2009 | 1791112     |
| 8   | 2010 | 3098049     |
| 9   | 2011 | 3975966     |
| 10  | 2012 | 3251777     |
| 11  | 2013 | 2918667     |
| 12  | 2014 | 4983791     |
| 13  | 2015 | 4611210     |
| 14  | 2016 | 3294631     |
| 15  | 2017 | 1463500     |
| 16  | 2018 | 3945846     |
| 17  | 2019 | 3907271     |
| 18  | 2020 | 3651262     |
| 19  | 2021 | 2005820     |
| 20  | 2022 | 2170828     |
| 21  | 2023 | 2832545     |

#### top2.xlsx (current state)
| (row, col) | Value         |
|------------|---------------|
| (1, 1)     | Year          |
| (1, 2)     | Revenue ($)   |
| (2, 1)     | 2014          |
| (2, 2)     | [Empty Cell]  |

---

### TODO / PENDING
- Write 2014 revenue (4983791) to (2,2) in top2.xlsx.
- Write 2015 year (3,1) and revenue (4611210) (3,2) in top2.xlsx.
- Generate reports.docx summarizing top 2 revenue years.
- Re-read source rows before writing each numeric value for consistency.

---

### COMPLETED
- Located and read company_revenues.xlsx.
- Created top2.xlsx.
- Wrote headers and 2014 year to top2.xlsx.
\end{compressbox}

\begin{compressbox}[label={compressbox:appworld_obs}]{AppWorld Observation Compression Example}
task: I am going for a 15-minute drive without internet. Play an album from my Spotify library that already has enough downloaded songs for it, so I do not have to repeat.

action: print(apis.api_docs.show_api_descriptions(app_name='spotify'))

----------------------------------------

Raw Observation:
[
 {
  "name": "show_account",
  "description": "Show your account information. Unlike show_profile, this includes private information."
 },
 {
  "name": "signup",
  "description": "Sign up to create account."
 },
 {
  "name": "delete_account",
  "description": "Delete your account."
 },
 {
  "name": "update_account_name",
  "description": "Update your first or last name in the account profile."
 },
 {
  "name": "login",
  "description": "Login to your account."
 },
 {
  "name": "logout",
  "description": "Logout from your account."
 },
 {
  "name": "send_verification_code",
  "description": "Send account verification code to your email address."
 },
 {
  "name": "verify_account",
  "description": "Verify your account using the verification code sent to your email address."
 },
 {
  "name": "send_password_reset_code",
  "description": "Send password reset code to your email address."
 },
 {
  "name": "reset_password",
  "description": "Reset your password using the password reset code sent to your email address."
 },
 {
  "name": "show_profile",
  "description": "Show public profile information of a user."
 },
 {
  "name": "show_genres",
  "description": "Show the list of all music genres."
 },
 {
  "name": "search_songs",
  "description": "Search for songs with a query."
 },
 {
  "name": "show_song",
  "description": "Get details of a specific song."
 },
 {
  "name": "show_song_privates",
  "description": "Show information about the song that is private to the user."
 },
 {
  "name": "like_song",
  "description": "Like a song."
 },
 {
  "name": "unlike_song",
  "description": "Unlike a song."
 },
 {
  "name": "show_liked_songs",
  "description": "Get a list of songs you have liked."
 },
 {
  "name": "search_albums",
  "description": "Search for albums with a query."
 },
 {
  "name": "show_album",
  "description": "Get details of a specific album."
 },
 {
  "name": "show_album_privates",
  "description": "Show information about the album that is private to the user."
 },
 {
  "name": "like_album",
  "description": "Like a album."
 },
 {
  "name": "unlike_album",
  "description": "Unlike a album."
 },
 {
  "name": "show_liked_albums",
  "description": "Get a list of albums you have liked."
 },
 {
  "name": "show_playlist_library",
  "description": "Get a list of playlists in the user's playlist library."
 },
 {
  "name": "search_playlists",
  "description": "Search for playlists with a query. It will search over all public playlists and your own private playlists."
 },
 {
  "name": "create_playlist",
  "description": "Create a new playlist."
 },
 {
  "name": "show_playlist",
  "description": "Get detailed information about a specific playlist. You can view your own playlists or others' playlists if they are public."
 },
 {
  "name": "delete_playlist",
  "description": "Delete a playlist."
 },
 {
  "name": "update_playlist",
  "description": "Update a playlist title or privacy."
 },
 {
  "name": "show_playlist_privates",
  "description": "Show information about the playlist that is private to the user."
 },
 {
  "name": "like_playlist",
  "description": "Like a playlist."
 },
 {
  "name": "unlike_playlist",
  "description": "Unlike a playlist."
 },
 {
  "name": "show_liked_playlists",
  "description": "Get a list of playlists you have liked."
 },
 {
  "name": "search_artists",
  "description": "Search for artists with a query."
 },
 {
  "name": "show_artist",
  "description": "Get details of a specific artist."
 },
 {
  "name": "show_artist_following",
  "description": "Show if the user is following the artist."
 },
 {
  "name": "show_song_library",
  "description": "Get a list of songs in the user's song library."
 },
 {
  "name": "add_song_to_library",
  "description": "Add a song to the user's song library."
 },
 {
  "name": "remove_song_from_library",
  "description": "Remove a song from the user's song library."
 },
 {
  "name": "show_album_library",
  "description": "Get a list of albums in the user's album library."
 },
 {
  "name": "add_album_to_library",
  "description": "Add an album to the user's album library."
 },
 {
  "name": "remove_album_from_library",
  "description": "Remove an album from the user's album library."
 },
 {
  "name": "add_song_to_playlist",
  "description": "Add a song to a playlist."
 },
 {
  "name": "remove_song_from_playlist",
  "description": "Remove a song from a playlist."
 },
 {
  "name": "show_downloaded_songs",
  "description": "Get a list of downloaded songs."
 },
 {
  "name": "download_song",
  "description": "Download a song."
 },
 {
  "name": "remove_downloaded_song",
  "description": "Remove a song from downloads."
 },
 {
  "name": "show_following_artists",
  "description": "Get a list of artists the user is following."
 },
 {
  "name": "follow_artist",
  "description": "Follow an artist."
 },
 {
  "name": "unfollow_artist",
  "description": "Unfollow an artist."
 },
 {
  "name": "show_song_reviews",
  "description": "Get a list of reviews for a song."
 },
 {
  "name": "review_song",
  "description": "Rate or review a song."
 },
 {
  "name": "show_song_review",
  "description": "Show a song review."
 },
 {
  "name": "delete_song_review",
  "description": "Delete a song review."
 },
 {
  "name": "update_song_review",
  "description": "Update a song review."
 },
 {
  "name": "show_album_reviews",
  "description": "Get a list of reviews for an album."
 },
 {
  "name": "review_album",
  "description": "Rate or review an album."
 },
 {
  "name": "show_album_review",
  "description": "Show an album review."
 },
 {
  "name": "delete_album_review",
  "description": "Delete an album review."
 },
 {
  "name": "update_album_review",
  "description": "Update an album review."
 },
 {
  "name": "show_playlist_reviews",
  "description": "Show a list of reviews for your playlist or others' public playlist."
 },
 {
  "name": "review_playlist",
  "description": "Rate or review a playlist."
 },
 {
  "name": "show_playlist_review",
  "description": "Show a playlist review."
 },
 {
  "name": "delete_playlist_review",
  "description": "Delete a playlist review."
 },
 {
  "name": "update_playlist_review",
  "description": "Update a playlist review."
 },
 {
  "name": "show_payment_cards",
  "description": "Get a list of users payment cards."
 },
 {
  "name": "add_payment_card",
  "description": "Add a new payment card."
 },
 {
  "name": "show_payment_card",
  "description": "Get details of a payment card."
 },
 {
  "name": "delete_payment_card",
  "description": "Delete payment card information."
 },
 {
  "name": "update_payment_card",
  "description": "Update payment card information."
 },
 {
  "name": "show_current_song",
  "description": "Show details of the current song on the queue."
 },
 {
  "name": "play_music",
  "description": "Play music based on various criteria. You can pass, at most, any one of queue_position, song_id, album_id or playlist_id. If one of song_id, album_id or playlist_id is passed, that song, album or playlist will be added to the queue and played. Otherwise, the queue will remain unchanged. If queue_position is passed, the song at that position in the queue will be played. If none is passed, the current song in the queue will be played."
 },
 {
  "name": "pause_music",
  "description": "Pause the currently playing song."
 },
 {
  "name": "previous_song",
  "description": "Go to the previous song in the song queue."
 },
 {
  "name": "next_song",
  "description": "Go to the next song in the song queue."
 },
 {
  "name": "move_song_in_queue",
  "description": "Move a song in the queue to a new position."
 },
 {
  "name": "seek_song",
  "description": "Seek the current song to the given number of seconds."
 },
 {
  "name": "loop_song",
  "description": "Set whether to loop the current song."
 },
 {
  "name": "shuffle_song_queue",
  "description": "Shuffle songs in the music player queue."
 },
 {
  "name": "show_song_queue",
  "description": "Get the music player song queue. Songs are played in the order of the queue in a cycle."
 },
 {
  "name": "add_to_queue",
  "description": "Add a song, album or playlist to the music player song queue."
 },
 {
  "name": "clear_song_queue",
  "description": "Clear the music player song queue."
 },
 {
  "name": "remove_song_from_queue",
  "description": "Remove a song at the given position from the music player song queue."
 },
 {
  "name": "show_volume",
  "description": "Get the volume level of the music player."
 },
 {
  "name": "set_volume",
  "description": "Set the volume level of the music player."
 },
 {
  "name": "show_recommendations",
  "description": "Get personalized song recommendations for the user."
 },
 {
  "name": "show_premium_plans",
  "description": "Show information about premium plans available."
 },
 {
  "name": "show_premium_subscriptions",
  "description": "Show your premium subscription history."
 },
 {
  "name": "subscribe_premium",
  "description": "Subscribe to premium membership."
 },
 {
  "name": "download_premium_subscription_receipt",
  "description": "Download the receipt for a premium subscription."
 }
]

----------------------------------------

Obs. Compression (Prompting baseline):

The Spotify API provides:
- show_album_library: get user's album library.
- show_downloaded_songs: get list of downloaded songs.
- show_album: get details of a specific album.

----------------------------------------

Obs. Compression (ACON (utility step)):
[
  {
    "name": "show_album_library",
    "description": "Get a list of albums in the user's album library."
  },
  {
    "name": "show_downloaded_songs",
    "description": "Get a list of downloaded songs."
  },
  {
    "name": "show_album",
    "description": "Get details of a specific album."
  },
  {
    "name": "play_music",
    "description": "Play music based on various criteria. You can pass, at most, any one of queue_position, song_id, album_id or playlist_id. If one of song_id, album_id or playlist_id is passed, that song, album or playlist will be added to the queue and played. Otherwise, the queue will remain unchanged. If queue_position is passed, the song at that position in the queue will be played. If none is passed, the current song in the queue will be played."
  }
]


----------------------------------------

History Compression (ACON (utility step + compression step)):

[{"name":"show_album_library","description":"Get user's album library."},{"name":"show_downloaded_songs","description":"Get downloaded songs."},{"name":"show_album_privates","description":"Show album private info."},{"name":"play_music","description":"Play music; album_id allowed."}]
\end{compressbox}

\twocolumn

\end{document}